\documentclass[reqno]{style/amsart}
\usepackage[margin=1.4in]{geometry}
\numberwithin{equation}{section}

\usepackage[utf8]{inputenc} 
\usepackage[T1]{fontenc}    
\usepackage{url}            
\usepackage{booktabs}       
\usepackage{amsfonts}       
\usepackage{nicefrac}       
\usepackage{microtype}      

\usepackage[colorlinks=true, citecolor=blue]{hyperref}
\usepackage{amsmath, amssymb, amsthm}
\usepackage{enumitem}
\usepackage{color}
\usepackage{algorithm}
\usepackage{algcompatible}
\usepackage[noend]{algpseudocode} 
\usepackage{comment}
\usepackage{graphicx} 

\usepackage{dsfont}
\usepackage{mathtools}

\DeclareMathOperator*{\argmin}{argmin}

\usepackage{subfig}
\usepackage[numbers]{natbib}
\newcommand{\probo}{{\fontfamily{lmss}\selectfont{ProBO}}}
\newcommand{\probosp}{{\fontfamily{lmss}\selectfont{ProBO }}}

\DeclarePairedDelimiter\floor{\lfloor}{\rfloor}
\newcommand{\norm}[1]{\left\lVert #1 \right\rVert}

\newenvironment{changemargin}[2]{%
\begin{list}{}{%
\setlength{\topsep}{0pt}%
\setlength{\leftmargin}{#1}%
\setlength{\rightmargin}{#2}%
\setlength{\listparindent}{\parindent}%
\setlength{\itemindent}{\parindent}%
\setlength{\parsep}{\parskip}%
}%
\item[]}{\end{list}}

\begin{document}
\title[\probo: Versatile Bayesian Optimization Using Any PPL]
{\probo: Versatile Bayesian Optimization Using Any Probabilistic
Programming Language}
\author[W. Neiswanger]{Willie Neiswanger$^{1,*}$}
\author[K. Kandasamy]{Kirthevasan Kandasamy$^{2,\dagger}$}
\author[B. P\'oczos]{Barnab\'as P\'oczos$^1$}
\author[J. Schneider]{Jeff Schneider$^1$}
\author[E. P. Xing]{Eric P. Xing$^{1,3}$}
\thanks{$^*$Correspondence to \texttt{willie@cs.cmu.edu}. $^\dagger$Work done
        while at CMU.  Last revised \today.}
\maketitle
\begin{abstract}
Optimizing an expensive-to-query function is a common task in science and
engineering, where it is beneficial to keep the number of queries to a minimum.
A popular strategy is Bayesian optimization (BO), which leverages probabilistic
models for this task.  Most BO today uses Gaussian processes (GPs), or a few
other surrogate models.  However, there is a broad set of Bayesian modeling
techniques that could be used to capture complex systems and reduce the number
of queries in BO.  Probabilistic programming languages (PPLs) are modern tools
that allow for flexible model definition, prior specification, model
composition, and automatic inference.  In this paper, we develop \probo, a BO
procedure that uses only standard operations common to most PPLs. This allows a
user to drop in a model built with an arbitrary PPL and use it directly in BO.
We describe acquisition functions for \probo, and strategies for efficiently
optimizing these functions given complex models or costly inference procedures.
Using existing PPLs, we implement new models to aid in a few challenging
optimization settings, and demonstrate these on model hyperparameter and
architecture search tasks.
\end{abstract}

\section{Introduction}
Bayesian optimization (BO) is a popular method for zeroth-order
optimization of an unknown (``black box'') system.  A BO procedure iteratively
queries the system to yield a set of input/output data points, computes the
posterior of a Bayesian model given these data, and optimizes an acquisition
function defined on this posterior in order to determine the next point to
query.


BO involves performing inference and optimization to choose each point to query,
which can incur a greater computational cost than simpler stategies, but may be
ultimately beneficial in settings where queries are expensive.  Specifically,
if BO
can reach a good optimization objective in fewer iterations than simpler methods,
it may be effective in cases where the expense of queries far outweighs the
extra cost of BO. Some examples of this are in science and engineering, where a
query could involve synthesizing and measuring the properties of a material,
collecting metrics from an industrial process, or training a large machine
learning model, which can be expensive in cost, time, or human labor. 

The most common model used in BO is the \emph{Gaussian process} (GP), for which
we can compute many popular acquisition functions. There has also been some
work deriving BO procedures for other flexible models including random forests
\cite{hutter2011sequential} and neural networks \cite{snoek2015scalable}.
In this paper, we argue that more-sophisticated models that better capture the
details of a system can help reduce the number of iterations needed in BO, and
allow for BO to be effectively used in custom and complex settings.  For
example, systems may have complex noise \cite{shah2014student,
jylanki2011robust}, yield multiple types of observations
\cite{gardner2014bayesian}, depend on covariates \cite{krause2011contextual},
have interrelated subsystems \cite{swersky2013multi}, and more.
%
%
%
To accurately capture these systems, we may want to design custom models using
a broader library of Bayesian tools and techniques. For example, we may want to
compose models---such as GPs, latent factor (e.g.  mixture) models, deep
Bayesian networks, hierarchical regression models---in various ways, and use
them in BO.

\emph{Probabilistic programming languages} (PPLs) are modern tools for
specifying Bayesian models and performing inference.  They allow for easy
incorporation of prior knowledge and model structure, composition of models,
quick deployment, and automatic inference, often in the form of samples from or
variational approximations to a posterior distribution.  PPLs may be used to
specify and run inference in a variety of models, such as graphical models,
GPs, deep Bayesian models, hierarchical models, and implicit (simulator-based)
models, to name a few \cite{carpenter2015stan, lunn2000winbugs,
salvatier2016probabilistic, tran2016edward, bingham2018pyro, minka2012infer,
hodlr, mansinghka2014venture, wood-aistats-2014, dillon2017tensorflow}.
We would like to be able to build an arbitrary model with any PPL and then
automatically carry out BO with this model. However, this comes with a few
challenges. In BO with GPs, we have the posterior in closed-form, and use this
when computing and optimizing acquisition functions. PPLs, however, use a
variety of approximate inference procedures, which can be costly to run and
yield different posterior representations (e.g.  samples
\cite{carpenter2015stan, wood-aistats-2014}, variational approximations
\cite{tran2016edward, bingham2018pyro}, implicit models
\cite{huszar2017variational,tran2017hierarchical}, or amortized distributions 
\cite{ritchie2016deep, le2016inference}).  We need a method that can compute
and optimize acquisition functions automatically, given the variety of
representations, and efficiently, making judicious use of PPL procedures.

Towards this end, we develop \probo, a BO system
for PPL models, which computes and optimizes acquisition functions via
operations that can be implemented in a broad variety of PPLs.  This system
comprises algorithms that cache and use these operations efficiently, which
allows it to be used in practice given complex models and expensive inference
procedures.  The overall goal of \probosp is to allow a custom model written in
an arbitrary PPL to be ``dropped in'' and immediately used in BO.

This paper has two main contributions: (1) We present \probo, a system for
versatile Bayesian optimization using models from any PPL. (2) We describe
optimization settings that are difficult
for standard BO methods and models, and then use PPLs to implement new models
for these settings, which are dropped into \probosp and show good optimization
performance.  Our open source release of \probosp is available at
\url{https://github.com/willieneis/ProBO}.



\section{Related Work}
\label{sec:relatedwork}
A few prior works make connections between PPLs and BO. BOPP
\citep{rainforth2016bayesian} describes a BO method for marginal maximum a
posteriori (MMAP) estimates of latent variables in a probabilistic program.
This work relates BO and PPLs, but differs from us in that the goal of BOPP is
to use BO (with GP models) to help estimate latent variables in a given PPL,
while we focus on using PPLs to build new surrogate models for BO.

BOAT \citep{dalibard2017boat} provides a custom PPL involving composed Gaussian
process models with parametric mean functions, for use in BO. For these models,
exact inference can be performed and the expected improvement acquisition
directly used. This work has similar goals as us, though we instead aim to
provide a system that can be applied to models from \emph{any} existing PPL
(not constrained to a certain family of GP models),
and specifically with PPLs that use approximate inference algorithms where we
cannot compute acquisition functions in standard ways.

\sectionnocaps{\probo}
\label{sec:framework}
We first describe a general abstraction for PPLs, and use this abstraction to
define \probosp and present algorithms for computing a few acquisition
functions.  We then show how to efficiently optimize these acquisition
functions.

\subsection{Abstraction for Probabilistic Programs}
\label{sec:ppformalism}
Suppose we are modeling a system which, given an input $x \in \mathcal{X}$,
yields observations $y \in \mathcal{Y}$, written $y \sim s(x)$. 
Let $f:\mathcal{Y} \rightarrow \mathbb{R}$ be an objective function that maps
observations $y$ to real values.
Observing the system $n$ times at different inputs yields a
dataset $\mathcal{D}_n = \{(x_i,y_i)\}_{i=1}^n$.  Suppose we have a
Bayesian model for $\mathcal{D}_n$, with likelihood $p(\mathcal{D}_n|z)$
$=$ $\prod_{i=1}^n p(y_i | z ; x_i)$, where $z\in \mathcal{Z}$ are latent
variables.
We define the joint model PDF to be $p(\mathcal{D}_n,z) = p(z)
p(\mathcal{D}_n|z)$, where $p(z)$ is the PDF of the prior on $z$.  The
posterior PDF is then $p(z|\mathcal{D}_n) = p(\mathcal{D}_n,z) /
\int p(\mathcal{D}_n,z) dz$.

Our abstraction assumes three basic PPL operations:
\begin{enumerate}[topsep=5pt,itemsep=5pt,parsep=0pt]
  \item $\texttt{inf}(\mathcal{D})$: given data $\mathcal{D}$, this runs an
    inference algorithm and returns an object \texttt{post}, which is a
    PPL-dependent representation of the posterior distribution. 
  \item $\texttt{post}(s)$: given a seed $s \in \mathbb{Z}^+$, this returns a
    sample from the posterior distribution.
  \item $\texttt{gen}(x,z,s)$: given an input $x \in \mathcal{X}$, a latent
    variable $z \in \mathcal{Z}$, and a seed $s \in \mathbb{Z}^+$, this returns
    a sample from the generative distribution $p(y|z;x)$.
\end{enumerate}
Note that \texttt{post} and \texttt{gen} are deterministic, i.e. for a fixed seed $s$,
\texttt{post}/\texttt{gen} produce the same output each time they are called.


\newlength{\minipagewidth}
\newlength{\minipageoffset}
\newlength{\columngap}
\newlength{\pretitlegap}
\newlength{\posttitlegap}
\newlength{\afterlettergap}

\setlength{\minipagewidth}{0.3\textwidth}
\setlength{\minipageoffset}{0mm}
\setlength{\columngap}{-3mm}
\setlength{\pretitlegap}{2mm}
\setlength{\posttitlegap}{3mm}
\setlength{\afterlettergap}{1mm}

\begin{figure*}[tp]
  \centering

  \makebox[\textwidth]{\makebox[1.25\textwidth]{
    \hspace{-4mm}
    \begin{minipage}[b]{\minipagewidth}
      \hspace{\minipageoffset}
      \centering
      {(a) \hspace{\afterlettergap} $a_\text{EI}(x)$}
    \end{minipage}
    \hspace{\columngap}
    \begin{minipage}[b]{\minipagewidth}
      \hspace{\minipageoffset}
      \centering
      {(b) \hspace{\afterlettergap} $a_\text{PI}(x)$}
    \end{minipage}
    \hspace{\columngap}
    \begin{minipage}[b]{\minipagewidth}
      \hspace{\minipageoffset}
      \centering
      {(c) \hspace{\afterlettergap} $a_\text{UCB}(x)$}
    \end{minipage}
    \hspace{\columngap}
    \begin{minipage}[b]{\minipagewidth}
      \hspace{\minipageoffset}
      \centering
      {(d) \hspace{\afterlettergap} $a_\text{TS}(x)$}
    \end{minipage}
  }}

  \makebox[\textwidth]{\makebox[1.25\textwidth]{
    \begin{minipage}[b]{\minipagewidth}
      \includegraphics[width=\textwidth]{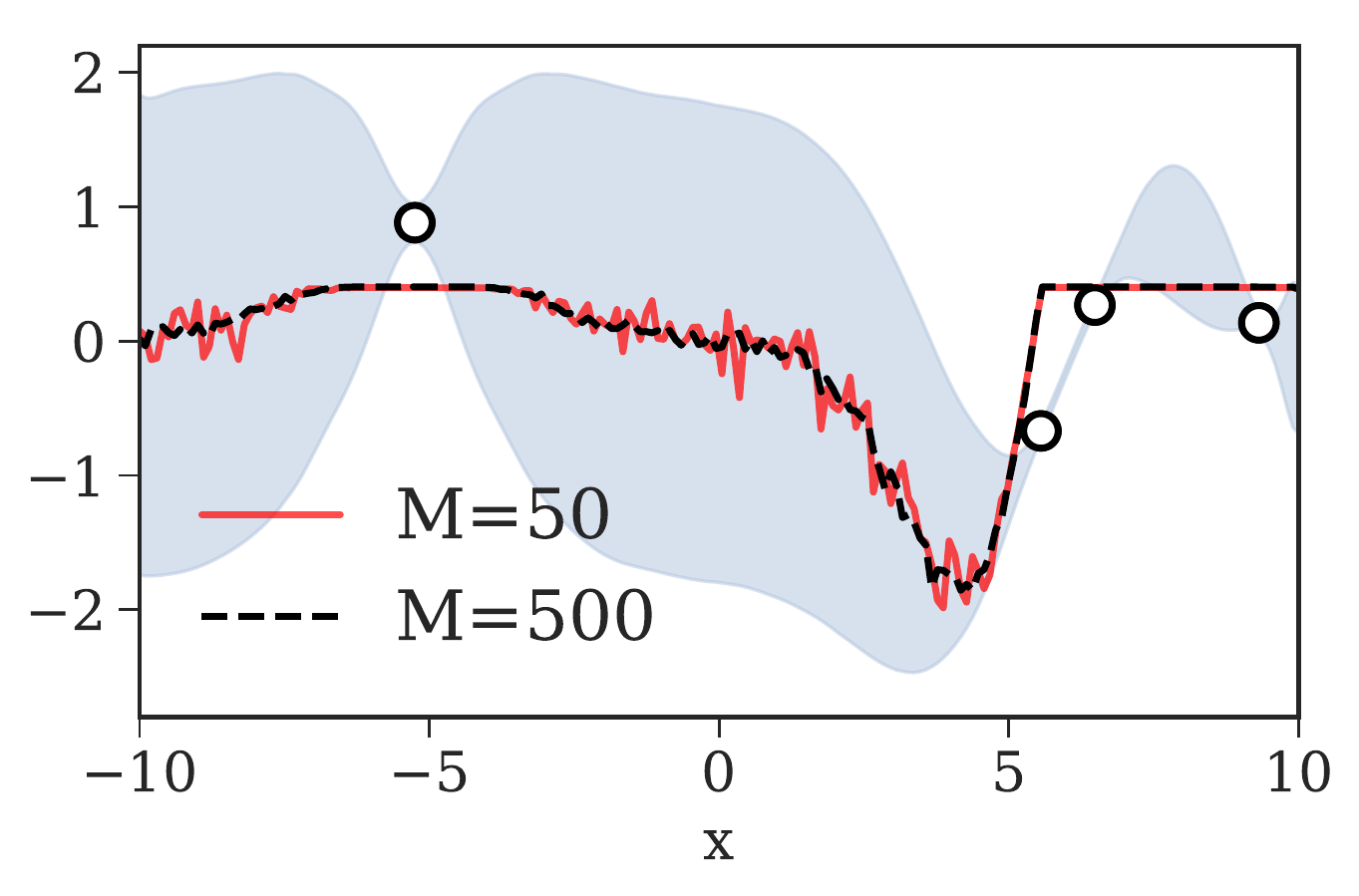}
    \end{minipage}
    \hspace{\columngap}
    \begin{minipage}[b]{\minipagewidth}
      \includegraphics[width=\textwidth]{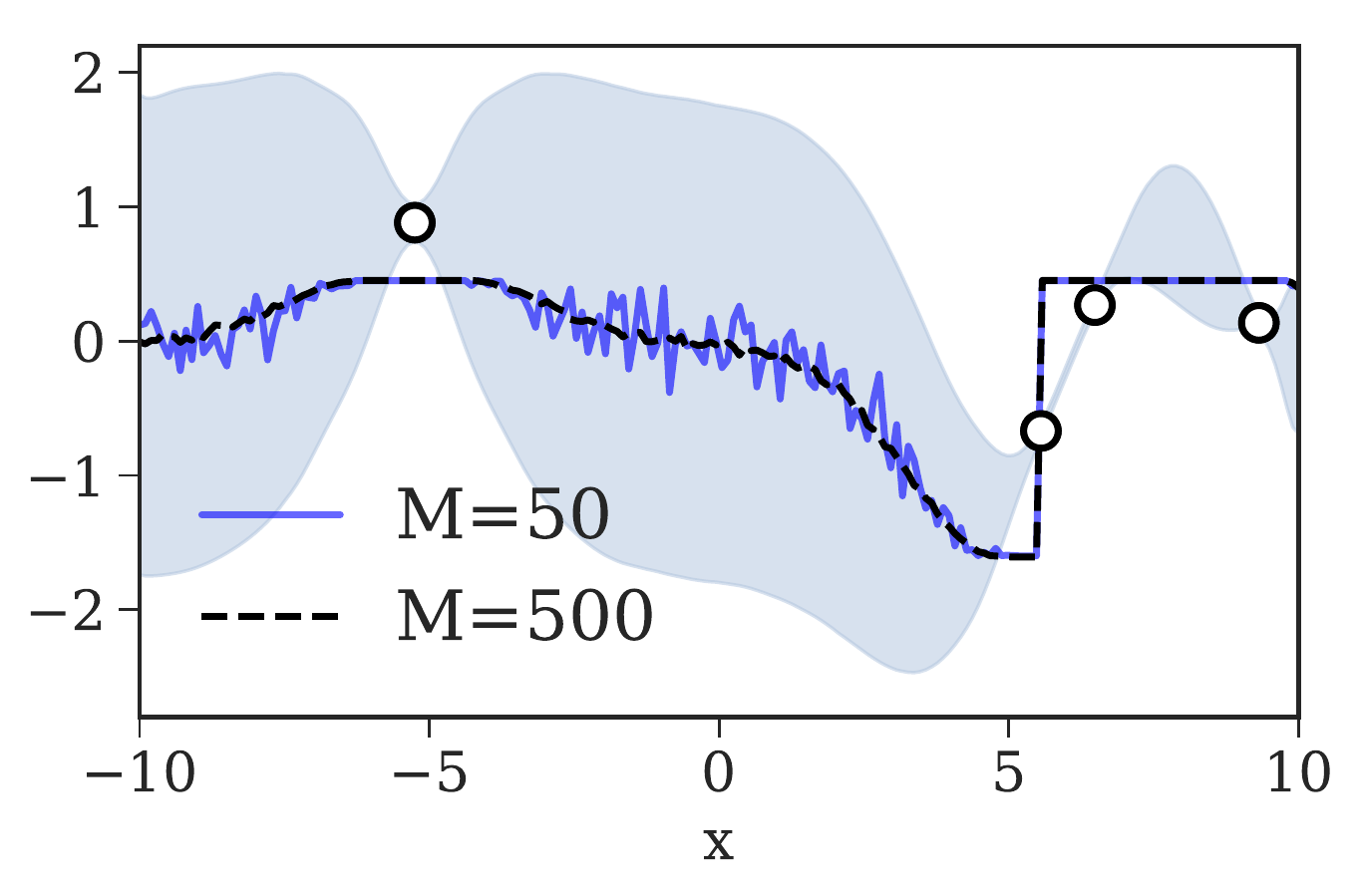}
    \end{minipage}
    \hspace{\columngap}
    \begin{minipage}[b]{\minipagewidth}
      \includegraphics[width=\textwidth]{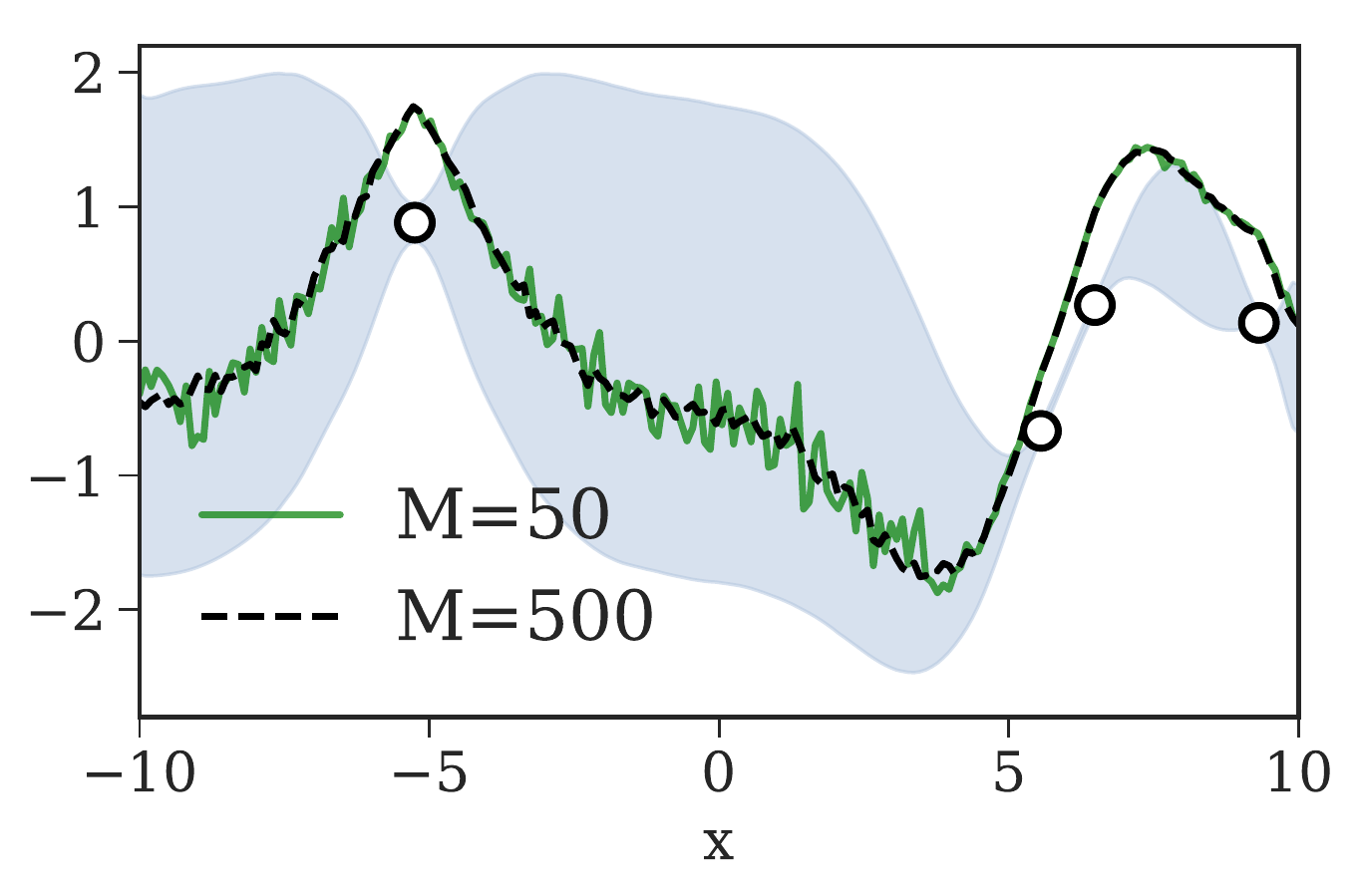}
    \end{minipage}
    \hspace{\columngap}
    \begin{minipage}[b]{\minipagewidth}
      \includegraphics[width=\textwidth]{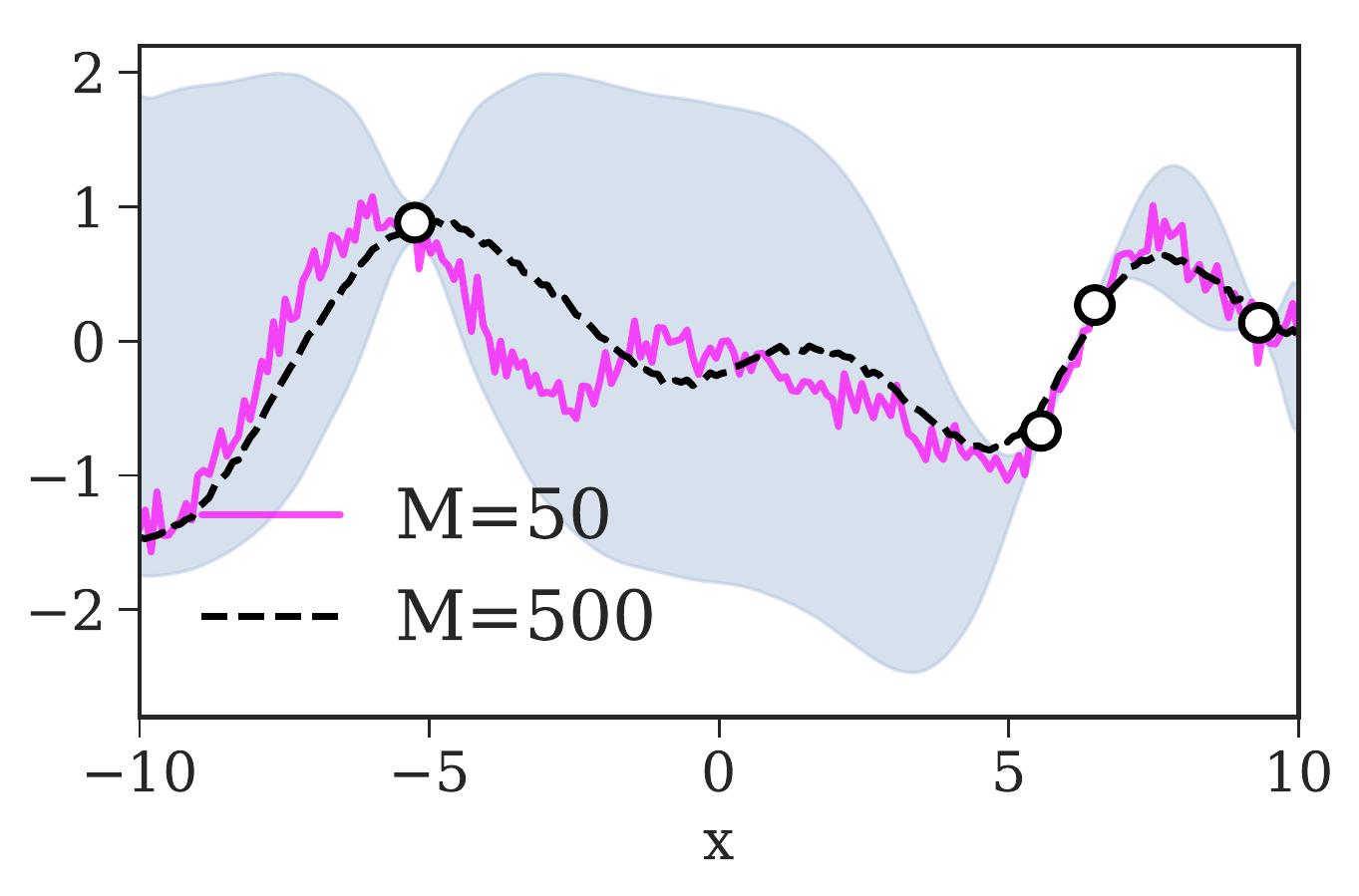}
    \end{minipage}
  }}

  \vspace{-0.1in}
  \caption{
    \small 
    Visualizations of PPL acquisition functions $a(x)$ given in
    Algs.~\ref{alg:ei}-\ref{alg:ts} for use in \probo. In each plot, the data
    and posterior predictive distribution are shown, and $a(x)$ is given for
    two fidelities: $M=50$ (solid color line) and $M=500$ (dashed black line).
    \label{fig:bbacq}
  }
  \vspace{0.05in}
\end{figure*}

\subsubsection*{Scope.}
This abstraction applies to a number of PPLs, which use a variety of inference
strategies and compute different representations for the posterior. For
example, in PPLs using Markov chain Monte Carlo (MCMC) or sequential Monte
Carlo (SMC) algorithms \cite{carpenter2015stan, salvatier2016probabilistic,
wood-aistats-2014}, \texttt{inf} computes a set of posterior samples and
\texttt{post} draws uniformly from this set. For PPLs using variational
inference (VI), implicit models,  or exact inference methods
\cite{tran2016edward, bingham2018pyro, gpy2014}, \texttt{inf} computes the
parameters of a distribution, and \texttt{post} draws a sample from this
distribution.  In amortized or compiled inference methods
\cite{ritchie2016deep, le2016inference, kingma2013auto}, \texttt{inf} trains or
calls a pretrained model that maps observations to a posterior or proposal
distribution, and \texttt{post} samples from this.




\subsection{Main Procedure}
Recall that we use PPLs to model a system $s$, which yields
observations $y \sim s(x)$ given a query $x$,
and where $f(y):\mathcal{Y} \rightarrow \mathbb{R}$ denotes the objective value
that we want to optimize.  The goal of \probosp is to return
$x^* = \argmin_{x \in \mathcal{X}} \mathbb{E}_{y \sim s(x)} \left[ f(y)
\right]$.
We give \probosp in Alg.~\ref{alg:probo}. Each iteration consists
of four steps: call an inference procedure \texttt{inf}, select an input $x$ by
optimizing an acquisition function $a$ that calls \texttt{post} and
\texttt{gen}, observe the system at $x$, and add new observations to the
dataset.
\begin{algorithm}[h!]
    \caption{\probo$\left(\mathcal{D}_0,\texttt{inf},\texttt{gen}\right)$}
    \label{alg:probo}
    \begin{algorithmic}[1]
      \For{$n = 1,\ldots,N$}
        \State $\texttt{post} \leftarrow \texttt{inf}(\mathcal{D}_{n-1})$
          \Comment{Run inference algorithm to compute \texttt{post}}
        \State $x_n \leftarrow \argmin_{x \in \mathcal{X}}
          a(x, \texttt{post}, \texttt{gen})$ \Comment{Optimize acquisition
          using \texttt{post} and \texttt{gen}}
        \State $y_n \sim s(x_n)$ \Comment{Observe system at $x_n$}
        \State $\mathcal{D}_n \leftarrow \mathcal{D}_{n-1} \cup (x_n,y_n)$
          \Comment{Add new observations to dataset}
      \EndFor 
      \State Return $\mathcal{D}_N$.
    \end{algorithmic}
\end{algorithm}

Note that acquisition optimization (line 3) involves only \texttt{post} and
\texttt{gen}, while \texttt{inf} is called separately beforehand.
We discuss the computational benefits of this, and the cost of \probo, in more
detail in Sec.~\ref{sec:computationalconsiderations}.
Also note that many systems can have extra observations $y \in \mathcal{Y}$, in
addition to the objective value, that provide information which aids
optimization \citep{wu2017bayesian, astudillo2017multi, swersky2013multi}.
For this reason, our formalism explicitly separates system observations $y$
from their objective values $f(y)$. We show examples of models that take
advantage of extra observations in Sec.~\ref{sec:examples}.


In the following sections, we develop algorithms for computing acquisition
functions $a(x)$ automatically using only \texttt{post} and \texttt{gen}
operations, without requiring any model specific derivation.  We refer to these
as \emph{PPL acquisition functions}.

%

\subsection{PPL Acquisition Functions via \texttt{post} and \texttt{gen}}
\label{sec:bbacq}

In \probosp (Alg.~\ref{alg:probo}), we denote the PPL acquisition
function with $a(x,\texttt{post},\texttt{gen})$.  Each PPL acquisition
algorithm includes a parameter $M$, which represents the fidelity of the
approximation quality of $a$.  We will describe an adaptive method for choosing
$M$ during acquisition optimization in Sec.~\ref{sec:mf}. Below, we will make
use of the posterior predictive distribution, which is defined to be
$p(y|\mathcal{D}_n;x)$ $=$
$\mathbb{E}_{p(z|\mathcal{D}_n)}\left[p(y|z;x)\right]$.

There are a number of popular acquisition functions used commonly in Bayesian
optimization, such as expected improvement (EI) \citep{movckus1975bayesian},
probability of improvement (PI) \citep{kushner1964new}, GP upper confidence
bound (UCB) \citep{srinivas2009gaussian}, and Thompson sampling (TS)
\citep{thompson1933likelihood}.
Here, we propose a few simple acquisition estimates that can be computed 
with \texttt{post} and \texttt{gen}.  Specifically, we give algorithms for EI
(Alg.~\ref{alg:ei}), PI (Alg.~\ref{alg:pi}), UCB (Alg.~\ref{alg:ucb}), and TS
(Alg.~\ref{alg:ts}) acquistion strategies, though similar algorithms could be
used for other acquisitions involving expectations or statistics of either
$p(y|D_n;x)$ or $p(y|z;x)$.

We now describe the PPL acquisition functions given in
Alg.~\ref{alg:ei}-\ref{alg:ts} in more detail, and discuss the approximations
given by each. Namely, we show that these yield versions of exact acquisition
functions as $M \rightarrow \infty$. These algorithms are related to Monte
Carlo acquisition function estimates for GP models \citep{snoek2012practical,
hennig2012entropy, hernandez2015predictive}, which have been developed for
specific acquisition functions.


\subsubsection*{Expected Improvement (EI), Alg.~\ref{alg:ei}}
Let $\mathcal{D}$ be the data at a given iteration of \probo.  In our setting,
the expected improvement (EI) acquisition function will return the
expected improvement that querying the system at $x \in \mathcal{X}$ will have
over the minimal observed objective value, $f_\text{min} = \min_{y \in
\mathcal{D}} f(y)$. We can write the exact EI acquisition function as 
\begin{align}
a_\text{EI}^*(x) = \int \mathds{1} \left\{f(y) \leq f_\text{min} \right\}
    \left(f_\text{min} - f(y) \right) p\left( y|\mathcal{D};x \right) dy.
\end{align}

In Alg.~\ref{alg:ei}, for a sequence of steps $m=1,\ldots,M$, we draw $z_m \sim
p(z|\mathcal{D})$ and $y_m \sim p(y|z_m;x)$ via \texttt{post} and
\texttt{gen}, and then compute $\lambda(y_{1:M}) = \sum_{m=1}^M
\mathds{1}\left[ f(y_m) \leq f_\text{min} \right] (f_\text{min} - f(y_m))$.
Marginally, $y_m$ is drawn from the posterior predictive distribution, i.e.
$y_m$ $\sim$ $\mathbb{E}_{p(z|\mathcal{D})}\left[p(y|z;x)\right]$ $=$
$p(y|\mathcal{D};x)$. Therefore, as the number of calls $M$ to \texttt{post}
and \texttt{gen} grows,
$a_\text{EI}(x)$ $\rightarrow$ $a_\text{EI}^*(x)$ (up to a multiplicative
constant) at a rate of $O(\sqrt{M})$.

\noindent \begin{minipage}[t]{0.52\textwidth}
  \begin{algorithm}[H]
      \caption{\hspace{1mm}$a_\text{EI}\left(x,\texttt{post},\texttt{gen}\right)$
        \hfill $\triangleright$ EI}
      \label{alg:ei}
      \begin{algorithmic}[1]
        \For{$m = 1,\ldots,M$}
          \State $z_m \leftarrow \texttt{post}(s_m)$
          \State $y_m \leftarrow \texttt{gen}(x,z_m,s_m)$
        \EndFor
        \State $f_\text{min} \leftarrow \min_{y\in\mathcal{D}} f(y)$
        \State Return $\sum_{m=1}^M \mathds{1}\left[ f(y_m) \leq
          f_\text{min} \right] (f_\text{min} - f(y_m))$
      \end{algorithmic}
  \end{algorithm}
\end{minipage}
\hfill
\begin{minipage}[t]{0.45\textwidth}
  \begin{algorithm}[H]
      \caption{\hspace{1mm}$a_\text{PI}\left(x,\texttt{post},\texttt{gen}\right)$
        \hfill$\triangleright$ PI}
      \label{alg:pi}
      \begin{algorithmic}[1]
        \For{$m = 1,\ldots,M$}
          \State $z_m \leftarrow \texttt{post}(s_m)$
          \State $y_m \leftarrow \texttt{gen}(x,z_m,s_m)$
        \EndFor
        \State $f_\text{min} \leftarrow \min_{y\in\mathcal{D}} f(y)$
        \State Return $\sum_{m=1}^M \mathds{1}\left[ f(y_m) \leq
          f_\text{min} \right]$
      \end{algorithmic}
  \end{algorithm}
\end{minipage}
\begin{minipage}[t]{0.52\textwidth}
  \begin{algorithm}[H]
      \caption{\hspace{1mm}$a_\text{UCB}\left(x,\texttt{post},\texttt{gen}\right)$
        \hfill$\triangleright$ UCB}
      \label{alg:ucb}
      \begin{algorithmic}[1]
        \For{$m = 1,\ldots,M$}
          \State $z_m \leftarrow \texttt{post}(s_m)$
          \State $y_m \leftarrow \texttt{gen}(x,z_m,s_m)$
        \EndFor
        \State Return $\widehat{\text{LCB}}\left(f(y_m)_{m=1}^M\right)$
          \Comment{See text for details}
      \end{algorithmic}
  \end{algorithm}
\end{minipage}
\hfill
\begin{minipage}[t]{0.45\textwidth}
  \begin{algorithm}[H]
      \caption{\hspace{1mm}$a_\text{TS}\left(x,\texttt{post},\texttt{gen}\right)$
        \hfill$\triangleright$ TS}
      \label{alg:ts}
      \begin{algorithmic}[1]
        \State $z \leftarrow \texttt{post}(s_1)$
        \For{$m = 1,\ldots,M$}
          \State $y_m \leftarrow \texttt{gen}(x,z,s_m)$ 
        \EndFor
        \State Return $\sum_{m=1}^M f(y_m)$
      \end{algorithmic}
  \end{algorithm}
\end{minipage}\\\\

\subsubsection*{Probability of Improvement (PI), Alg.~\ref{alg:pi}}
In our setting, the probability of improvement (PI) acquisition function will
return the probability that observing the system at query $x \in \mathcal{X}$
will improve upon the minimally observed objective value, $f_\text{min} =
\min_{y \in \mathcal{D}} f(y)$. We can write the exact PI acquisition function
as
\begin{align}
a_\text{PI}^*(x) = \int \mathds{1} \left\{f(y) \leq f_\text{min} \right\}
  p \left( y|\mathcal{D};x \right) dy.
\end{align}
In Alg.~\ref{alg:pi}, for a sequence of steps $m=1,\ldots,M$, we draw $z_m \sim
p(z|\mathcal{D})$, $y_m \sim p(y|z_m;x)$ via \texttt{post} and \texttt{gen},
and then compute $\lambda(y_{1:M}) = \sum_{m=1}^M \mathds{1}\left[ f(y_m) \leq
f_\text{min} \right]$.
As before, $y_m$ is drawn (marginally) from the posterior predictive
distribution. Therefore, as the number of calls $M$ to \texttt{post} and
\texttt{gen} grows,
$a_\text{PI}(x)$
$\rightarrow$ $a_\text{PI}^*(x)$ (up to a multiplicative constant) at a rate of
$O(\sqrt{M})$.

\subsubsection*{Upper Confident Bound (UCB), Alg.~\ref{alg:ucb}} We propose an
algorithm based on the principle of optimization under uncertainty (OUU), which
aims to compute a lower confidence bound for $p(f(y)|\mathcal{D};x)$, which we
denote by $\text{LCB}\left[ p(f(y)|\mathcal{D}_n;x) \right]$. In
Alg.~\ref{alg:ucb}, we use an estimate of this,
$\widehat{\text{LCB}}(f(y_m)_{m=1}^M)$.  Note that we use a lower confidence
bound since we are performing minimization, though we denote our acquisition
function with the more commonly used title UCB.
Two simples strategies for estimating this LCB are
\begin{enumerate}[topsep=5pt,itemsep=3pt,parsep=0pt,leftmargin=10mm]
\item \emph{Empirical quantiles:} Order $f(y_m)_{m=1}^M$ into $f_{(1)}\leq
\ldots \leq f_{(M)}$, and return $f_{(b)}$ if $b \in \mathbb{Z}$,
or else return $\frac{1}{2} (f_{(\floor{b})} + f_{(\floor{b} + 1)})$, where $b \in
[0,M+1]$ is a tradeoff parameter.
\item \emph{Parametric assumption:} As an example, if we model
$p(f(y)|\mathcal{D}_n;x)$ $=$ $\mathcal{N}(f(y)|\mu,\sigma^2)$, we can compute
$\hat{\mu}$ $=$ $\frac{1}{M} \sum_{m=1}^M f(y_m)$ and $\hat{\sigma}^2$ $=$
$\frac{1}{M-1} \sum_{m=1}^M (f(y_m) - \hat{\mu})^2$, and return $\hat{\mu} -
\beta \hat{\sigma}^2$, where $\beta>0$ is a trade-off parameter.
\end{enumerate}
The first proposed estimate (empirical quantiles) is a consistent estimator,
though may yield worse performance in practice than the second proposed
estimate in cases where we can approximate $p(f(y_m)|\mathcal{D};x)$ with some
parametric form.


\subsubsection*{Thompson Sampling (TS), Alg.~\ref{alg:ts}}
Thompson sampling (TS) proposes proxy values for unknown model variables by drawing a
posterior sample, and then performs optimization as if this sample were the
true model variables, and returns the result.
In Alg.~\ref{alg:ts}, we provide an acquisition function to carry out a
TS strategy in \probosp, using \texttt{post} and \texttt{gen}.
At one given iteration of BO, a specified seed is used so that each call to
$a_\text{TS}(x)$ produces the same posterior latent variable sample $\tilde{z}
\sim p(z|\mathcal{D})$ via \texttt{post}. After, \texttt{gen} is called
repeatedly to produce $y_m \sim p(y|\tilde{z};x)$ for $m=1,\ldots,M$, and the
objective values of these are averaged to yield $\lambda(y_{1:M}) =
\sum_{m=1}^M f(y_m)$.  Here, each $f(y_m) \sim
\mathbb{E}_{p(y|\tilde{z};x)} \left[f(y)\right]$.
Optimizing this acquisition function serves as a proxy for optimizing our model
given the true model variables, using posterior sample $\tilde{z}$ in place of
unknown model variables.

\subsubsection*{In Summary}
As $M \rightarrow \infty$, for constants $c_1$, $c_2$, $c_3$, and
$c_4$,
\begin{align}
  a_\text{EI}(x,\texttt{post},\texttt{gen}) &\rightarrow
    c_1 \int \mathds{1} \left\{f(y)\leq \min_{y' \in \mathcal{D}} f(y') \right\}
    \left(\min_{y' \in \mathcal{D}} f(y')-f(y)\right) p(y|\mathcal{D};x) dy \\
  a_\text{PI}(x,\texttt{post},\texttt{gen}) &\rightarrow
    c_2 \int \mathds{1} \left\{f(y)\leq \min_{y' \in \mathcal{D}} f(y') \right\}
    p(y|\mathcal{D};x) dy \\
  a_\text{UCB}(x,\texttt{post},\texttt{gen}) &\rightarrow
    c_3 \hspace{1mm} \text{LCB} \left[
    p(f(y)|\mathcal{D};x) \right] \\
  a_\text{TS}(x,\texttt{post},\texttt{gen}) &\rightarrow
    c_4 \int f(y) \hspace{1mm} p(y|\tilde{z};x) dy,
    \hspace{2mm} \text{ for } \tilde{z} \sim p(z|\mathcal{D}),
\end{align}

We visualize Alg.~\ref{alg:ei}-\ref{alg:ts} in Fig.~\ref{fig:bbacq}~(a)-(d), 
showing $M \in \{50,500\}$.



\subsection{Computational Considerations}
\label{sec:computationalconsiderations}
%
In \probo, we run a PPL's inference procedure when we call \texttt{inf}, which
has a cost dependent on the underlying inference algorithm. For example, most
MCMC methods have complexity $O(n)$ per iteration \citep{bardenet2017markov}.
However, \probosp only runs \texttt{inf} once per query; acquisition
optimization, which may be run hundreds of times per query, instead uses only
\texttt{post} and \texttt{gen}. For many PPL models, \texttt{post} and
\texttt{gen} can be implemented cheaply. For example, \texttt{post} often
involves drawing from a pool of samples or from a known distribution, and
\texttt{gen} often involves sampling from a fixed-length sequence of known
distributions and transformations, both of which typically have $O(1)$
complexity.  However, for some models, \texttt{gen} can involve running a more
costly simulation.  For these cases, we provide acquisition optimization
algorithms that use \texttt{post} and \texttt{gen} efficiently in
Sec.~\ref{sec:mf}.


\subsection{Efficient Optimization of PPL Acquisition Functions}
\label{sec:mf}

In \probo, we must optimize over the acquisition algorithms defined in the
previous section, i.e. compute $x_n = \argmin_{x \in \mathcal{X}}
a(x,\texttt{post},\texttt{gen})$.  Note that \texttt{post} and \texttt{gen} are
not in general analytically differentiable, so in contrast with
\citep{wilson2018maximizing}, we cannot optimize $a(x)$ with gradient-based
methods. We therefore explore strategies for efficient zeroth-order
optimization.

In Alg.~\ref{alg:ei}-\ref{alg:ts}, $M$ denotes the number of times
\texttt{post} and \texttt{gen} are called in an evaluation of $a(x)$.  As seen
in Fig.~\ref{fig:bbacq}, a small $M$ will return a noisy estimate of $a(x)$,
while a large $M$ will return a more-accurate estimate.  However, for some
PPLs, the \texttt{post} and/or \texttt{gen} operations can be costly (e.g. if 
\texttt{gen} involves a complex simulation \cite{toni2008approximate,
mansinghka2013approximate}), and we'd like to minimize the number of times they
are called.

This is a special case of a multi-fidelty optimization problem
\citep{forrester2007multi}, with fidelity parameter $M$. Unlike typical
multi-fidelity settings, our goal is to reduce the number of calls to
\texttt{post} and \texttt{gen} for a single $x$ only, via modifying the
acquisition function $a(x,\texttt{post},\texttt{gen})$.
This way, we can drop in any off-the-shelf optimizer that makes calls to $a$.
Suppose we have $F$ fidelities ranging from a small number of samples
$M_\text{min}$ to a large number $M_\text{max}$, i.e.  $M_\text{min} = M_1 <
\ldots < M_F = M_\text{max}$.
Intuitively, when calling $a(x,\texttt{post},\texttt{gen})$ on a given $x$,
we'd like to use a small $M$ if $a(x)$ is far from the minimal value $a(x^*)$,
and a larger $M$ if $a(x)$ is close to $a(x^*)$.

We propose the following procedure: Suppose $a_\text{min}$ is the minimum value
of $a$ seen so far during optimization (for any $x$).  For a given
fidelity $M_f$ (starting with $f$$=$$1$), we compute a lower confidence bound
(LCB) for the sampling distribution of $a(x,\texttt{post},\texttt{gen})$ with
$M_f$ calls to \texttt{post} and \texttt{gen}.
We can do this via the bootstrap method \citep{efron1992bootstrap} along with
the LCB estimates described in Sec.~\ref{sec:bbacq}. If this LCB is below
$a_\text{min}$, it remains plausible that the acquisition function minimum is
at $x$, and we repeat these steps at fidelity $M_{f+1}$. After reaching a
fidelity $f^*$ where the LCB is above $a_\text{min}$ (or upon reaching the
highest fidelity $f^* = F$), we return the estimate $a(x,\texttt{post},
\texttt{gen})$ with $M_{f^*}$ calls.  We give this procedure in
Alg.~\ref{alg:mf}.

\noindent \begin{minipage}[t]{0.46\textwidth}
  \begin{algorithm}[H]
      \caption{\hspace{1mm}$a_\text{MF} \left( x, \texttt{post}, \texttt{gen} \right)$
        }
      \label{alg:mf}
      \begin{algorithmic}[1]
        \State $a_\text{min} \leftarrow $ Min value of $a$ seen so far
        \State $\ell = -\infty$, $f=1$
        \While{$\ell \leq a_\text{min}$}
          \State $\ell \leftarrow \text{LCB-bootstrap}\left(\texttt{post},
            \texttt{gen}, M_f \right)$
          \State $f \leftarrow f+1$
        \EndWhile
        \State Return $a(x,\texttt{post},\texttt{gen})$ using $M=M_f$
      \end{algorithmic}
  \end{algorithm}
\end{minipage}
\hfill
\begin{minipage}[t]{0.49\textwidth}
  \begin{algorithm}[H]
      \caption{\hspace{1mm} LCB-bootstrap$(\texttt{post}, \texttt{gen}, M_f)$}
      \label{alg:mfinner}
      \begin{algorithmic}[1]
          \State $y_{1:M_f} \leftarrow$ Call \texttt{post} and \texttt{gen} $M_f$ times 
          \For{$j=1,\ldots,B$} 
            \State $\tilde{y}_{1:M_f} \leftarrow$ Resample$(y_{1:M_f})$ 
            \State $a_j \leftarrow \lambda(\tilde{y}_{1:M_f})$ \Comment{See text for details}
          \EndFor
          \State Return $\text{LCB}\left(a_{1:B}\right)$ 
      \end{algorithmic}
  \end{algorithm}
\end{minipage}\\\\

In Alg.~\ref{alg:mfinner} we use notation $\lambda(y_{1:M})$ to
denote the final operation (last line) in one of
Algs.~\ref{alg:ei}-\ref{alg:ts} (e.g. $\lambda_a(y_{1:M})$ $=$ $\sum_{m=1}^M
\mathds{1} [f(y_m) \leq f_\text{min}]$ in the case of PI).
As a simple example, we could run a two-fidelity algorithm, with $M \in
\{M_1, M_2\}$,
where $M_1 \ll M_2$. For a given $x$, $a_\text{MF}$  would first call
\texttt{post} and \texttt{gen} $M_1$ times, and compute the LCB with the
bootstrap. If the LCB is greater than $a_\text{min}$, it would return an $a(x,
\texttt{post}, \texttt{gen})$ with the $M_1$ calls; if not, it would return it
with $M_2$ calls. Near optima, this will make $M_1+M_2$ calls to \texttt{post}
and \texttt{gen}, and will make $M_1$ calls otherwise.

One can apply any derivative-free (zeroth-order) global optimization procedure
that iteratively calls $a_\text{MF}$. In general, we can replace the
optimization step in \probosp (Alg.~\ref{alg:probo}, line 3) with $x_n
\leftarrow \argmin_{x \in \mathcal{X}}
a_\text{MF}(x)$,
for each of the PPL acquisition functions described in Sec.~\ref{sec:bbacq}. In
Sec.~\ref{sec:mfexp}, we provide experimental results for this method, showing
favorable performance relative to high fidelity acquisition functions, as well
as reduced calls to \texttt{post} and \texttt{gen}.

\section{Examples and Experiments}
\label{sec:examples}
We provide examples of models to aid in complex optimization scenarios,
implement these models with PPLs, and show empirical results.  Our main goals
are to demonstrate that we can plug models built with various PPLs into \probo,
and use these to improve BO performance (i.e. reduce the number of iterations)
when compared with standard methods and models. We also aim to verify that our
acquisition functions and extensions (e.g.  multi-fidelity $a_\text{MF}$)
perform well in practice.

\subsubsection*{PPL Implementations:} 
We implement models with Stan \cite{carpenter2015stan} and Edward
\cite{tran2016edward}, which (respectively) make use of the No U-Turn Sampler
\cite{hoffman2011no} (a form of Hamiltonian Monte Carlo) and black box
variational inference \cite{ranganath2014black}. We also use George
\cite{hodlr} and GPy \cite{gpy2014} for GP comparisons.

\subsection{BO with State Observations}
\label{sec:exstate}

\subsubsection*{Setting:}
Some systems exhibit unique behavior in different regions of the input space
$\mathcal{X}$ based on some underlying state.  We often do not know these state
regions apriori, but can observe the state of an $x$ when it is queried.
%
%
%
%
%
%
%
Two examples of this, for computational systems, are:
\begin{itemize}[topsep=5pt,itemsep=5pt,parsep=0pt,leftmargin=10mm]
  \item Timeouts or failures: there may be regions where queries fail or time
    out. We can observe if a query has a ``pass'' or ``fail'' state 
    \citep{gardner2014bayesian, gelbart2014bayesian, lam2017lookahead}.
  \item Resource usage regions: queries can have distinct resource usage
    patterns. We can observe this pattern for a query, and use it to assign
    a state \citep{alipourfard2017cherrypick, dalibard2017boat}.
\end{itemize}
%
Assume that for each query $x \in \mathcal{X}$, $s(x)$ returns a $y \in
\mathcal{Y} = \mathbb{R} \times \mathbb{Z}^+$,  with two types of information:
an objective value $y_0$ and an state observation $y_1$ indicating the region
assignment. We take the objective function to be $f(y) = y_0$.

\subsubsection*{Model:}
Instead of using a single black box model for the entire input space
$\mathcal{X}$ we provide a model that infers the distinct regions and learns a
model for each. For the case of two states, we can write the generative model
for this as: $c \sim \text{Bernoulli}(\cdot|C(x))$, $y \sim c M_1(\cdot|x) + (1-c)
M_2(\cdot|x)$, where $M_1$ and $M_2$ are models for $y|x$ (e.g. GP regression
models) and $C$ is a classification model (e.g. a Bayesian NN) that models the
probability of $M_1$.  We refer to this model as a \emph{switching model}.
This model could be extended to more states. We show inference in this model in
Fig.~\ref{fig:stateobs}~(d).  Comparing this with a a GP
(Fig.~\ref{fig:stateobs}~(c)), we see that GP hyperparameter estimates can be
negatively impacted due to the nonsmooth landscape around region boundaries.

\subsubsection*{Empirical Results:}
We demonstrate the switching model on the task of neural network architecture
and hyperparameter search \cite{zoph2016neural, kandasamy2018neural} with
timeouts, where in each query we train a network and return accuracy on a held
out validation set. However, training must finish within a given time
threshold, or it \emph{times out}, and instead returns a preset
(low accuracy) value.
%
We optimize over multi-layer perceptron (MLP) neural networks, where we
represent each query as a vector $x \in \mathbb{R}^4$, where $x$ = $($number of
layers, layer width, learning rate, and batch size$)$.  We train and validate
each network on the  Pima Indians Diabetes Dataset \cite{smith1988using}.
Whenever training has not converged within 60 seconds, the system times out and
returns a fixed accuracy value of $30\%$.  We use a GP regression model for
$M_1$, and a Gaussian model (with mean and variance latent variables) for
$M_2$.  We compare \probosp with this switching model against standard BO using
GPs, plotting the maximum validation accuracy found vs iteration $n$, averaged
over 10 trials, in Fig.~\ref{fig:stateobs}~(e)-(f).


\setlength{\minipagewidth}{0.4\textwidth}
\setlength{\minipageoffset}{0mm}
\setlength{\columngap}{2mm}
\setlength{\pretitlegap}{5mm}
\setlength{\posttitlegap}{3mm}

\begin{figure}[bp]
  \centering

  \begin{minipage}[b]{\textwidth}
    \centering
    \vspace{\pretitlegap}
    \underline{\emph{BO with State Observations}}
    \vspace{\posttitlegap}
  \end{minipage}\\

  \begin{minipage}[b]{\minipagewidth}
    \hspace{\minipageoffset}
    \centering
    {\small (a) Data and true system mean}
  \end{minipage}
  \hspace{\columngap}
  \begin{minipage}[b]{\minipagewidth}
    \hspace{\minipageoffset}
    \centering
    {\small (b) GP (on state 1 data only)}
  \end{minipage}\\

  \begin{minipage}[b]{\minipagewidth}
    \includegraphics[width=\textwidth]{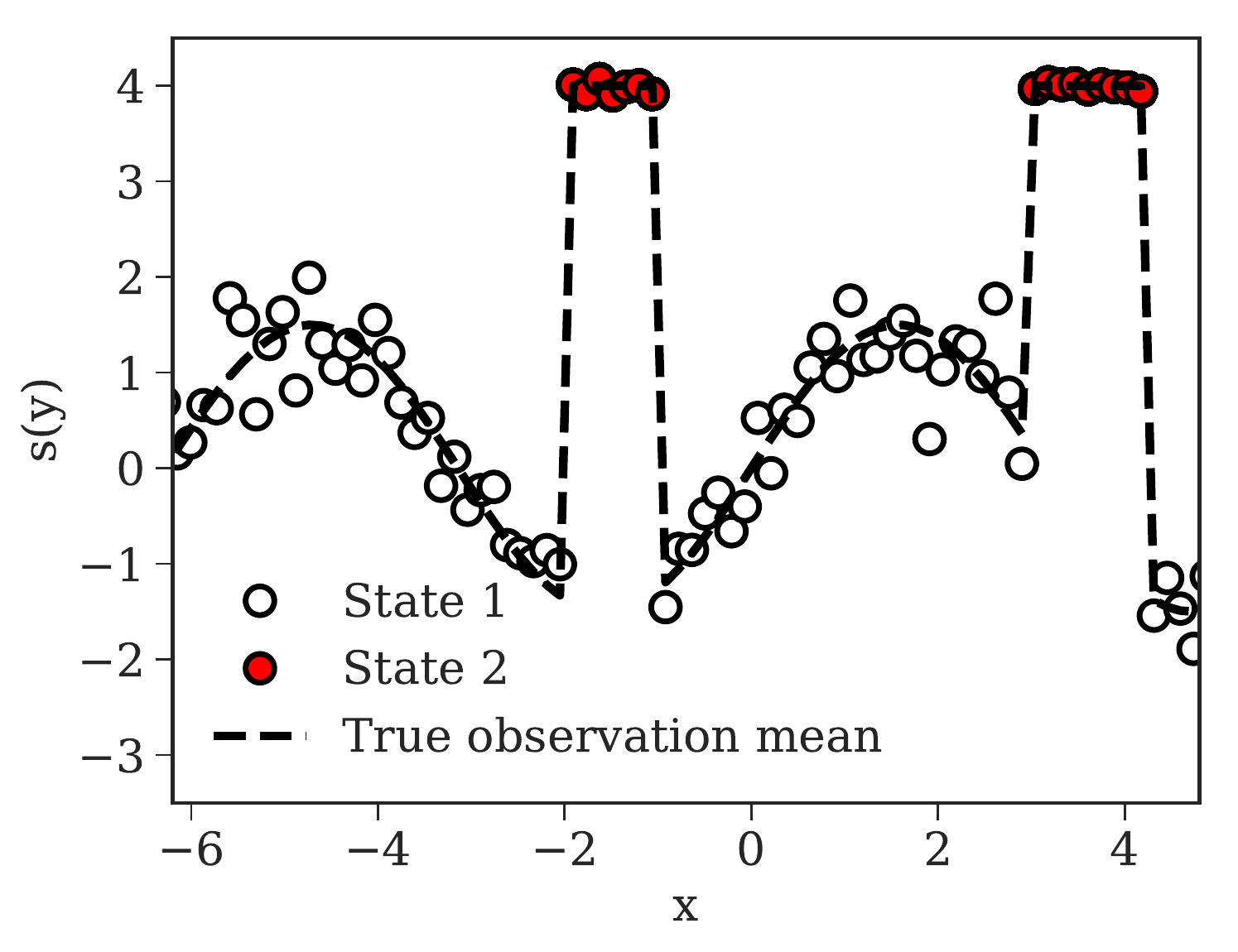}
  \end{minipage}
  \hspace{\columngap}
  \begin{minipage}[b]{\minipagewidth}
    \includegraphics[width=\textwidth]{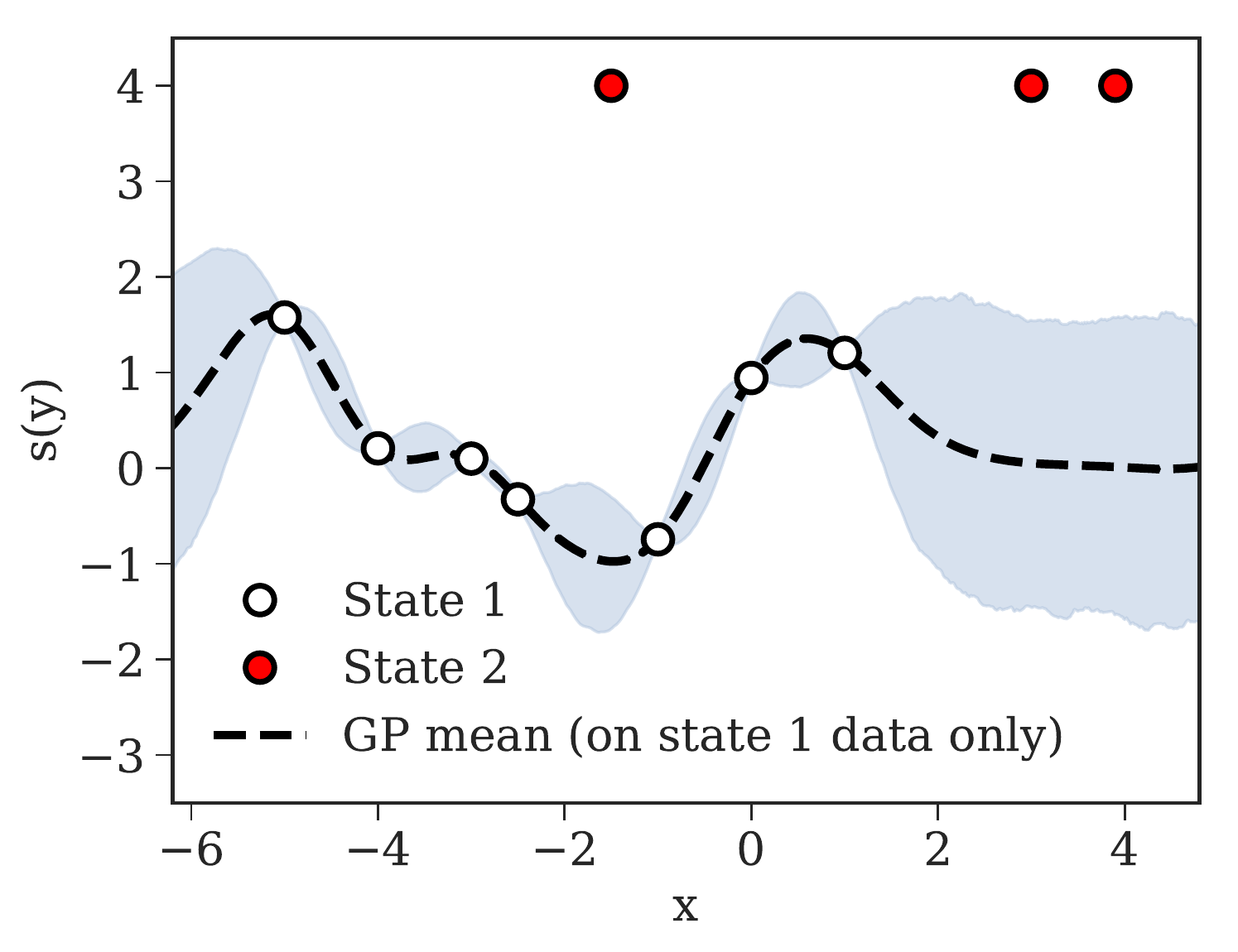}
  \end{minipage}\\

  \begin{minipage}[b]{\minipagewidth}
    \hspace{\minipageoffset}
    \centering
    {\small (c) GP (on all data)}
  \end{minipage}
  \hspace{\columngap}
  \begin{minipage}[b]{\minipagewidth}
    \hspace{\minipageoffset}
    \centering
    {\small (d) Switching model (on all data)}
  \end{minipage}\\

  \begin{minipage}[b]{\minipagewidth}
    \includegraphics[width=\textwidth]{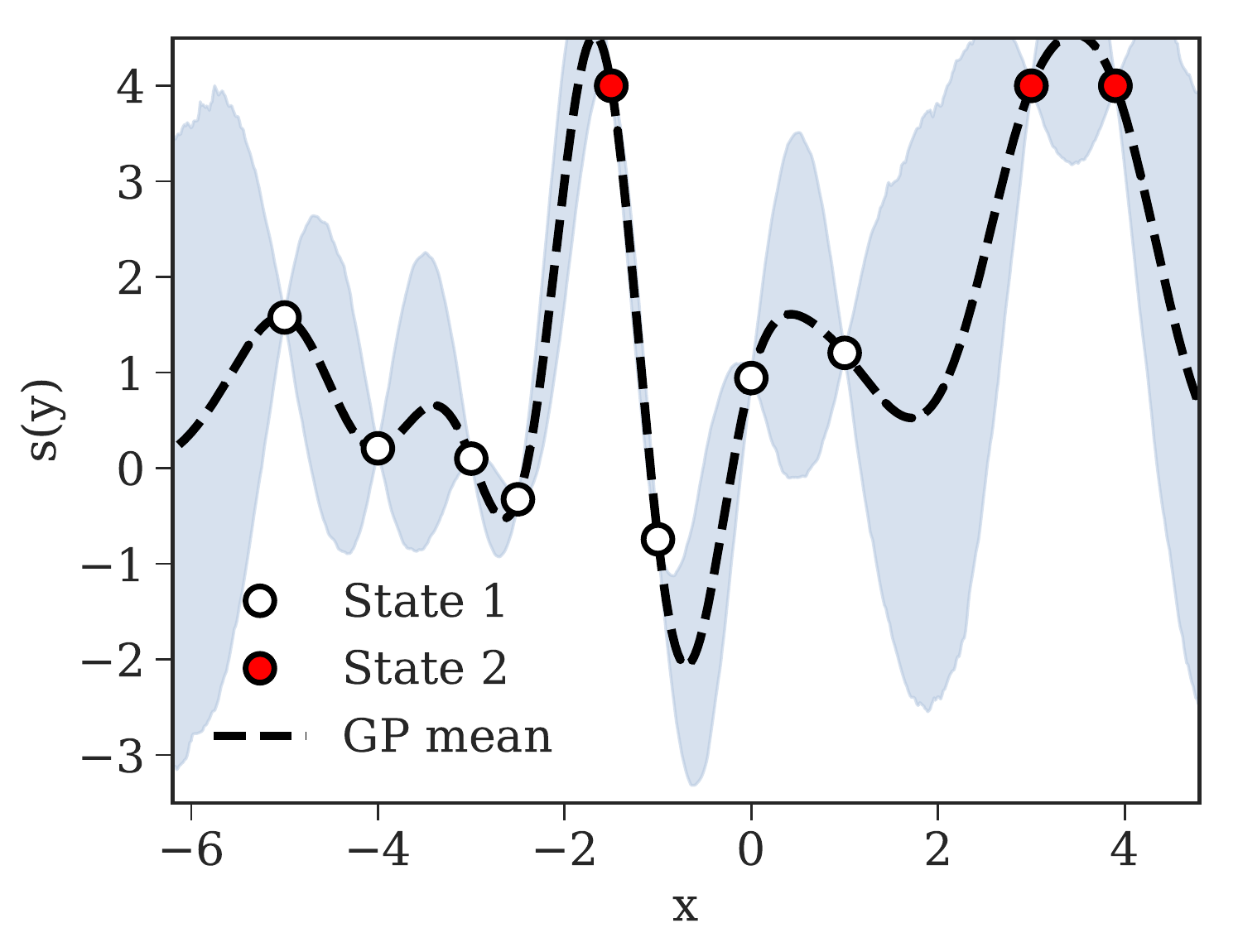}
  \end{minipage}
  \hspace{\columngap}
  \begin{minipage}[b]{\minipagewidth}
    \includegraphics[width=\textwidth]{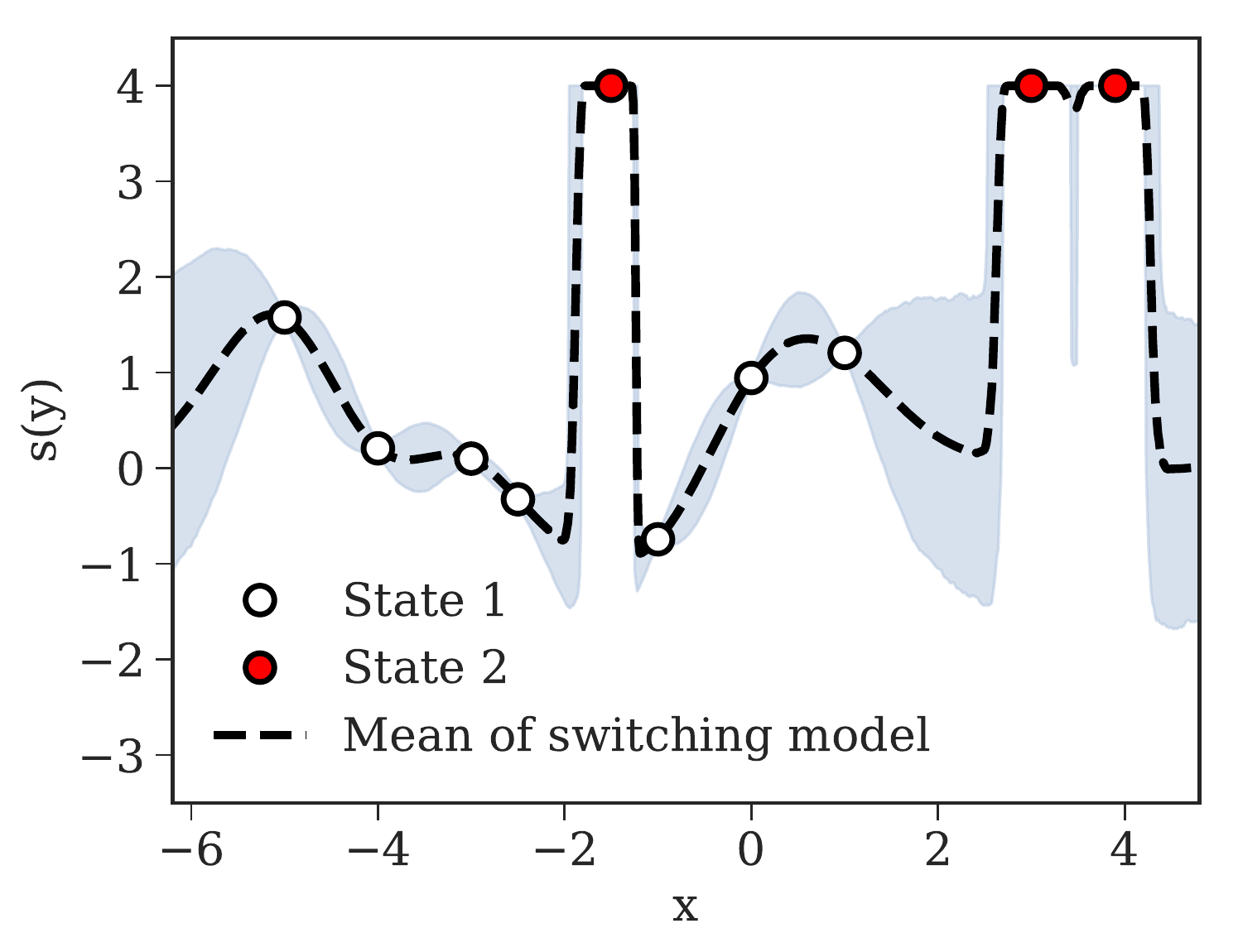}
  \end{minipage}\\

  \begin{minipage}[b]{\minipagewidth}
    \hspace{\minipageoffset}
    \centering
    {\small (e) EI and UCB}
  \end{minipage}
  \hspace{\columngap}
  \begin{minipage}[b]{\minipagewidth}
    \hspace{\minipageoffset}
    \centering
    {\small (f) PI and TS}
  \end{minipage}\\

  \begin{minipage}[b]{\minipagewidth}
    \includegraphics[width=\textwidth]{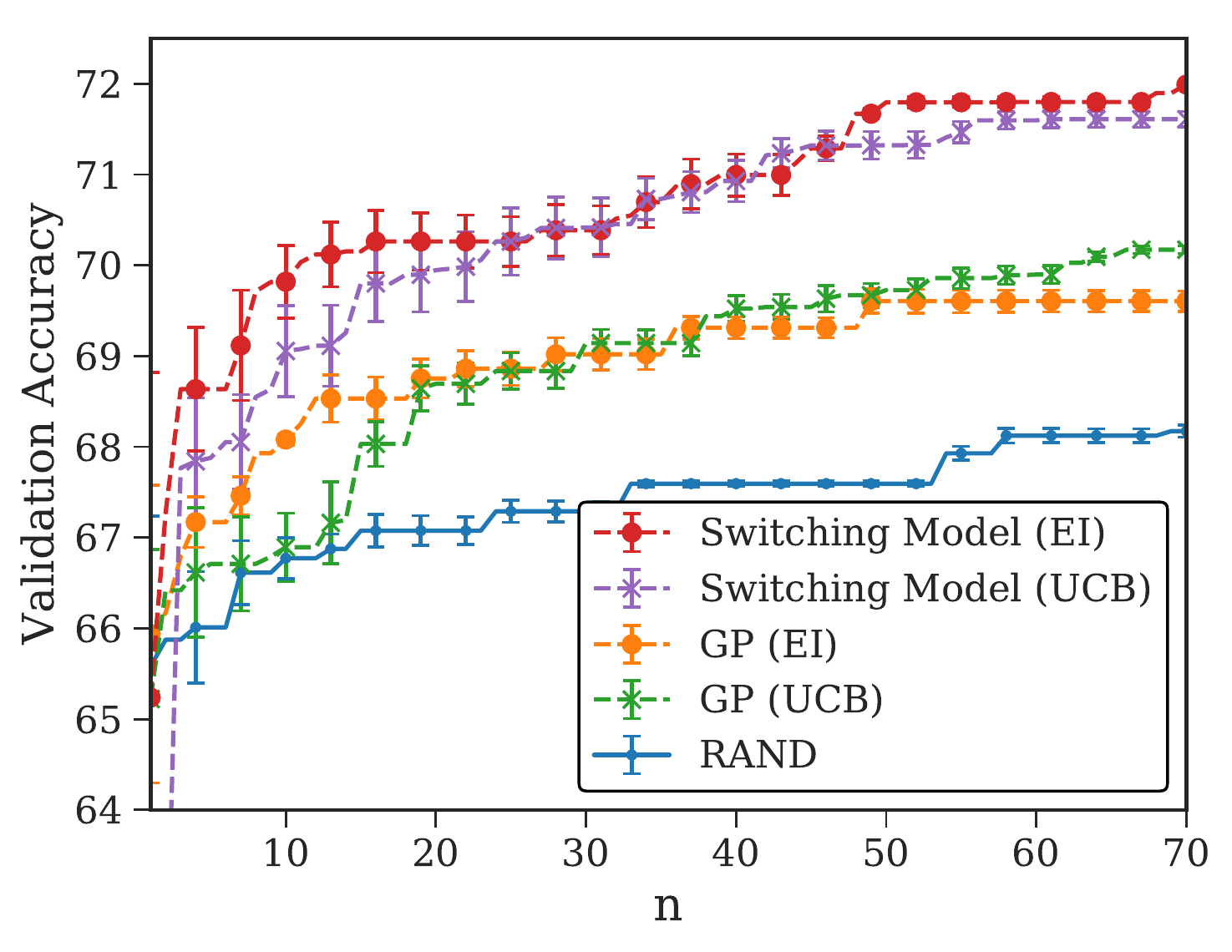}
  \end{minipage}
  \hspace{\columngap}
  \begin{minipage}[b]{\minipagewidth}
    \includegraphics[width=\textwidth]{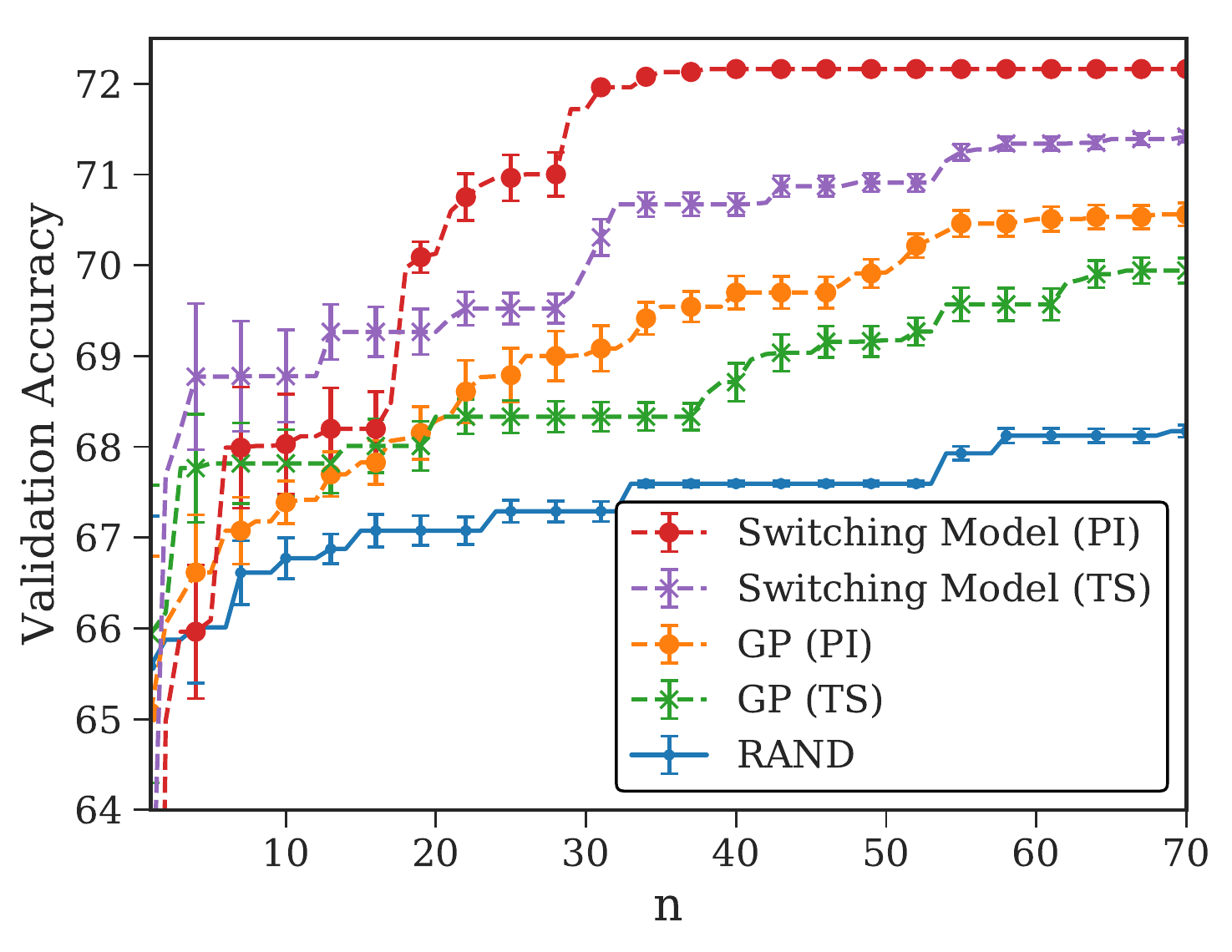}
  \end{minipage}\\

  \caption{
    BO with state observations (Sec.~\ref{sec:exstate}). We show (a) the true
    system, and inference results on (b) a GP model fit on state 1 data only,
    (c) the same GP model fit on all data, where hyperparameter estimates are
    badly influenced, and (d) our switching model fit on all data.
    In (e)-(f) we show results on the task of neural network architecture and
    hyperparameter search with timeouts, comparing \probosp using a switching
    model to BO using GPs.
    Curves are averaged over 10 trials, and error bars represent one standard
    error.
    \label{fig:stateobs}
  }
\end{figure}


\setlength{\minipagewidth}{0.4\textwidth}
\setlength{\minipageoffset}{0mm}
\setlength{\columngap}{1mm}
\setlength{\pretitlegap}{0mm}
\setlength{\posttitlegap}{1mm}

\begin{figure}[bp]
  \centering

  \begin{minipage}[b]{\textwidth}
    \centering
    \vspace{\pretitlegap}
    \underline{\emph{Contaminated BO}}
    \vspace{\posttitlegap}
  \end{minipage}\\

  \begin{minipage}[b]{\minipagewidth}
    \hspace{\minipageoffset}
    \centering
    {\small (a)  GP ($n=20$)}
  \end{minipage}
  \hspace{\columngap}
  \begin{minipage}[b]{\minipagewidth}
    \hspace{\minipageoffset}
    \centering
    {\small (b)  GP ($n=50$)}
  \end{minipage}\\

  \begin{minipage}[b]{\minipagewidth}
    \includegraphics[width=\textwidth]{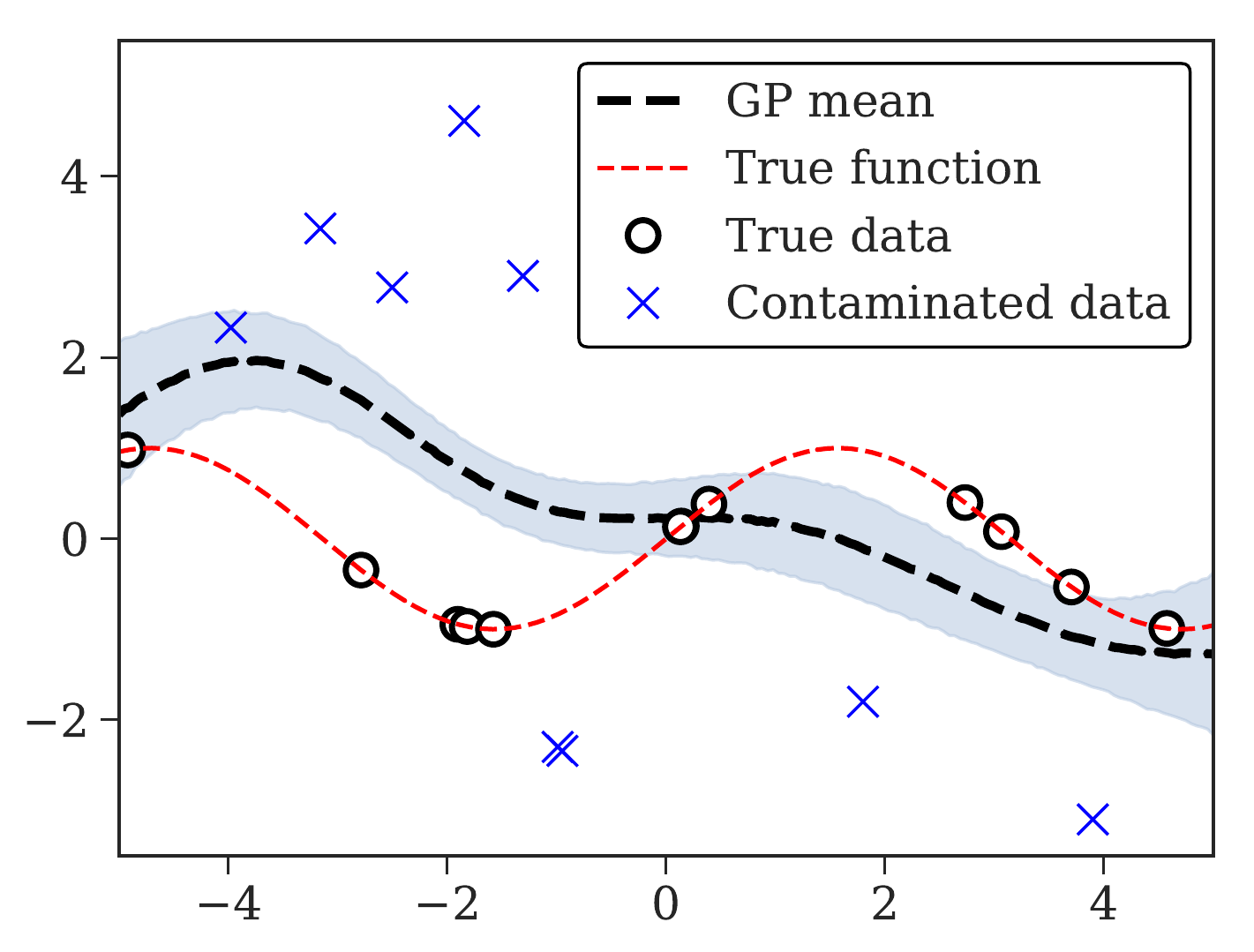}
  \end{minipage}
  \hspace{\columngap}
  \begin{minipage}[b]{\minipagewidth}
    \includegraphics[width=\textwidth]{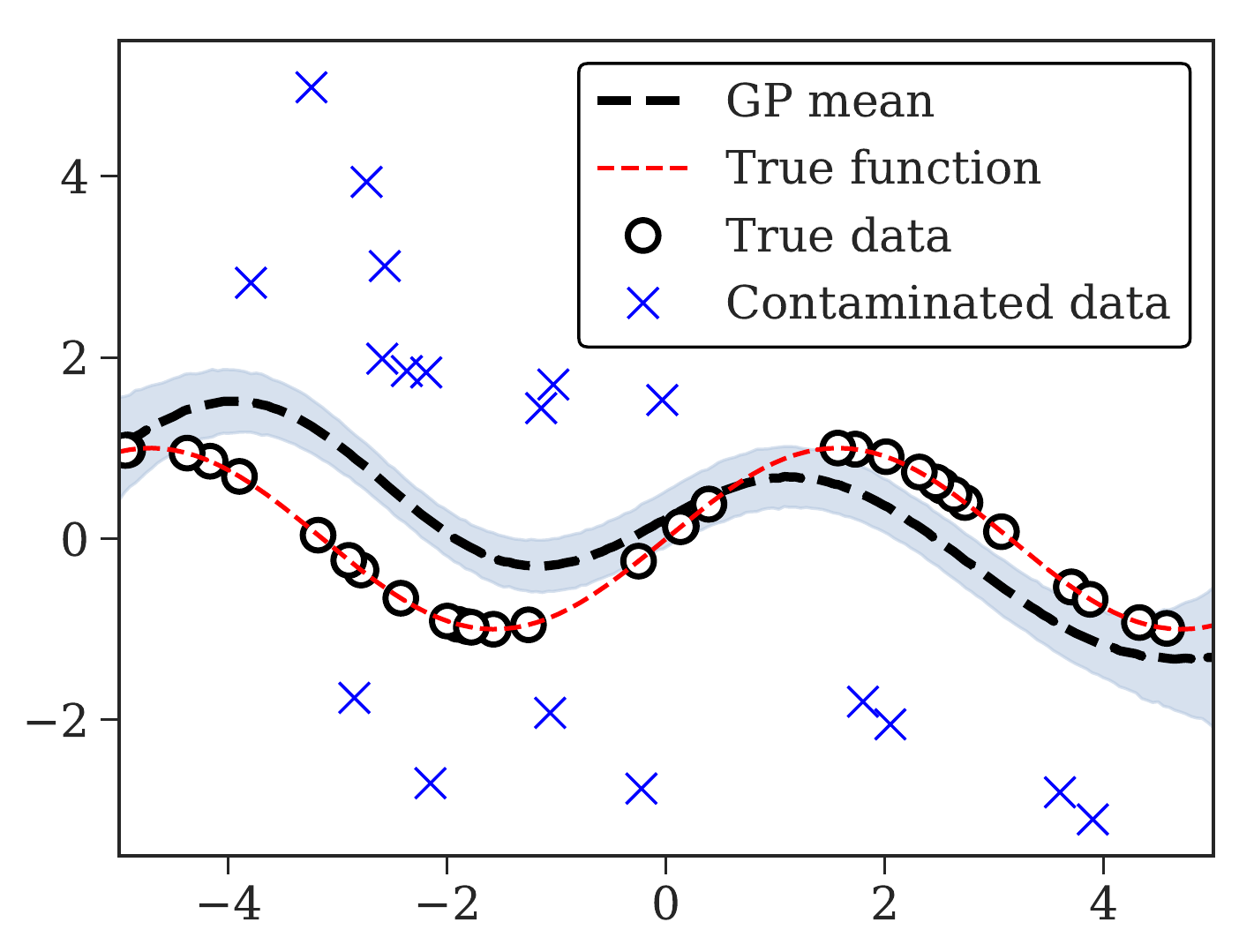}
  \end{minipage}\\

  \begin{minipage}[b]{\minipagewidth}
    \hspace{\minipageoffset}
    \centering
    {\small (c) Denoising GP ($n=20$)}
  \end{minipage}
  \hspace{\columngap}
  \begin{minipage}[b]{\minipagewidth}
    \hspace{\minipageoffset}
    \centering
    {\small (d) Denoising GP ($n=50$)}
  \end{minipage}\\

  \begin{minipage}[b]{\minipagewidth}
    \includegraphics[width=\textwidth]{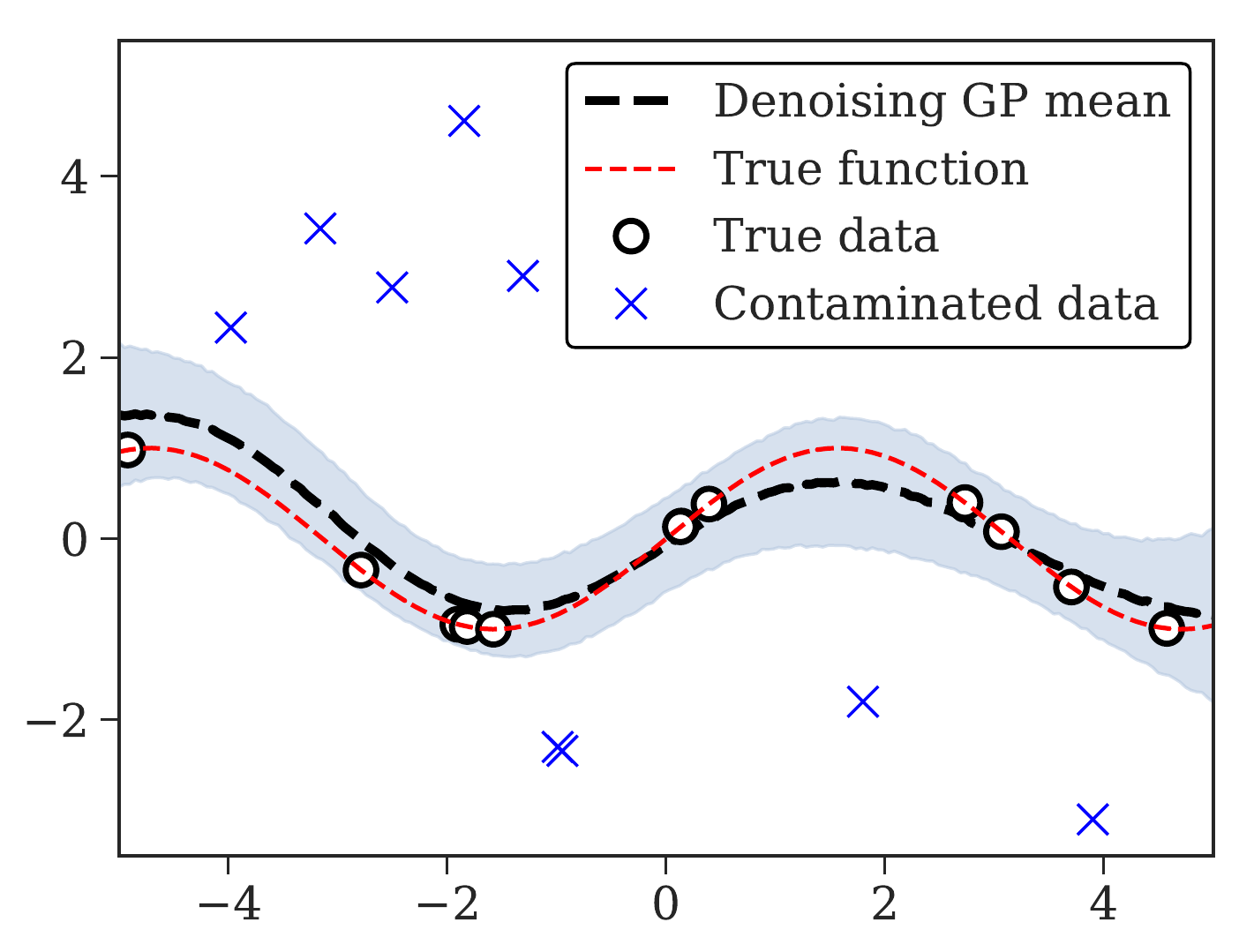}
  \end{minipage}
  \hspace{\columngap}
  \begin{minipage}[b]{\minipagewidth}
    \includegraphics[width=\textwidth]{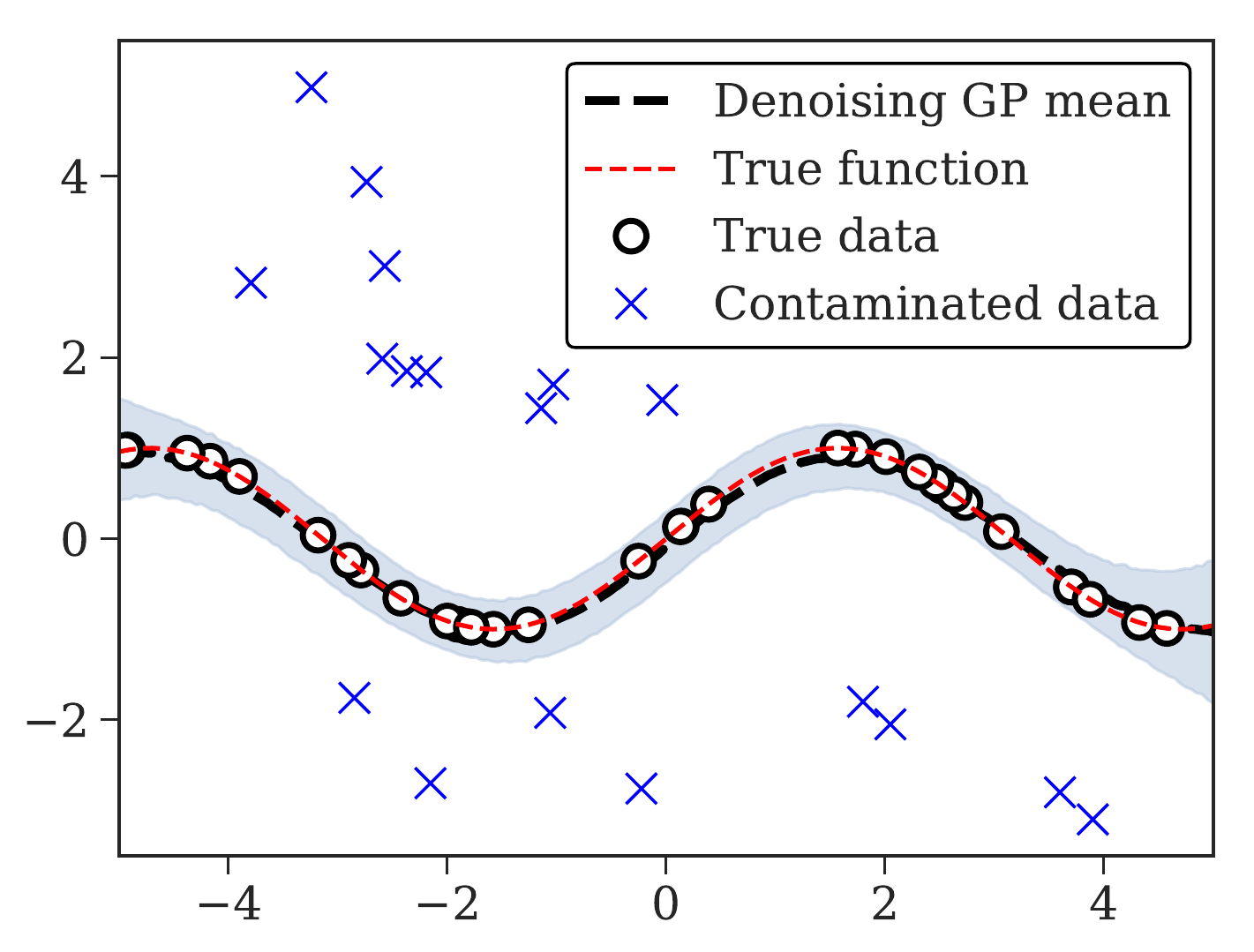}
  \end{minipage}\\

  \begin{minipage}[b]{\minipagewidth}
    \hspace{\minipageoffset}
    \centering
    {\small (e) EI and UCB ($p=.01$)}
  \end{minipage}
  \hspace{\columngap}
  \begin{minipage}[b]{\minipagewidth}
    \hspace{\minipageoffset}
    \centering
    {\small (f) PI and TS ($p=.01$)}
  \end{minipage}\\

  \begin{minipage}[b]{\minipagewidth}
    \includegraphics[width=\textwidth]{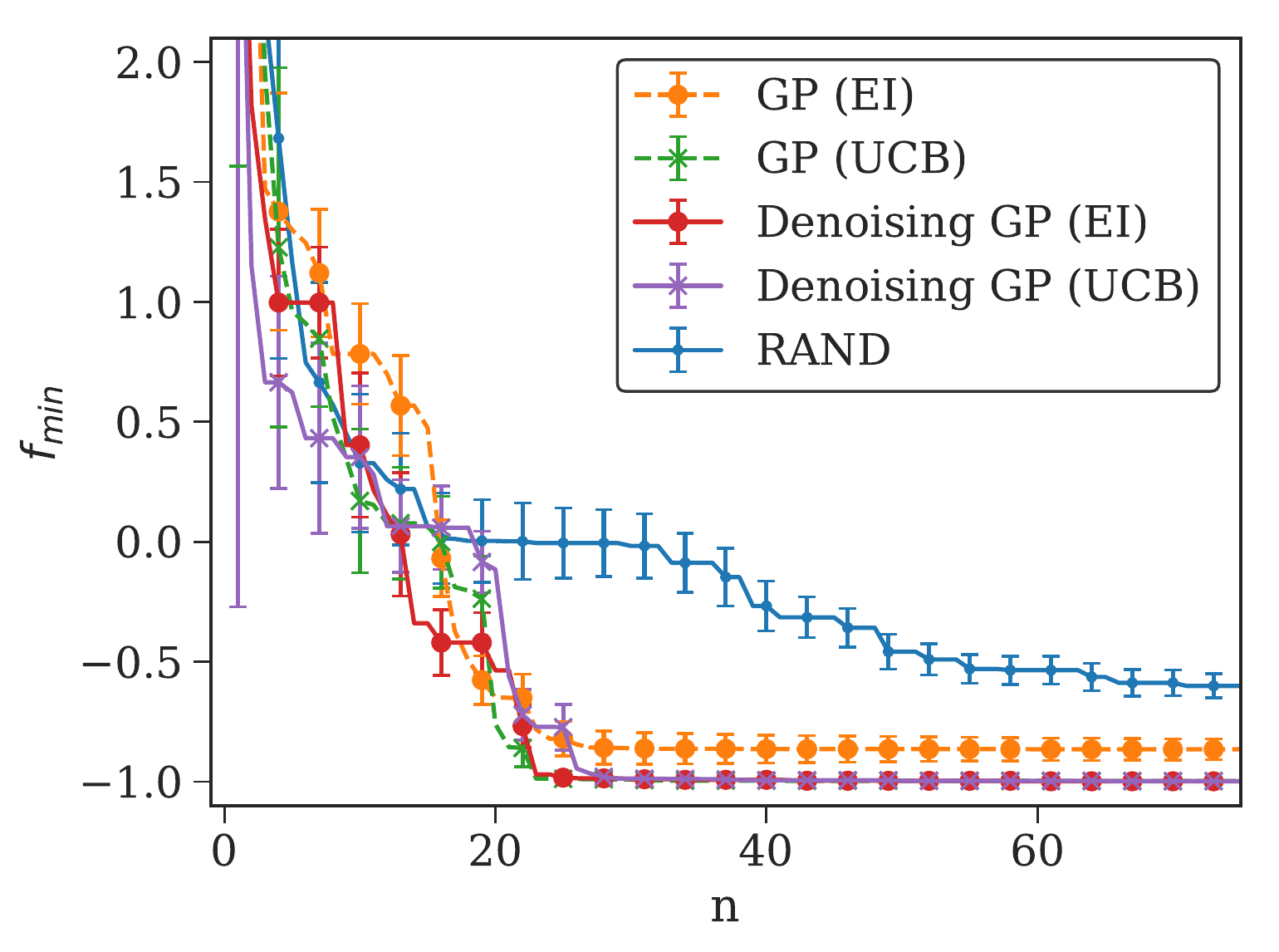}
  \end{minipage}
  \hspace{\columngap}
  \begin{minipage}[b]{\minipagewidth}
    \includegraphics[width=\textwidth]{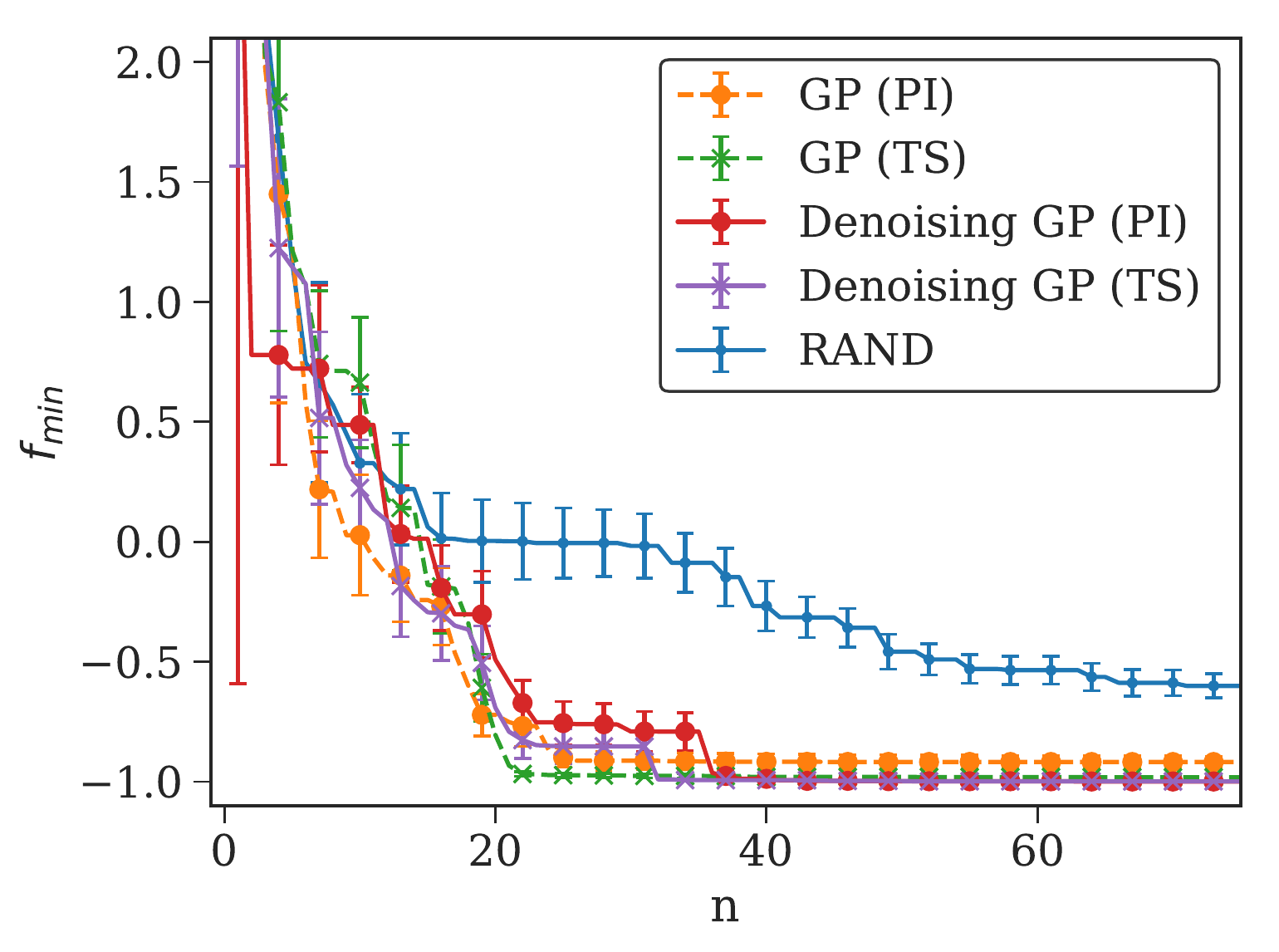}
  \end{minipage}\\

  \begin{minipage}[b]{\minipagewidth}
    \hspace{\minipageoffset}
    \centering
    {\small (g) EI and UCB ($p=.33$)}
  \end{minipage}
  \hspace{\columngap}
  \begin{minipage}[b]{\minipagewidth}
    \hspace{\minipageoffset}
    \centering
    {\small (h) PI and TS ($p=.33$)}
  \end{minipage}\\

  \begin{minipage}[b]{\minipagewidth}
    \includegraphics[width=\textwidth]{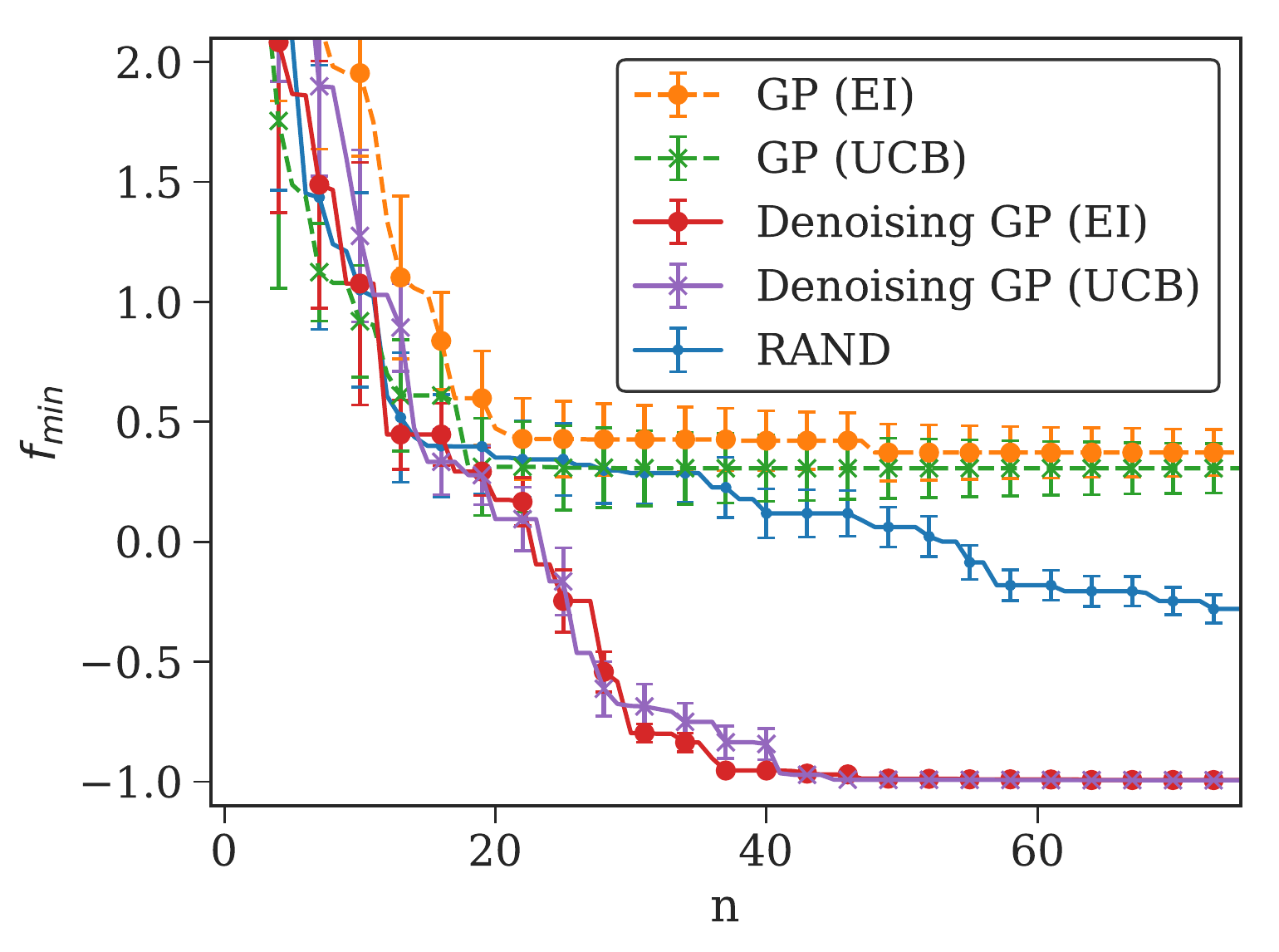}
  \end{minipage}
  \hspace{\columngap}
  \begin{minipage}[b]{\minipagewidth}
    \includegraphics[width=\textwidth]{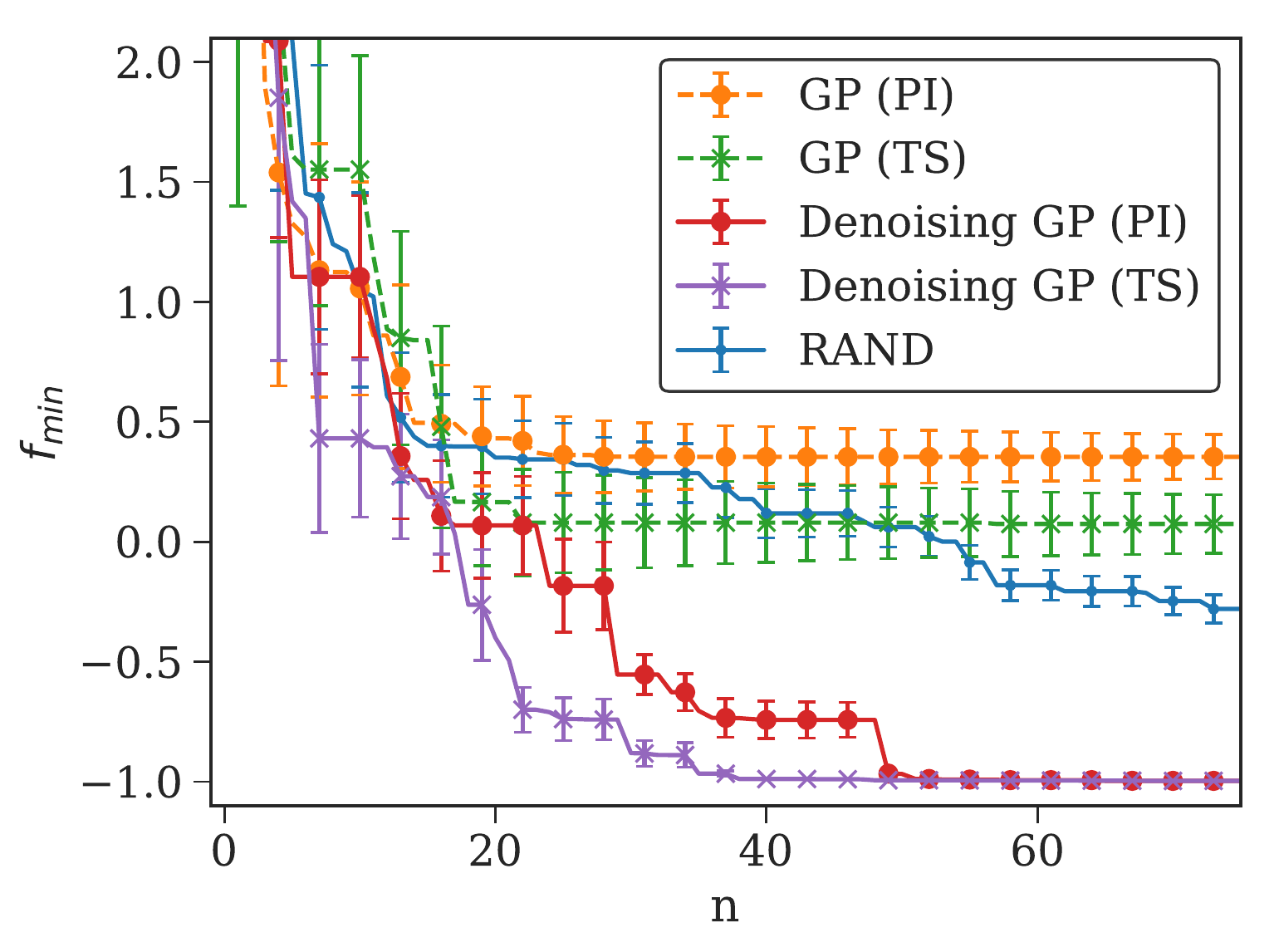}
  \end{minipage}

  \vspace{-3mm}
  \caption{
    Contaminated BO (Sec.~\ref{sec:excontam}).
    We show (a) inference in a GP with n=20 and (b) n=50, and (c)-(d) inference
    in a denoising GP.
    In (e)-(f), for low corruption ($p=.01$), \probosp using denoising GPs is
    competitive with standard BO using GP models. In (g)-(h), for higher
    corruption ($p=.33$), \probosp converges to the optimal value while
    standard BO does not.
    Curves are averaged over 10 trials, and error bars represent one standard
    error.
    \label{fig:contambo}
  }
\end{figure}


\setlength{\minipagewidth}{0.4\textwidth}
\setlength{\minipageoffset}{0mm}
\setlength{\columngap}{2mm}
\setlength{\pretitlegap}{2mm}
\setlength{\posttitlegap}{3mm}

\begin{figure}[bp]
  \centering

  \begin{minipage}[b]{\textwidth}
    \centering
    \vspace{\pretitlegap}
    \underline{\emph{BO with Prior Structure on the Objective Function}}
    \vspace{\posttitlegap}
  \end{minipage}\\
  \vspace{-1mm}

  \begin{minipage}[b]{\minipagewidth}
    \hspace{\minipageoffset}
    \centering
    {\small (a) GP}
  \end{minipage}
  \hspace{\columngap}
  \begin{minipage}[b]{\minipagewidth}
    \hspace{\minipageoffset}
    \centering
    {\small (b) Basin model}
  \end{minipage}\\

  \begin{minipage}[b]{\minipagewidth}
    \includegraphics[width=\textwidth]{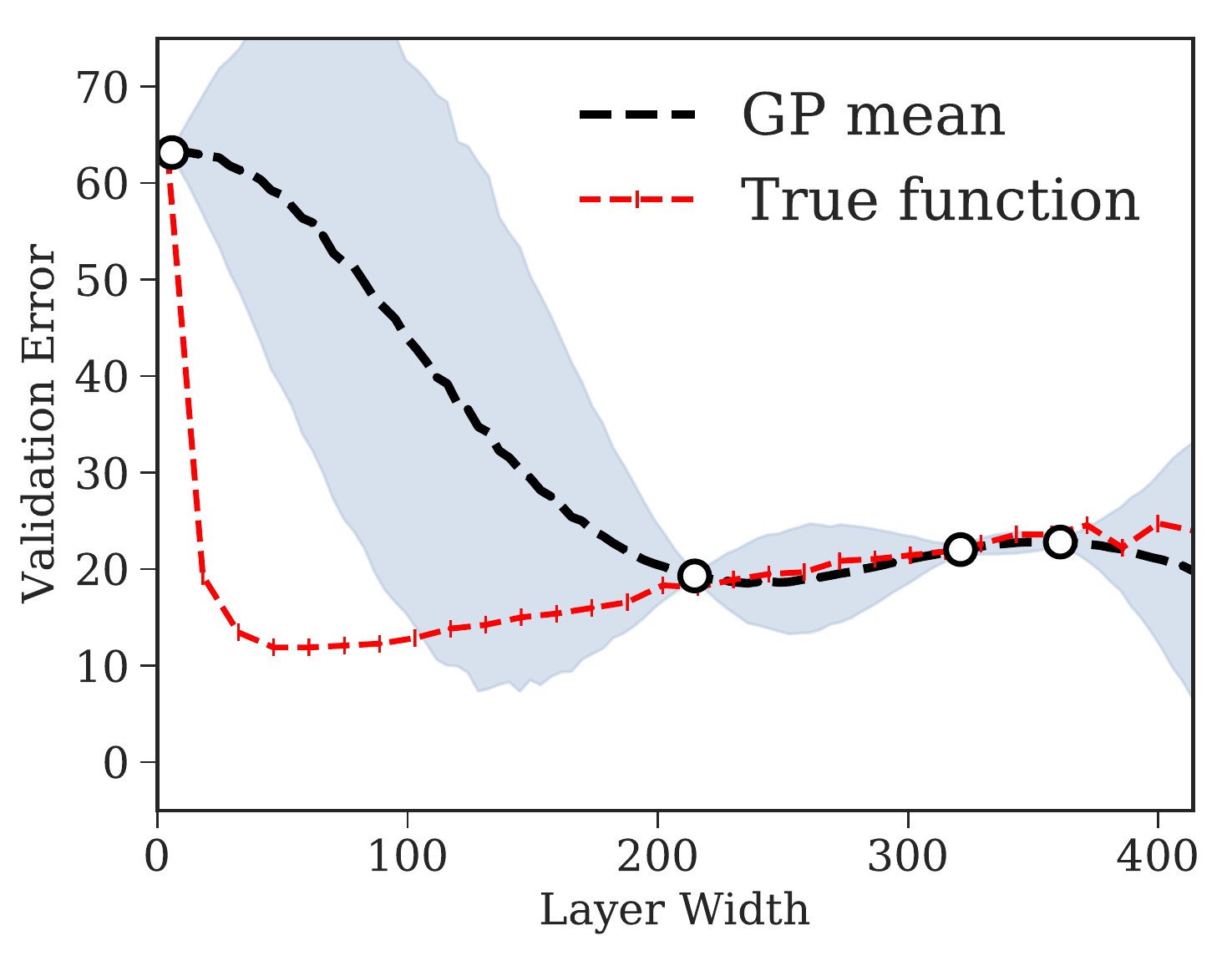}
  \end{minipage}
  \hspace{\columngap}
  \begin{minipage}[b]{\minipagewidth}
    \includegraphics[width=\textwidth]{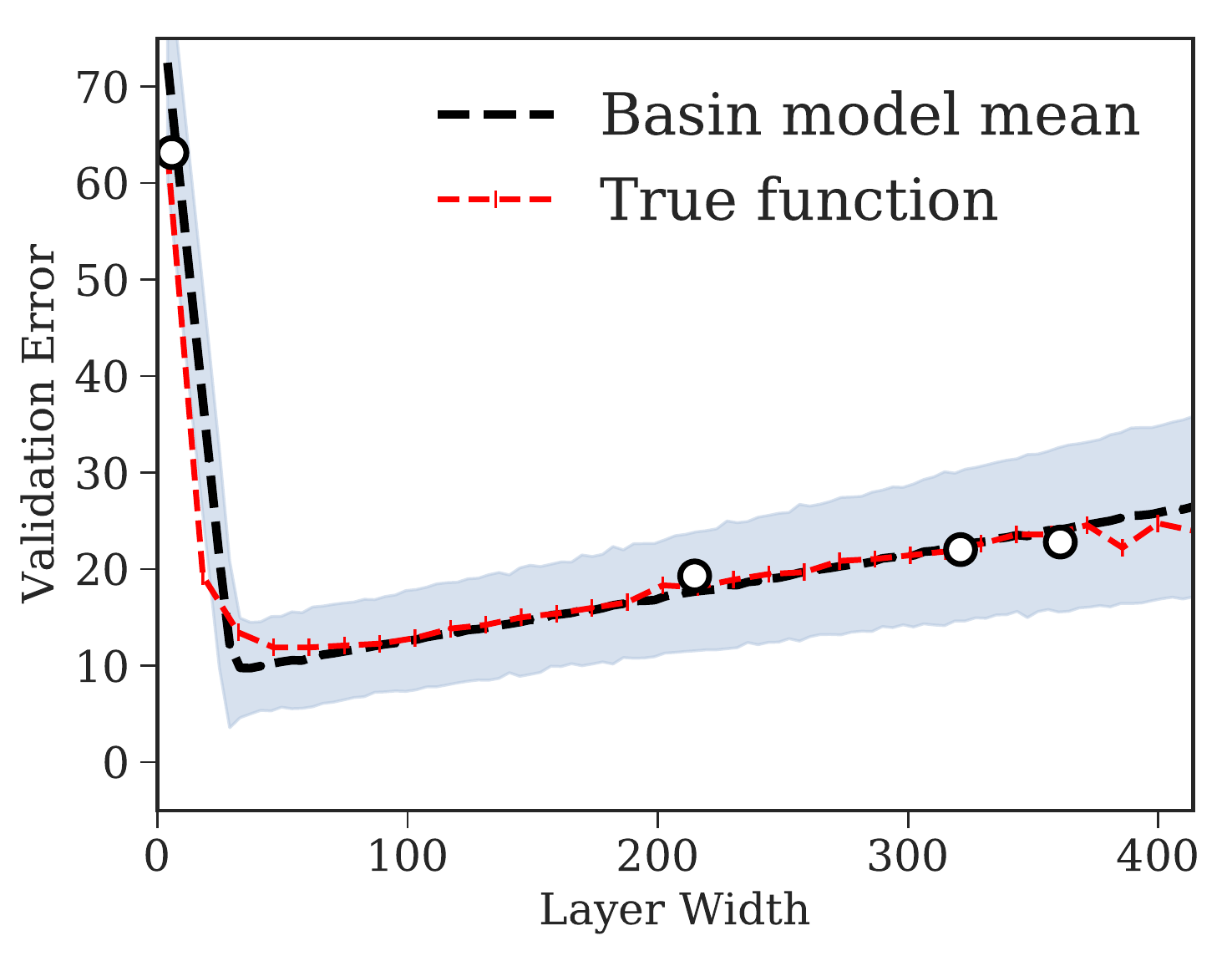}
  \end{minipage}\\

  \begin{minipage}[b]{\minipagewidth}
    \hspace{\minipageoffset}
    \centering
    {\small (c) EI and UCB}
  \end{minipage}
  \hspace{\columngap}
  \begin{minipage}[b]{\minipagewidth}
    \hspace{\minipageoffset}
    \centering
    {\small (d) PI and TS}
  \end{minipage}\\

  \begin{minipage}[b]{\minipagewidth}
    \includegraphics[width=\textwidth]{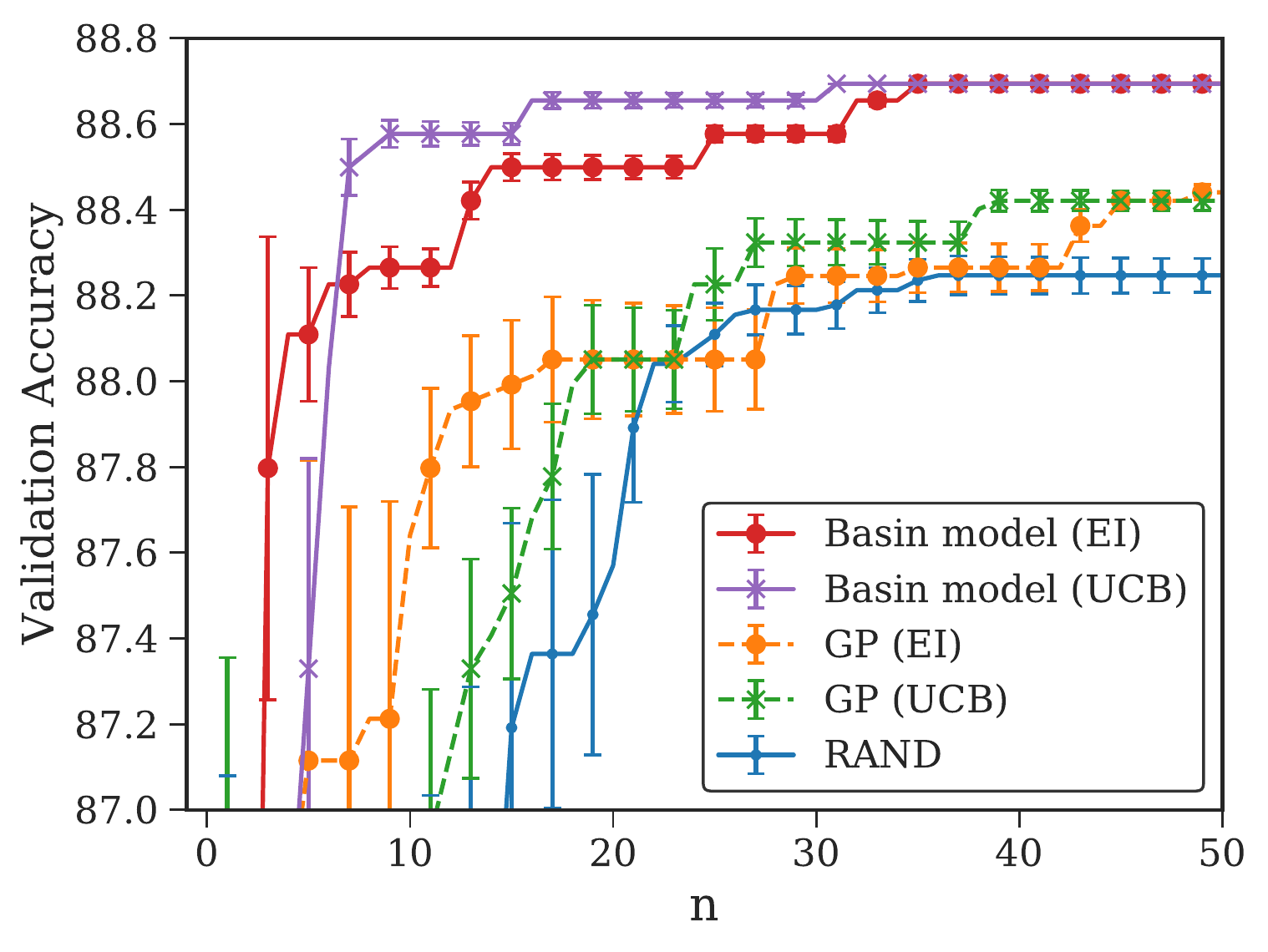}
  \end{minipage}
  \hspace{\columngap}
  \begin{minipage}[b]{\minipagewidth}
    \includegraphics[width=\textwidth]{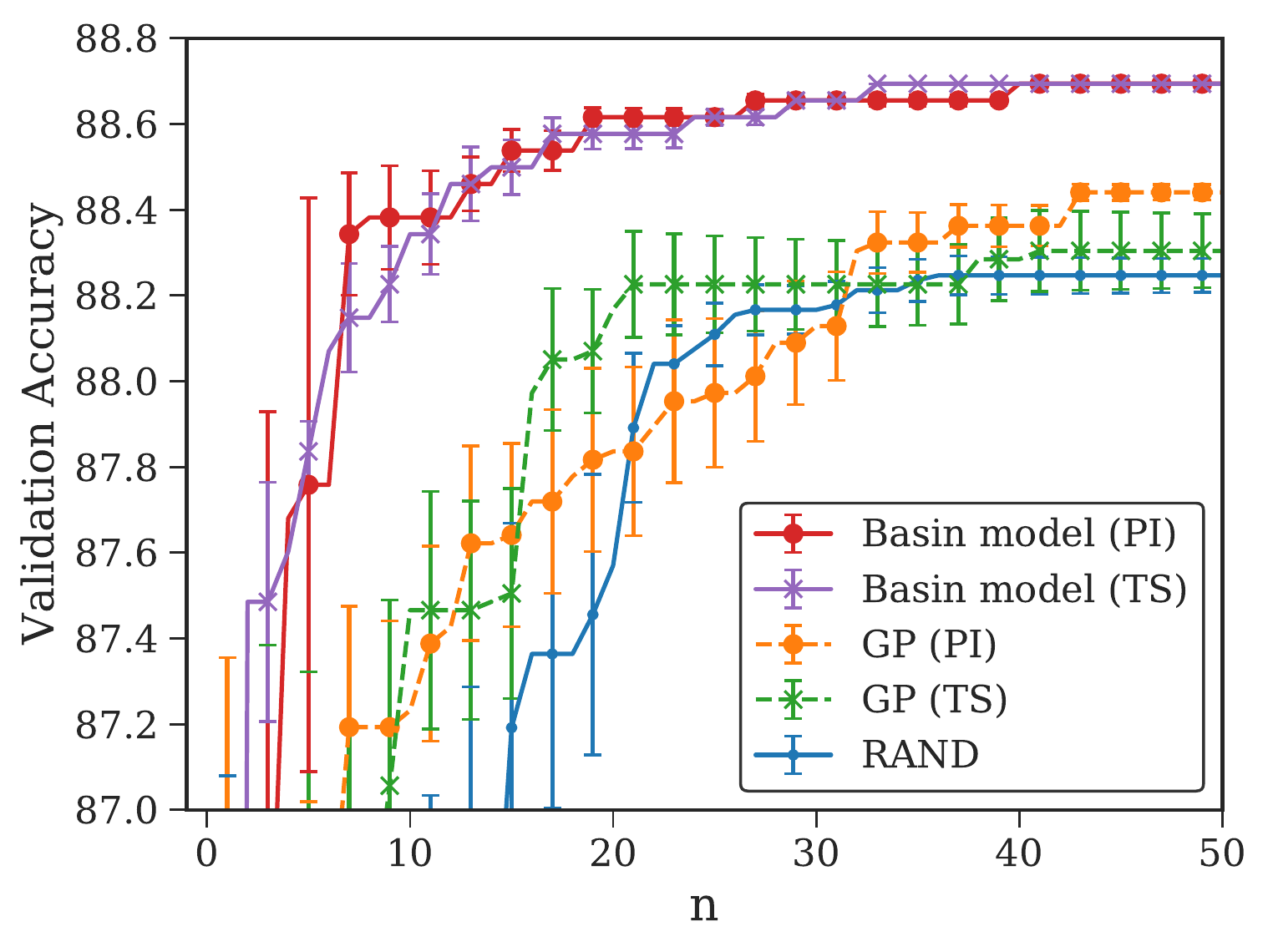}
  \end{minipage}\\

  \vspace{-2mm}
  \caption{
    Basin model for overfitting (Sec.~\ref{sec:nas}). We plot validation
    accuracy vs layer width for a small dataset, and show inference in (a) a GP
    and (b) our basin model.  In (c-d) we show results of model complexity
    hyperparameter tuning experiments, comparing \probosp using a basin model
    with BO using GPs.
    Curves are averaged over 10 trials, and error bars represent one standard
    error.
    \label{fig:priorstruc}
  }
\end{figure}

\subsection{Robust Models for Contaminated BO}
\label{sec:excontam}
\subsubsection*{Setting:}
We may want to optimize a system that periodically yields ``contaminated
observations,'' i.e. outliers drawn from a second noise distribution.  Examples
of this are queries involving unstable simulations \cite{lucas2013failure}, or faulty
computer systems \cite{schroeder2009large}.  This is similar to the setting of Huber's
$\epsilon$-contamination model \cite{huber1992robust}, and we refer to this as
\emph{contaminated BO}.  The contaminating distribution may have some
dependence on input $\mathcal{X}$ (e.g. may be more prevalent in a window
around the optimum value $x^*$). Note that this differs from
Sec.~\ref{sec:exstate} because we do not have access to state observations, and
the noise distributions are not in exclusive regions of $\mathcal{X}$. To
perform accurate BO in this setting, we need models that are robust to the
contamination noise.

\subsubsection*{Model:}
We develop a \emph{denoising model}, which infers (and ignores) contaminated
data points.
Given a system model $M_s$ and contamination model $M_c$ we write our denoising
model as $y \sim w_s M_s(\cdot|z_s;x) + w_c M_c(\cdot|z_c;x)$, where $z_s, z_c
\sim \text{Prior}(\cdot)$, and $w_s, w_c \sim \text{Prior}(\cdot|x)$ (and where
$\text{Prior}$ denotes some appropriate prior density).  This is a mixture
where weights $(w_s,w_c)$ can depend on input $x$.  We show inference in this
model in Fig.~\ref{fig:contambo}~(c)-(d).
%

\subsubsection*{Empirical Results:}
We show experimental results for \probosp on a synthetic optimization task.
This allows us to know the true optimal value $x^*$ and objective $f(s(x^*))$,
which may be difficult to judge in real settings (given the contaminations),
and to show results under different contamination levels. For an $x \in
\mathbb{R}^d$, with probability $1-p$ we query the function $f(x)$ $=$ $
\norm{x}_2-\frac{1}{d}\sum_{i=1}^d \text{cos}(x_i)$, which has a minimum value
of $f(x^*)=-1$ at $x^* = 0_d$, and with probability $p$, we receive a
contaminated value with distribution $f(x)$ $\sim$
$\text{Unif}([f_\text{max}/10,f_\text{max}])$, where $f_\text{max}$ is
$\max_{x\in \mathcal{X}} f(x)$.
%
We compare \probosp using a \emph{denoising GP} model with standard BO using GPs. We
show results for both a low contamination setting $(p=.01)$ and a high
contamination setting $(p=.33)$, in Fig.~\ref{fig:contambo}~(e)-(h), where we
plot the minimal found value (under the noncontaminated model) $f_{\min}$ $=$
$\min_{t\leq n} f(y_t)$ vs iteration $n$, averaged over 10 trials. In the low
contamination setting, both models converge to a near-optimal value and perform
similarly, while in the high contamination setting, \probosp with denoising GP
converges to a near optimal value while standard BO with GPs does not.
%

\subsection{BO with Prior Structure on the Objective Function}
\label{sec:nas}

\subsubsection*{Setting:}
In some cases, we have prior knowledge about properties of the objective
function, such as trends with respect to $x \in \mathcal{X}$.  For example,
consider the task of tuning hyperparameters of a machine learning model, where
the hyperparameters correspond with model complexity.  For datasets of
moderate size, there are often two distinct phases as model complexity grows: a
phase where the model underfits, where increasing modeling complexity reduces
error on a held-out validation set; and a phase where the model overfits, where
validation error increases with respect to model complexity. We can design a
model that leverages trends such as these.

\subsubsection*{Model:}
We design a model for tuning model complexity, which we refer to as a basin
model. Let
$y \sim \mathcal{N}(R(x-\mu;a,b)+c,\sigma^2)$ where $R(x;a,b)$ $=$ $a^T
\text{ReLU}(x) + b^T \text{ReLU}(-x)$, with priors on parameters $\mu\in
\mathbb{R}_d$, $a, b\in\mathbb{R}_d^+$, $c\in \mathbb{R}$, and $\sigma^2>0$.
This model captures the inflection point with variable $\mu$, and uses
variables $a$ and $b$ to model the slope of the optimization landscape above
and below (respectively) $\mu$.  We give a one dimensional view of validation
error data from an example where $x$ corresponds to neural network layer width,
and show inference with a basin model for this data in
Fig.~\ref{fig:priorstruc}~(b).

\subsubsection*{Empirical Results:}
In this experiment, we optimize over the number of units (i.e. layer width) of
the hidden layers in a four layer MLP 
trained on the Wisconsin Breast Cancer
Diagnosis dataset \cite{blake1998uci}. We compare \probosp using a basin model,
with standard BO using a GP. We see in Fig~\ref{fig:priorstruc}~(c)-(d) that
\probosp with the basin model can significantly outperform standard BO with
GPs.  In this optimization task, the landscape around the inflection point (of
under to over fitting) can be very steep, which may hurt the performance of GP
models. In contrast, the basin model can capture this shape and quickly
identify the inflection point via inferences about $\mu$.


\setlength{\minipagewidth}{0.4\textwidth}
\setlength{\minipageoffset}{0mm}
\setlength{\columngap}{0mm}
\setlength{\pretitlegap}{2mm}
\setlength{\posttitlegap}{3mm}

\begin{figure}[tp]
  \centering

  \begin{minipage}[b]{\textwidth}
    \centering
    \vspace{\pretitlegap}
    \underline{\emph{Structured Multi-task BO}}
    \vspace{\posttitlegap}
  \end{minipage}\\

  \makebox[\textwidth]{\makebox[1.25\textwidth]{
    \begin{minipage}[b]{\minipagewidth}
      \hspace{\minipageoffset}
      \centering
      {\small (a) GP (two task)}
    \end{minipage}
    \hspace{\columngap}
    \begin{minipage}[b]{\minipagewidth}
      \hspace{\minipageoffset}
      \centering
      {\small (b) Warp model (two task)}
    \end{minipage}
    \hspace{\columngap}
    \begin{minipage}[b]{\minipagewidth}
      \hspace{\minipageoffset}
      \centering
      {\small (c) EI}
    \end{minipage}
  }}

  \makebox[\textwidth]{\makebox[1.25\textwidth]{
    \begin{minipage}[b]{\minipagewidth}
      \includegraphics[width=\textwidth]{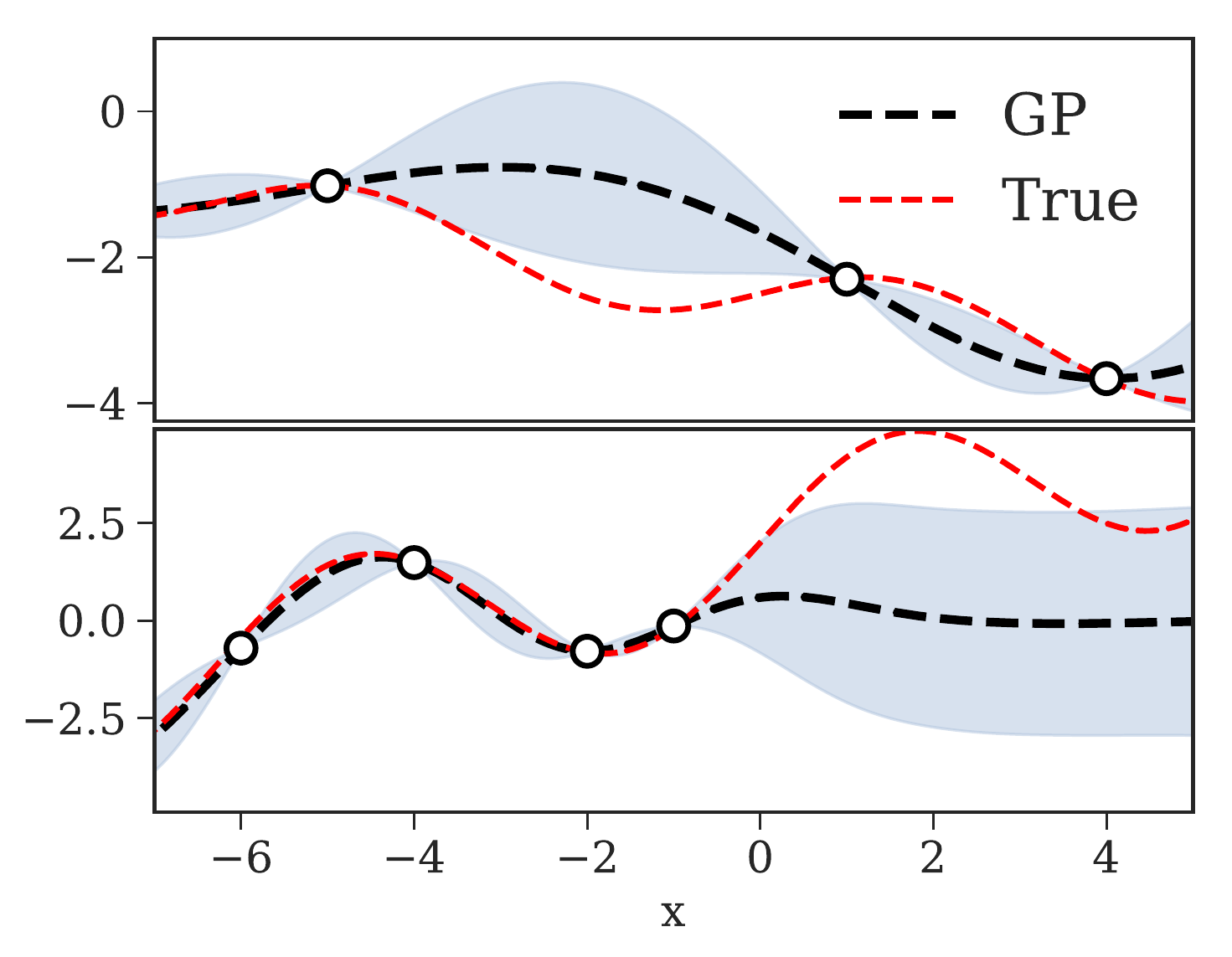}
    \end{minipage}
    \hspace{\columngap}
    \hspace{-4mm}
    \begin{minipage}[b]{\minipagewidth}
      \includegraphics[width=\textwidth]{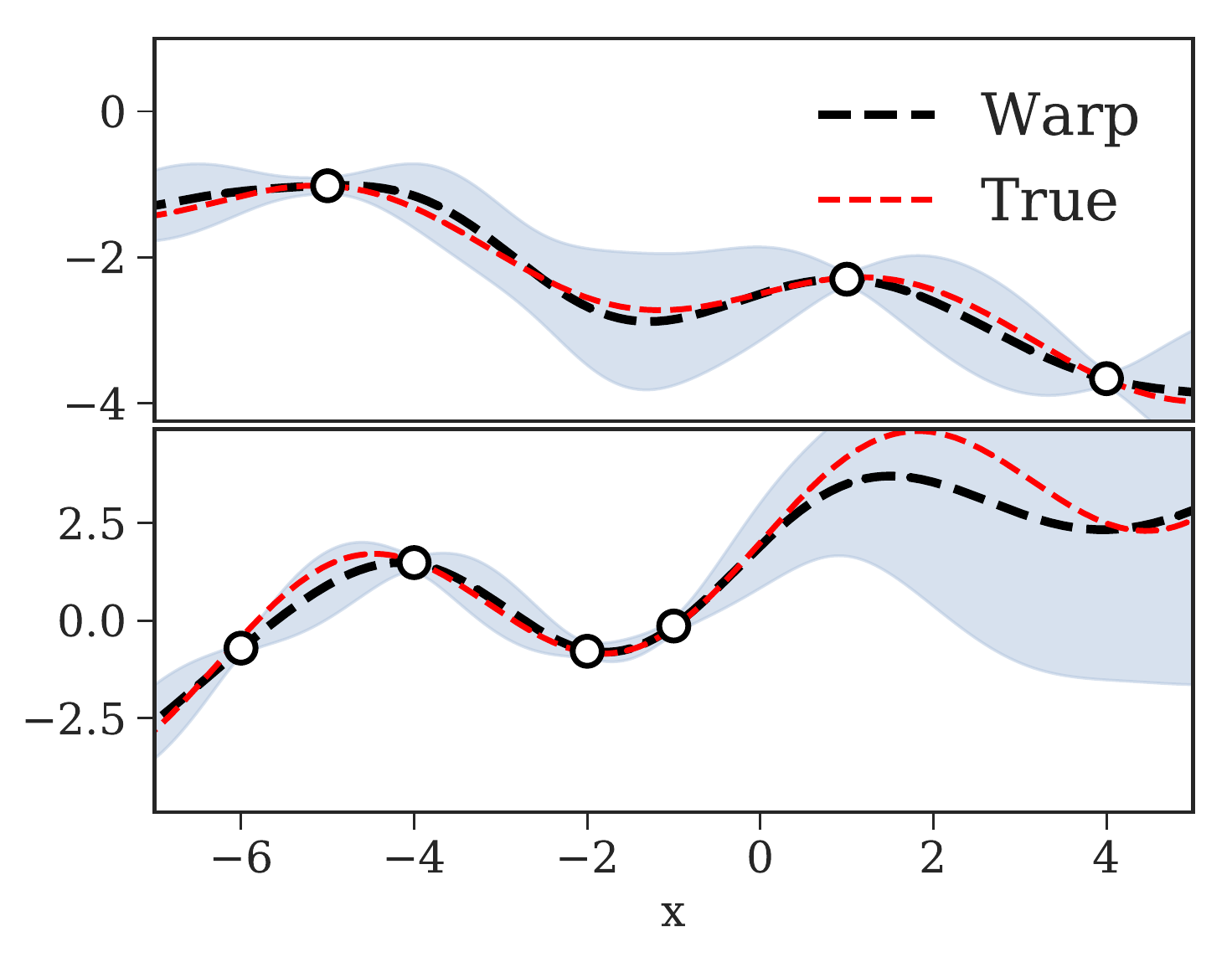}
    \end{minipage}
    \hspace{\columngap}
    \begin{minipage}[b]{\minipagewidth}
      \includegraphics[width=\textwidth]{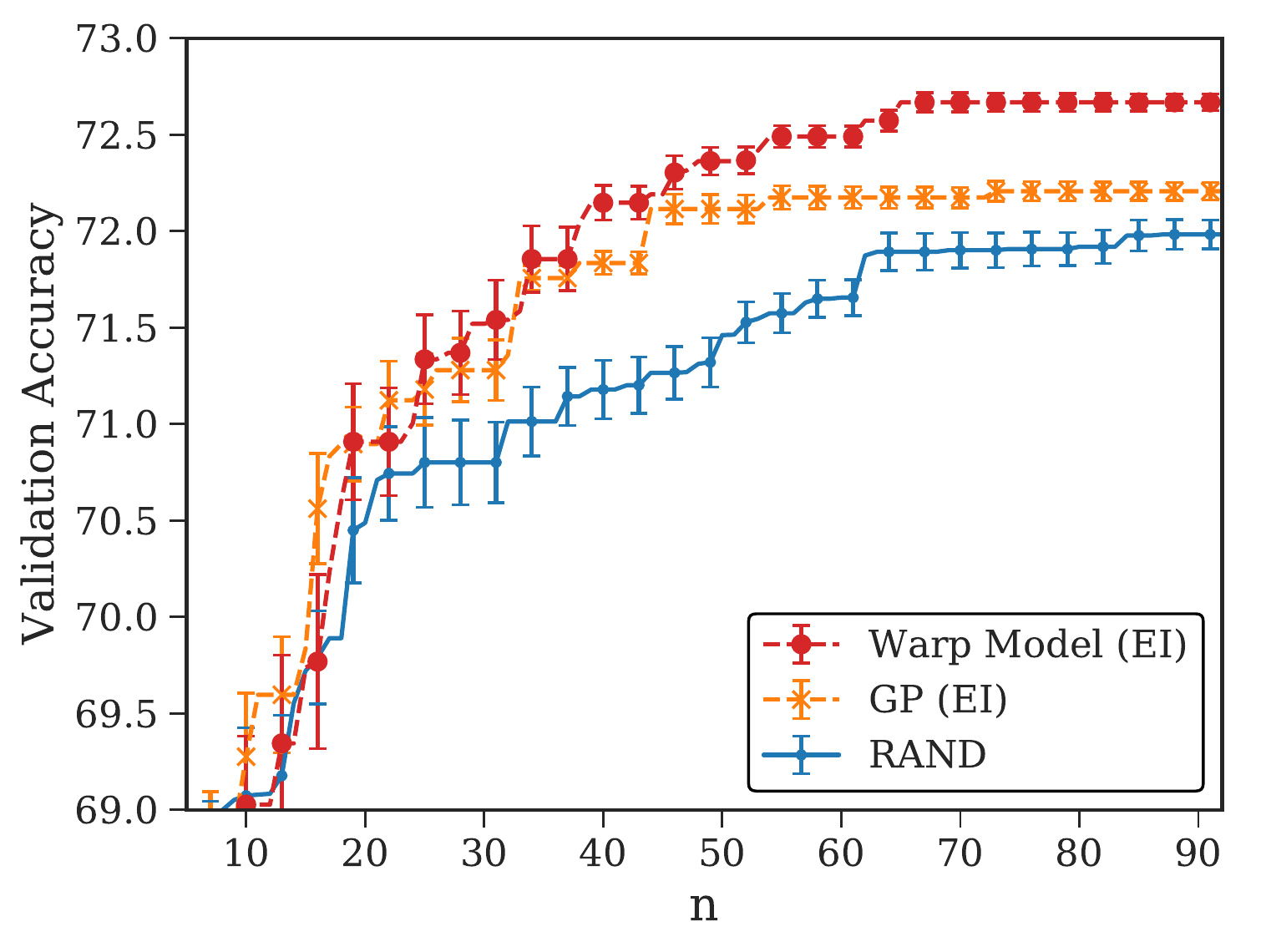}
    \end{minipage}
  }}

  \caption{
    Structured multi-task BO (Sec.~\ref{sec:multitask_and_ensembles}). We show
    (a) independent GPs and (b) our warp model, in a two-task setting (task one
    on top, task two on bottom).
    In (c) we show results for structured multi-task BO on a neural network
    hyperparameter search problem (details in Appendix
    Sec.~\ref{sec:multitaskbo}).  Curves are averaged over 10 trials, and error
    bars represent one standard error.
    \label{fig:multitask}
  }
\end{figure}


\setlength{\minipagewidth}{0.4\textwidth}
\setlength{\minipageoffset}{0mm}
\setlength{\columngap}{0mm}
\setlength{\pretitlegap}{2mm}
\setlength{\posttitlegap}{1mm}

\begin{figure}[tp]
  \centering

  \begin{minipage}[b]{\textwidth}
    \centering
    \vspace{\pretitlegap}
    \underline{\emph{Multi-fidelity Acquisition Functions}}
    \vspace{\posttitlegap}
  \end{minipage}\\

  \makebox[\textwidth]{\makebox[1.25\textwidth]{
    \begin{minipage}[b]{\minipagewidth}
      \hspace{5mm}
      \hspace{\minipageoffset}
      \centering
      {\small (a) EI}
    \end{minipage}
    \hspace{\columngap}
    \begin{minipage}[b]{\minipagewidth}
      \hspace{\minipageoffset}
      \centering
      {\small (b) UCB}
    \end{minipage}
    \hspace{\columngap}
    \begin{minipage}[b]{\minipagewidth}
      \hspace{\minipageoffset}
      \centering
      {\small (c) Calls to \texttt{gen}}
    \end{minipage}
  }}

  \makebox[\textwidth]{\makebox[1.25\textwidth]{
    \begin{minipage}[c]{\minipagewidth}
      \centering \vspace{0pt}
      \includegraphics[width=\textwidth]{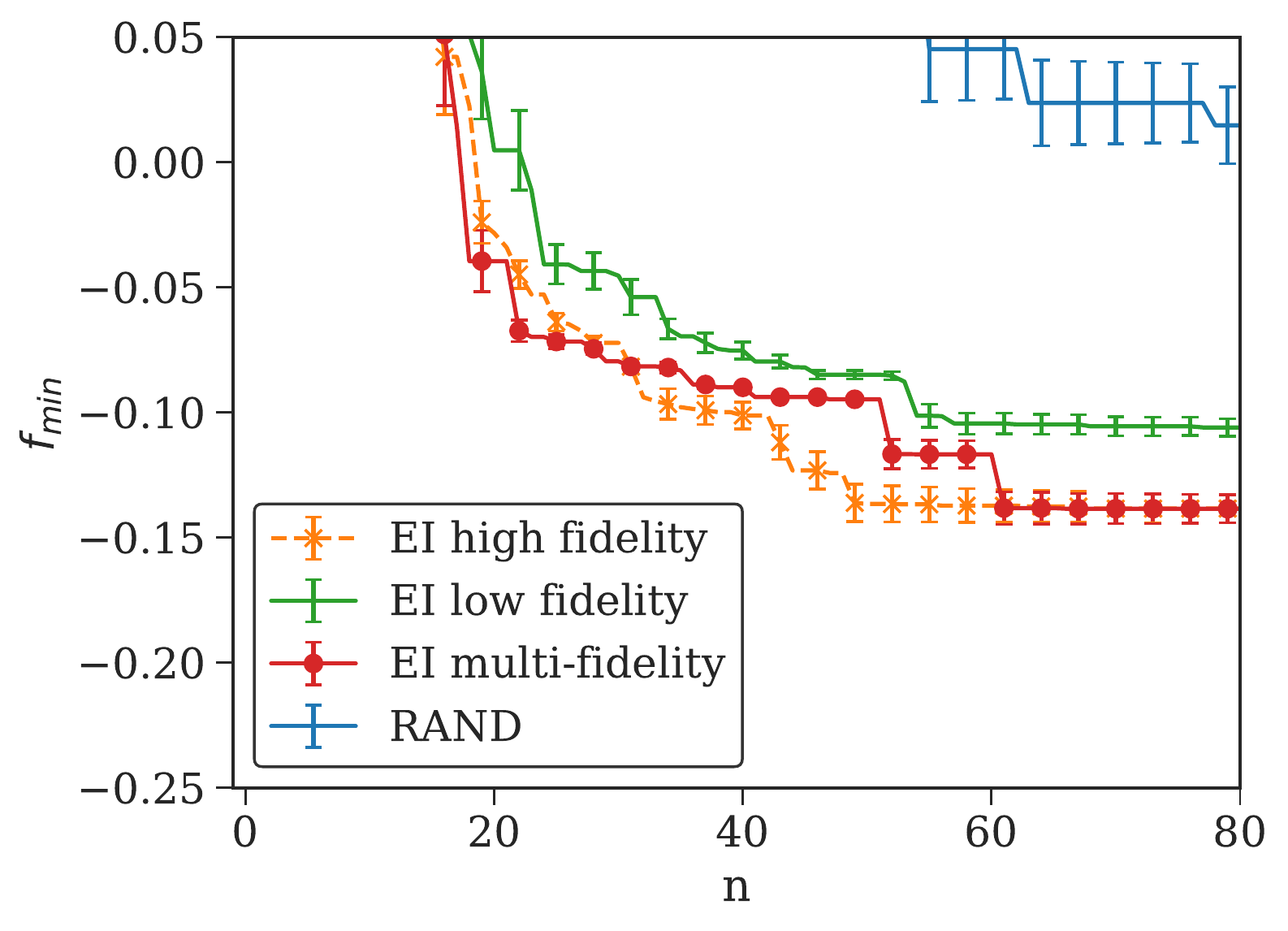}
    \end{minipage}
    \hspace{\columngap}
    \hspace{-4mm}
    \begin{minipage}[c]{\minipagewidth}
      \centering \vspace{0pt}
      \includegraphics[width=\textwidth]{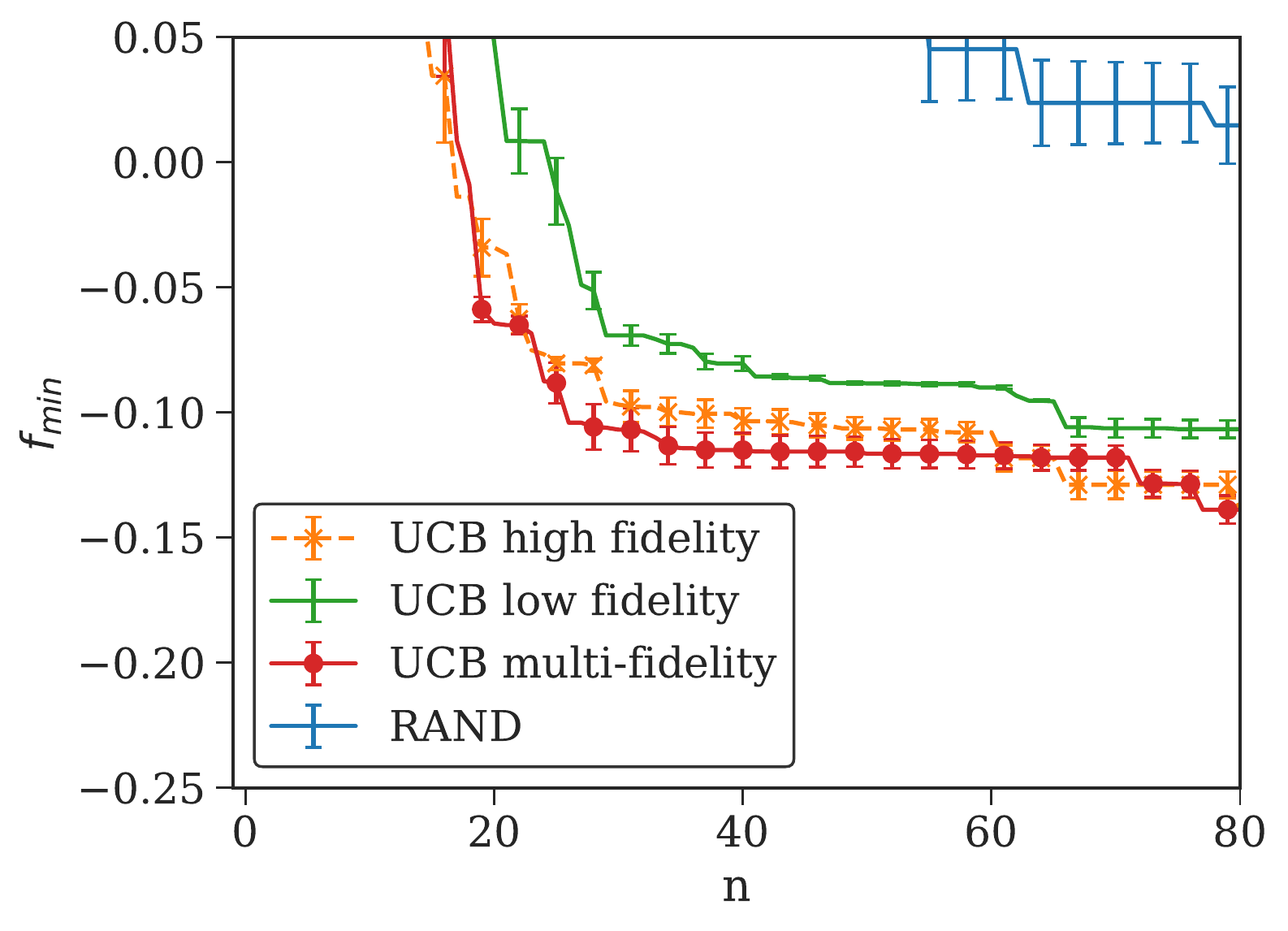}
    \end{minipage}
    \hspace{\columngap}
    \begin{minipage}[c]{\minipagewidth}
      \centering \vspace{0pt}
      \includegraphics[width=.9\textwidth]{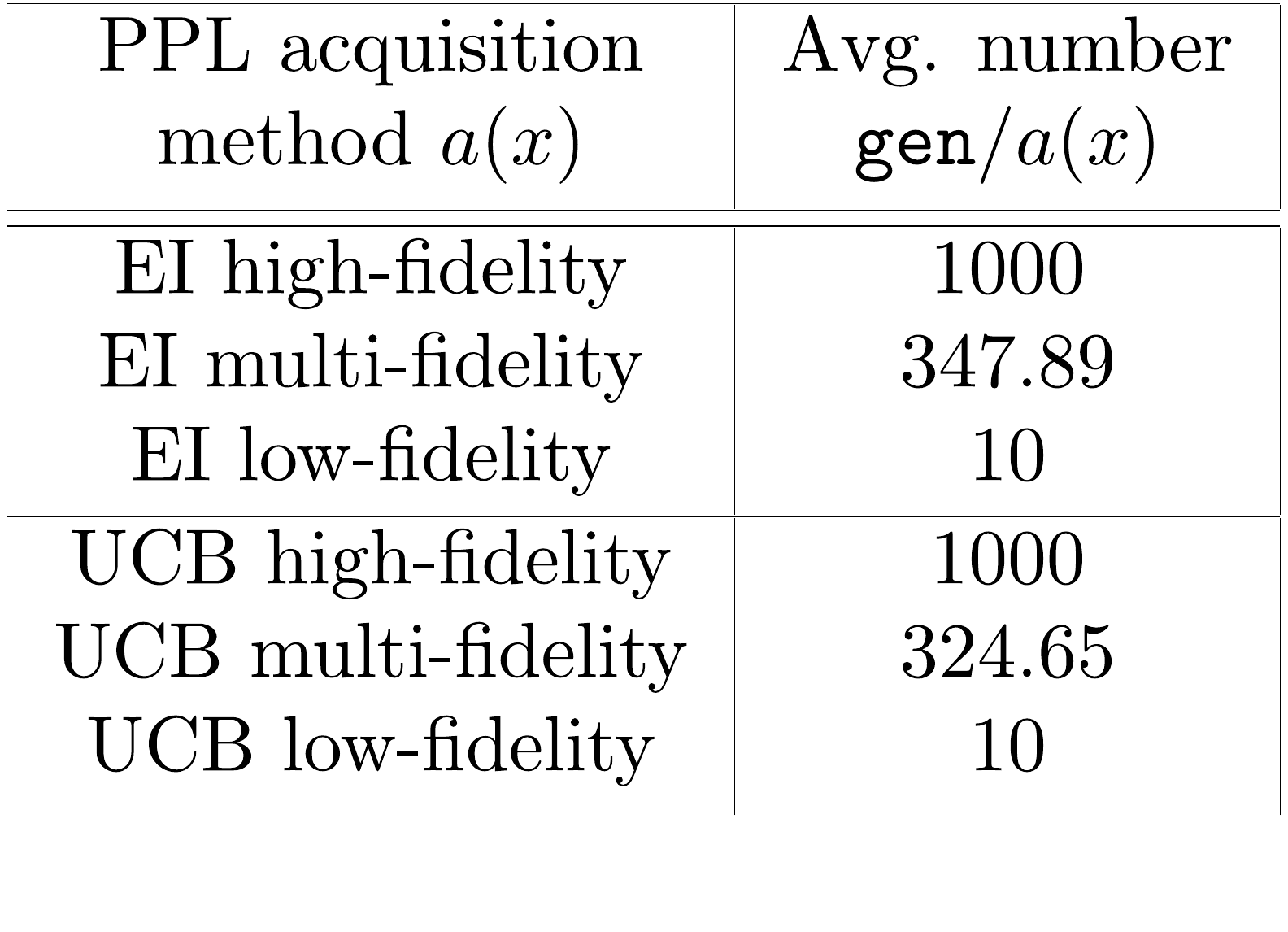}
    \end{minipage}
  }}

  \caption{
    Results on $a_\text{MF}$ experiments (Sec.~\ref{sec:mfexp}), showing
    (a)-(b) \probosp using $a_\text{MF}$ (Alg.~\ref{alg:mf}) vs using a fixed
    high-fidelity $a$ ($M=1000$) and a fixed low-fidelity $a$ ($M=10$). Here,
    $a_\text{MF}$ performs competitively with the high-fidelity $a$, while low
    fidelity $a$ performs worse. In (c) we show the average number of
    \texttt{post}/\texttt{gen} calls per evaluation of $a$.  We see that the
    $a_\text{MF}$ reduces the number of calls.
    Curves are averaged over 10 trials, and error bars represent one standard
    error.
    \label{fig:mfacqopt}
  }
  \vspace{-5mm}
\end{figure}

\subsection{Structured Models for Multi-task and Contextual BO, and Model Ensembles}
\label{sec:multitask_and_ensembles}
We may want to optimize multiple systems jointly, where there is some known
relation between the systems. In some instances, we have a finite set of
systems (multi-task BO) and in some cases systems are each indexed
by a context vector $c \in \mathbb{R}^d$ (contextual BO). 
We develop a model that can incorporate prior structure about the relationship
among these systems.
Our model warps a latent model based on context/task-specific parameters, so we
call this a warp model.  We show inference in this model in
Fig.~\ref{fig:multitask}~(b).  In Appendix Sec.~\ref{sec:multitaskbo} we define
this model, and describe experimental results shown in
Fig.~\ref{fig:multitask}~(c).

Alternatively, we may have multiple models that capture different aspects of a system, or we
may want to incorporate information given by, for instance, a parametric model
(e.g. a model with a specific trend, shape, or specialty for a subset of the
data) into a nonparametric model (e.g. a GP, which is highly flexible, but has
fewer assumptions). To incorporate multiple sources of information or bring in
side information, we want a valid way to create ensembles of multiple PPL
models. We develop strategies to combine the posterior predictive densities of
multiple PPL models, using only our three PPL operations. We describe 
this in Appendix Sec.~\ref{sec:bpoe}.




\subsection{Multi-fidelity Acquisition Optimization}
\label{sec:mfexp}
We empirically assess our multi-fidelity acquisition function
optimization algorithm (Sec.~\ref{sec:mf}). Our goal is to demonstrate that
increasing the fidelity $M$ in black box acquisitions can yield better
performance in \probo, and that our multi-fidelity method (Alg.~\ref{alg:mf})
maintains the performance of the high-fidelity acquisitions while reducing the
number of calls to \texttt{post} and \texttt{gen}.
We perform an experiment in a two-fidelity setting, where $M \in \{10,1000\}$,
and we apply $a_\text{MF}$ to EI and UCB, using a GP model and the
(non-corrupted) synthetic system described in Sec.~\ref{sec:excontam}. Results
are shown in Fig.~\ref{fig:mfacqopt}~(a)-(c), where we compare
high-fidelity $a$ ($M = 1000$), low-fidelity $a$ ($M=10$), and multi-fidelity
$a_\text{MF}$, for EI and UCB acquisitions.  For both, the high-fidelity and
multi-fidelity methods show comparable performance, while the low-fidelity
method performs worse. We also see in Fig.~\ref{fig:mfacqopt}~(c) that the
multi-fidelity method reduces the number of calls to \texttt{post}/\texttt{gen}
by a factor of 3, on average, relative to the high fidelity method.

\section{Conclusion}
\label{sec:conclusion}

In this paper we presented \probo, a system for performing Bayesian
optimization using models from any probabilistic programming language.  We
developed algorithms to compute acquisition functions using common PPL
operations (without requiring model-specific derivations), and showed how to
efficiently optimize these functions. We presented a few models for challenging
optimization scenarios, and we demonstrated promising empirical results on the
tasks of BO with state observations, contaminated BO, BO with prior structure
on the objective function, and structured multi-task BO, where we were able to
drop-in models from existing PPL implementations.


{\small \bibliographystyle{amsplain}
        \bibliography{main}}
\markboth{APPENDIX}{APPENDIX}
\appendix
\onecolumn
\begin{changemargin}{0cm}{0cm}
\appsection{Structured Models for Multi-task and Contextual BO}
\label{sec:multitaskbo}

In this section we provide details about our models for multi-task and
contextual BO, described in Sec.~\ref{sec:multitask_and_ensembles} and
shown in Fig.~\ref{fig:multitask}.
Many prior methods have been proposed for for multi-task BO
\citep{swersky2013multi} and contextual BO \citep{krause2011contextual}, though
these often focus on new acquisition strategies for GP models. Here we propose
a model for the case where we have some structured prior information about the
relation between systems, or some parametric relationship that we want to
incorporate into our model.

We first consider the multi-task setting and then extend this to the contextual
setting.  Suppose that we have  $T$ tasks
(i.e. subsystems to optimize) with data subsets $\mathcal{D}$ $=$
$\left\{\mathcal{D}_1,\ldots,\mathcal{D}_T\right\}$, where $\mathcal{D}_t$ has
data $\{x_{t,i}, y_{t,i}\}_{i=1}^{n_t}$, and where $n_t$ denotes the number of
observations in the $t^\text{th}$ task.  For each $(x_{t,i},y_{t,i})$ pair
within $\mathcal{D}$, we have a latent variable $z_{t,i} \in \mathcal{Z}$.
Additionally, for each task $t$, we have task-specific latent ``warp''
variables, denoted $w_t$.

Given these, we define our warp model to be
\begin{enumerate}[topsep=5pt,itemsep=5pt,parsep=0pt]
\item For $t=1,\ldots,T$:
  \begin{enumerate}
  \item $w_t \sim \text{Prior}(w)$
  \item For $i=1,\ldots,n_t$:
    \begin{enumerate}
      \item $z_{t,i} \sim p(z|x_{t,i})$
      \item $y_{t,i} \sim p(y|z_{t,i}, x_{t,i}, w_t)$
    \end{enumerate}
  \end{enumerate}
\end{enumerate}

We call $p(z|x_{t,i})$ our latent model, and $p(y|z_{t,i}, x_{t,i}, w_t)$ our
warp model, which is parameterized by warp variables $w_t$.  Intuitively, we
can think of the variables $z_{t,i}$ as latent ``unwarped'' versions of
observations $y_{t,i}$, all in a single task.  Likewise, we can intuitively
think of observed variables $y_{t,i}$ as ``warped versions of $z_{t,i}$'',
where $w_t$ dictates the warping for each subset of data $\mathcal{D}_t$.

\begin{figure}[bp]
  \centering

  \begin{minipage}[b]{0.49\textwidth}
    \hspace{4mm}
    \centering
    \textbf{Task 1 ($n_1=7$)}
  \end{minipage}
  \hfill
  \begin{minipage}[b]{0.49\textwidth}
    \hspace{4mm}
    \centering
    \textbf{Task 2 ($n_2=5$)}
  \end{minipage}\\

  \begin{minipage}[b]{0.49\textwidth}
    \includegraphics[width=\textwidth]{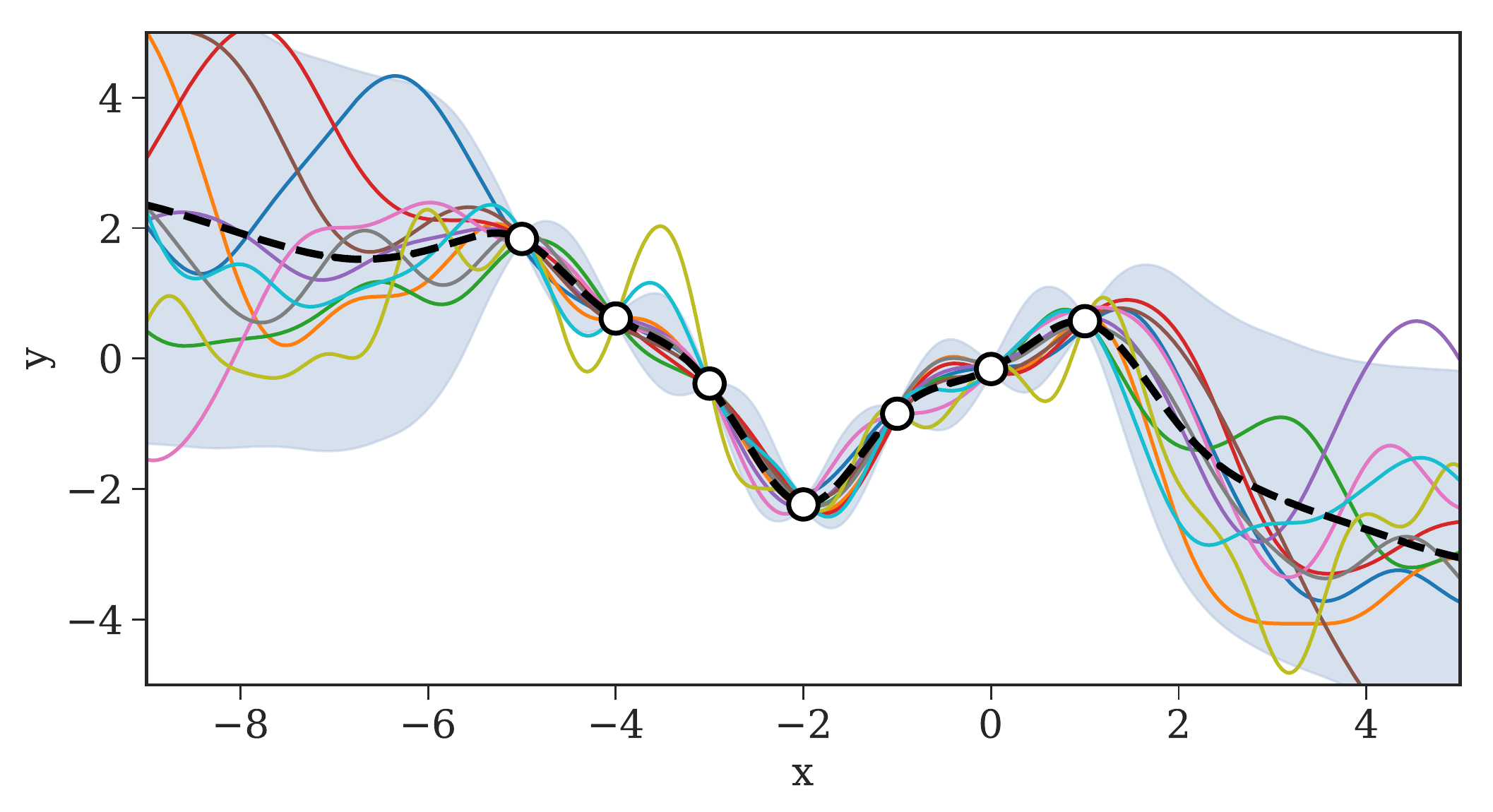}
  \end{minipage}
  \hfill
  \begin{minipage}[b]{0.49\textwidth}
    \includegraphics[width=\textwidth]{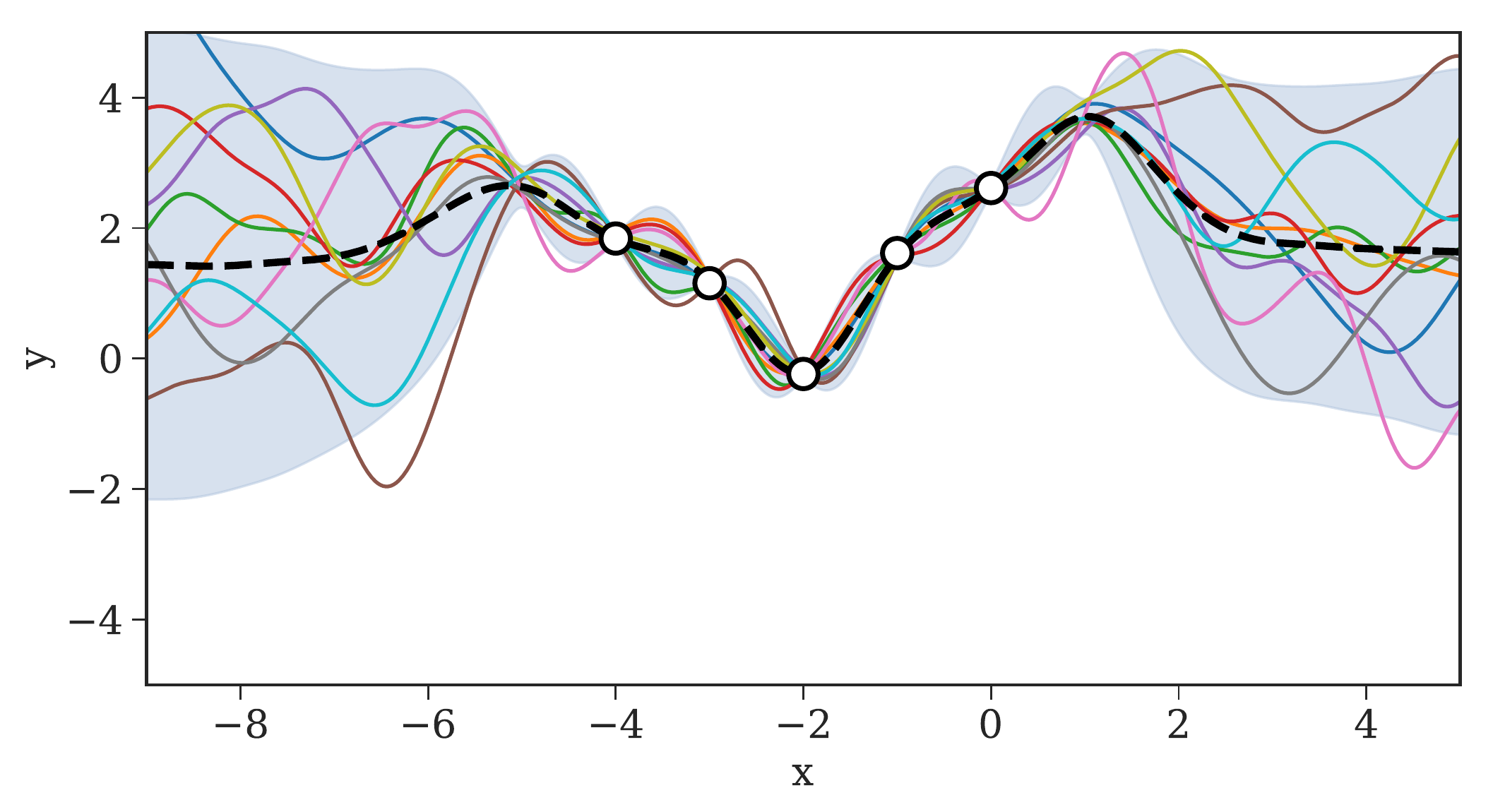}
  \end{minipage}

  \caption{Warp model inference on tasks one (left) and two (right), where
    $n_1=7$ and $n_2=5$.  This warp model assumes a linear warp with respect to
    both the latent variables $z$ and inputs $x$. Posterior mean, posterior
    samples, and posterior predictive distribution are shown.
    \label{fig:warpviz1}}
\end{figure}

\begin{figure}[!tbp]
  \centering

  \begin{minipage}[b]{0.49\textwidth}
    \hspace{4mm}
    \centering
    \textbf{Task 1 ($n_1=7$)}
  \end{minipage}
  \hfill
  \begin{minipage}[b]{0.49\textwidth}
    \hspace{4mm}
    \centering
    \textbf{Task 2 ($n_2=3$)}
  \end{minipage}\\

  \begin{minipage}[b]{0.49\textwidth}
    \includegraphics[width=\textwidth]{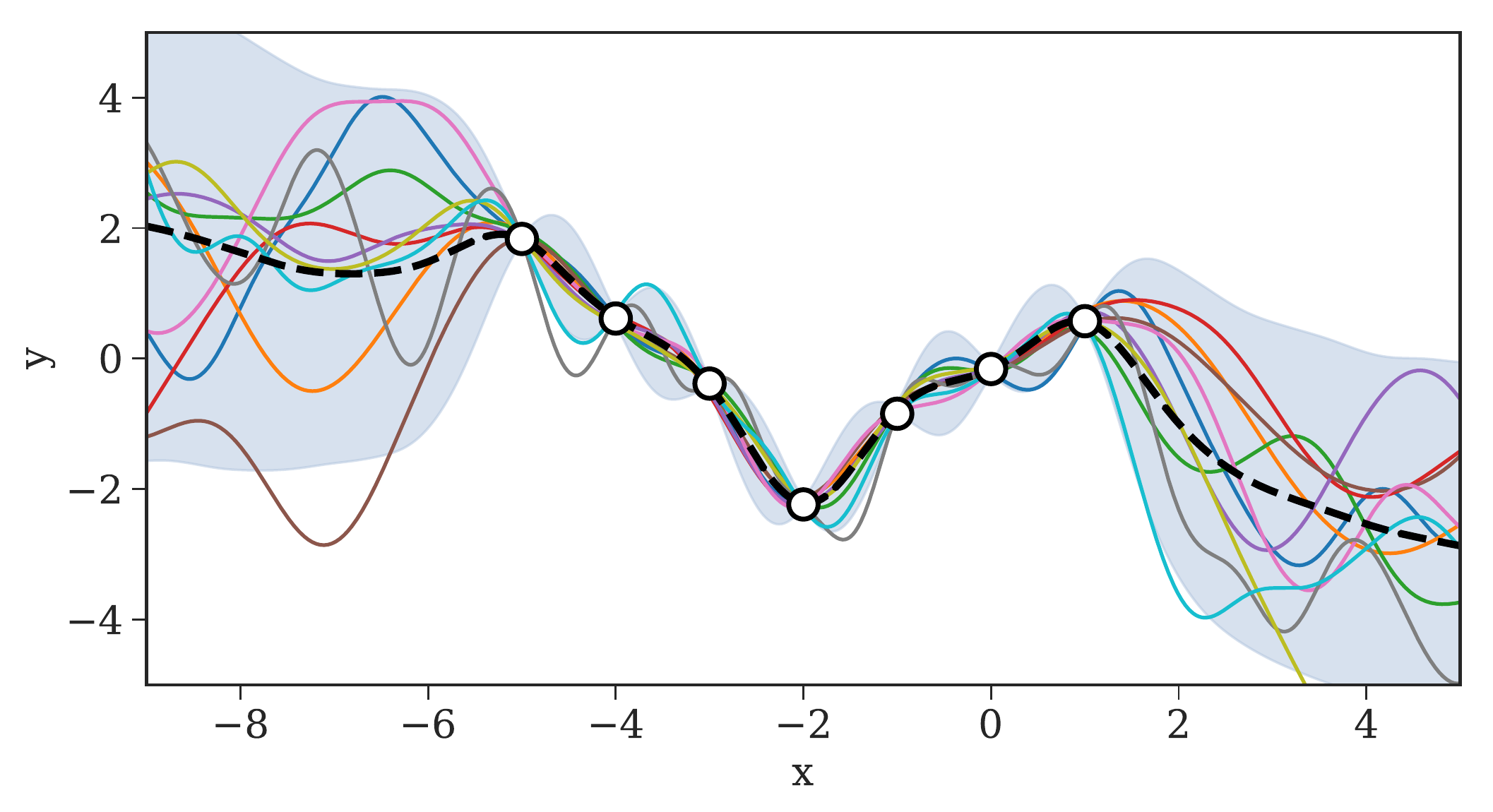}
  \end{minipage}
  \hfill
  \begin{minipage}[b]{0.49\textwidth}
    \includegraphics[width=\textwidth]{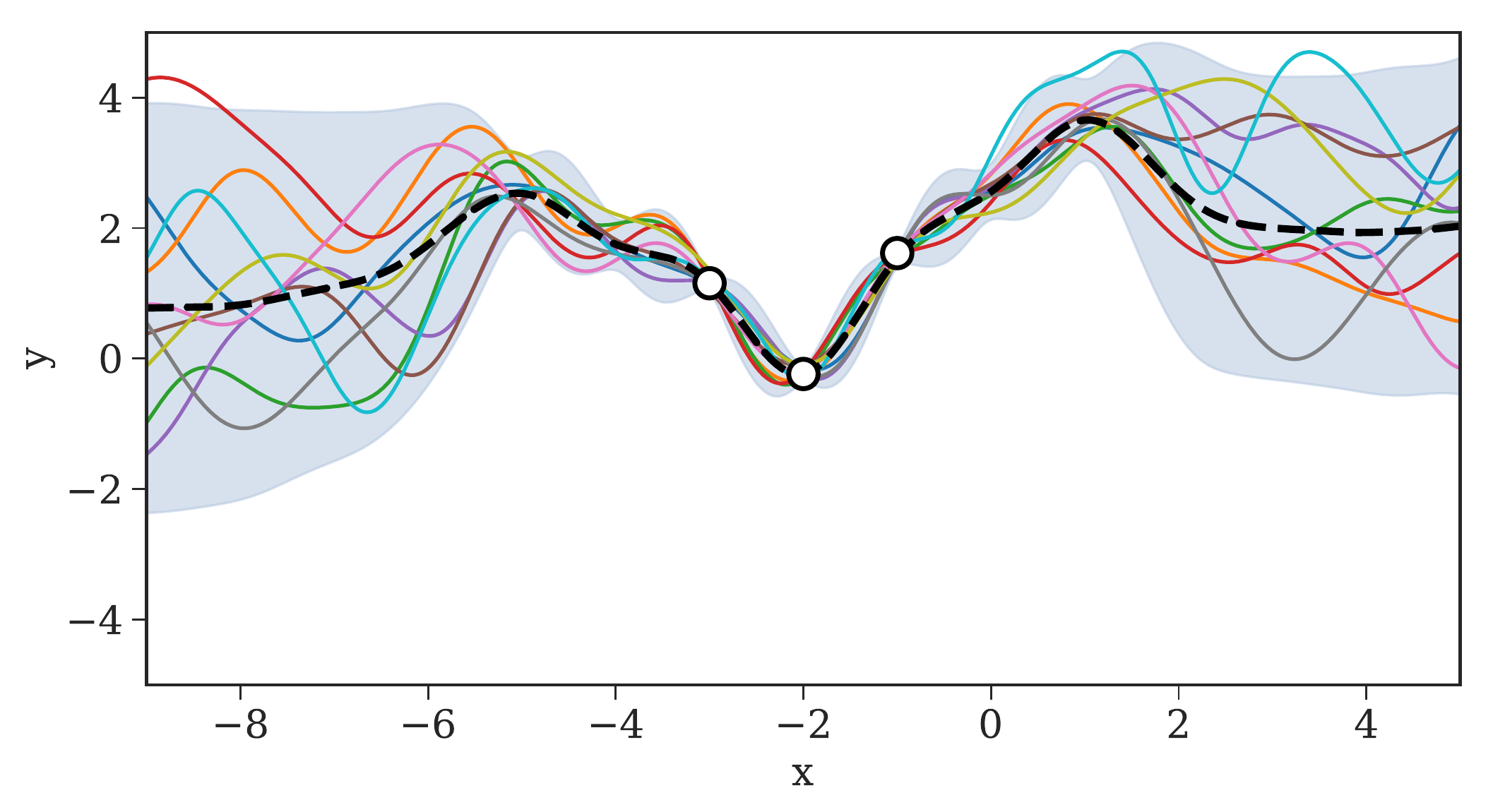}
  \end{minipage}

  \caption{Warp model inference on tasks one (left) and two (right), where
    $n_1=7$ and $n_2$ is reduced to $n_2=3$.  This warp model assumes a linear
    warp with respect to both the latent variables $z$ and inputs $x$.
    Posterior mean, posterior samples, and posterior predictive distribution
    are shown. Here, we see more uncertainty around the two removed points in
    task two, relative to Fig.~\ref{fig:warpviz1}.
    \label{fig:warpviz2}}
\end{figure}



We now give a concrete instantiation of this model.  Let the latent model be
$p(z|x_{t,i})$ $=$ $\mathcal{GP}\left(\mu(x), k(x,x')\right)$, i.e. we put a
Gaussian process prior on the latent variables $z$.
For a given task, let the warping model for $y$ be a linear function of both
$z$ and $x$ (with some added noise), where warping parameters $w$ are
parameters of this linear model, i.e. $y_{t,i} \sim w_0 + w_1 x_{t,i} + w_2
z_{t,i} + \epsilon$.

Intuitively, this model assumes that there is a latent GP, which is warped via
a linear model of both the GP output $z$ and input $x$ to yield observations
$y$ for a given task (and that there is a separate warp for each task). We
illustrate this model in Fig.~\ref{fig:warpviz1} and \ref{fig:warpviz2} (where
$n_1 = 7$ in both, $n_2=5$ in the former, and $n_2=3$ in the latter).  As we
remove points $x$ in task two, we see more uncertainty in the posterior
predictive distribution at these points.

We can also extend this warp model for use in a contextual optimization
setting, where we want to jointly optimize over a set of systems each indexed
by a context vector $c \in \mathbb{R}^d$. In practice, we observe the context
$c_i$ for input $x_i \in \mathcal{X}$, and therefore perform inference on a
dataset $\mathcal{D}_n = \{x_i,c_i,y_i\}_{i=1}^n$.

To allow for this, we simply let our warp model also depend on $c$, i.e. let
the warp model be $p(y|z_{t,i}, x_{t,i}, w_t, c_{t,i})$. Intuitively, this
model assumes that there is a single latent system, which is warped by various
factors (e.g. the context variables) to produce observations.

\subsection{Empirical Results}
Here we describe the empirical results shown in Fig.~\ref{fig:multitask}~(c).
We aim to perform the neural architecture and hyperparameter search task
from Sec.~\ref{sec:exstate}, but for two different settings, each with a unique
preset batch size. Based on prior observations, we believe that the validation
accuracy of both systems at a given query $x$ can be accurately modeled as a
linear transformation of some common latent system, and we apply the warp model
described above. We compare \probosp using this warp model with a single GP
model over the full space of tasks and inputs. We show results in
Fig.~\ref{fig:multitask}~(c), where we plot the best validation accuracy found
\emph{over both tasks} vs iteration $n$.  Both methods use the EI acquisition
function, and we compare these against a baseline that chooses queries
uniformly at random. Here, \probosp with the warp model is able to find a query
with a better maximum validation accuracy, relative to standard BO with GP
model.

\appsection{Ensembles of PPL Models within \probo}
\label{sec:bpoe}

We may have multiple models that capture different aspects of a system, or we
may want to incorporate information given by, for instance, a parametric PPL
model (e.g. a model with a specific trend, shape, or specialty for a subset of
the data) into a nonparametric PPL model (e.g. a GP, which is flexible, but has
fewer assumptions).

To incorporate multiple sources of information or bring in side information, we
want a valid way to create ensembles of multiple models. Here, we develop a
method to combine the posterior predictive densities of multiple PPL models,
using only our three PPL operations. Our procedure constructs a model similar
to a product of experts model \cite{hinton2002training}, and we call our
strategy a Bayesian product of experts (BPoE). This model can then be used in
our \probosp framework.

As an example, we show an ensemble of two models, $\mathcal{M}_1$ and
$\mathcal{M}_2$, though this could be extended to an arbitrarily large group of
models.  Assume $\mathcal{M}_1$ and $\mathcal{M}_2$ are both plausible models
for a dataset $\mathcal{D}_n = \{(x_i,y_i)\}_{i=1}^n$.

Let $\mathcal{M}_1$ have likelihood $p_1(\mathcal{D}_n|z_1)$ $=$ $\prod_{i=1}^n p_1(y_i |
z_1 ; x_i)$, where $z_1 \in \mathcal{Z}_1$ are latent variables with prior
$p_1(z_1)$. We define the joint model PDF for $\mathcal{M}_1$ to be
$p_1(\mathcal{D}_n,z_1)$ $=$ $p_1(z_1) \prod_{i=1}^n p_1(y_i | z_1 ; x_i)$.  The
posterior (conditional) PDF for $\mathcal{M}_1$ can then be written
$p_1(z_1|\mathcal{D}_n) = p_1(\mathcal{D}_n,z_1) / p_1(\mathcal{D}_n)$.  We can
write the posterior predictive PDF for $\mathcal{M}_1$ as
\begin{align}
  p_1(y|\mathcal{D}_n;x) = \mathbb{E}_{p_1(z_1|\mathcal{D}_n)}\left[ p_1(y|z_1;x) \right].
\end{align}

Similarly, let $\mathcal{M}_2$ have likelihood $p_2(\mathcal{D}_n|z_2)$ $=$
$\prod_{i=1}^n p_2(y_i | z_2 ; x_i)$, where $z_2 \in \mathcal{Z}_2$ are latent
variables with prior PDF $p_2(z_2)$. We define the joint model PDF for
$\mathcal{M}_2$ to be $p_2(\mathcal{D}_n,z_2)$ $=$ $p_2(z_2) \prod_{i=1}^n
p_2(y_i | z_2 ; x_i)$, the posterior (conditional) PDF to be
$p_2(z_2|\mathcal{D}_n)$ $=$ $p_2(\mathcal{D}_n,z_2) / p_2(\mathcal{D}_n)$, and
the posterior predictive PDF to be
\begin{align}
  p_2(y|\mathcal{D}_n;x) = \mathbb{E}_{p_2(z_2|\mathcal{D}_n)}\left[ p_2(y|z_2;x) \right].
\end{align}

Note that $z_1 \in \mathcal{Z}_1$ and $z_2 \in \mathcal{Z}_2$ need not be in
the same space nor related.

Given models $\mathcal{M}_1$ and $\mathcal{M}_2$, we define the
Bayesian Product of Experts (BPoE) ensemble model, $\mathcal{M}_e$,
with latent variables $z = (z_1,z_2) \in \mathcal{Z}_1 \times \mathcal{Z}_2$,
to be the model with posterior predictive density
\begin{align}
  p(y|\mathcal{D}_n;x) \propto p_1(y|\mathcal{D}_n;x) p_2(y|\mathcal{D}_n;x).
\end{align}
The posterior predictive PDF for the BPoE ensemble model $\mathcal{M}_e$
is proportional to the product of the posterior predictive PDFs of the
constituent models $\mathcal{M}_1$ and $\mathcal{M}_2$.  Note that this uses
the product of expert assumption \cite{hinton2002training} on $y$, which
intuitively means that $p(y|\mathcal{D}_n;x)$ is high where both
$p_1(y|\mathcal{D}_n;x)$ and $p_2(y|\mathcal{D}_n;x)$ agree (i.e. an ``and''
operation). Intuitively, this model gives a stronger posterior belief over $y$ 
in regions where both models have consensus, and weaker posterior belief over
$y$ in regions given by only one (or neither) of the models.

Given this model, we need an algorithm for computing and using the posterior
predictive for $\mathcal{M}_e$ within the \probosp framework. In our
acquisition algorithms, we use \texttt{gen} to generate samples from predictive
distributions.  We can integrate these with combination algorithms from the
embarrassingly parallel MCMC literature \cite{neiswanger2013asymptotically2,
wang2015parallelizing, neiswanger2017post} to develop an algorithm that
generates samples from the posterior predictive of the ensemble model
$\mathcal{M}_e$ and uses these in a new acquisition algorithm. We give this
procedure in Alg.~\ref{alg:bpoe}, which introduces the \texttt{ensemble}
operation. Note that \texttt{ensemble} takes as input two operations
\texttt{gen1} and \texttt{gen2} (assumed to be from two PPL models), as well as
two sets of $M$ posterior samples $z_1^M$ and $z_2^M$ (assumed to come from
calls to \texttt{post1} and \texttt{post2} from the two PPL models). Also note
that in Alg.~\ref{alg:bpoe} we've used $\text{Combine}(y_1,y_2)$ to denote a
combination algorithm, which we detail in appendix Sec.~\ref{sec:app_combine}.

We can now swap the \texttt{ensemble} operation in for the \texttt{gen} operation
in Algs.~\ref{alg:ei}-\ref{alg:ts}.
Note that the BPoE allows us to easily ensemble models written in different PPLs.
For example, a hierarchical regression model written in Stan
\cite{carpenter2015stan} using Hamiltonian Monte Carlo for inference could be
combined with a deep Bayesian neural network written in Pyro
\cite{bingham2018pyro} using variational inference and with a GP written in GPy
\cite{gpy2014} using exact inference.

{\centering
\begin{algorithm}[H]
    \caption{\hspace{1mm} $\texttt{ensemble}$$(x,\texttt{gen1},$
      $\texttt{gen2},z_1^M,z_2^M)$
      \hfill $\triangleright$ PPL model ensemble with BPoE}
    \label{alg:bpoe}
    \begin{algorithmic}[1]
      \For{$m = 1,\ldots,M$}
        \State $s_1,s_2 \sim \text{Unif}\left(\{1,\ldots,M\}\right)$ 
        \State $\tilde{y}_{1,m} \leftarrow \texttt{gen1}(x,z_{1,s_1},s_1)$
        \State $\tilde{y}_{2,m} \leftarrow \texttt{gen2}(x,z_{2,s_2},s_2)$
      \EndFor
      \State $y_{1:M} \leftarrow \text{Combine}(\tilde{y}_{1,M},\tilde{y}_{2,M})$
      \State Return $y_{1:M}$.
    \end{algorithmic}
\end{algorithm}
}

\subsection{Example: Combining Phase-Shift and GP Models.}
\label{sec:phaseshift}

We describe an example and illustrate it in Fig.~\ref{fig:combine}. Suppose we
expect a few phase shifts in our input space $\mathcal{X}$, which partition
$\mathcal{X}$ into regions with uniform output. We can model this system
with
$y$ $\sim$ $\mathcal{N}( y|\sum_{k=1}^K
\text{logistic}(x;m_k,s_k,\mu_k)+b_k,\sigma^2 )$, where latent variables
$m_{1:K}$, $s_{1:K}$, $\mu_{1:K}$, and $b_{1:K}$ are assigned appropriate
priors, and
where $\text{logistic}(x;m,s,\mu)$ $=$ $\frac{m}{1+\exp(-s(x-\mu))}$.  This
model may accurately describe general trends in the system, but it may be
ultimately misspecified, and underfit as the number of observations $n$ grows.

Alternatively, we could model this system as a black box using a Gaussian
process.  The GP posterior predictive may converge to the correct landscape
given enough data, but it is nonparametric, and does not encode our
assumptions. 

We can use the BPoE model to combine both the phase shift and GP models. We see
in Fig.~\ref{fig:combine} that when $n=2$ (first row), the BPoE model resembles
the phase shift model, but when $n=50$ (second row), it more closely resembles
the true landscape modeled by the GP.

\newlength{\tempheight}
\newcommand{\rowname}[1]
{\rotatebox{90}{\makebox[\tempheight][c]{#1}}}

\newlength{\tempwidth}
\newcommand{\columnname}[1]
{\makebox[\tempwidth][c]{#1}}
\newcommand{\columnnameleft}[1]
{\makebox[\tempwidth][l]{#1}}

\newcommand{\plotwidthThird}{0.3\textwidth} 
\newcommand{\plotwidthQuarter}{0.25\textwidth} 
\newcommand{\plotwidthOneColHalf}{0.23\textwidth} 
\newcommand{\betweenspace}{\hspace{-0.1in}}
\newcommand{\leftspace}{\hspace{-5mm}}

\begin{figure*}
\setlength{\tempwidth}{0.3\textwidth}
\settoheight{\tempheight}{\includegraphics[width=\tempwidth]{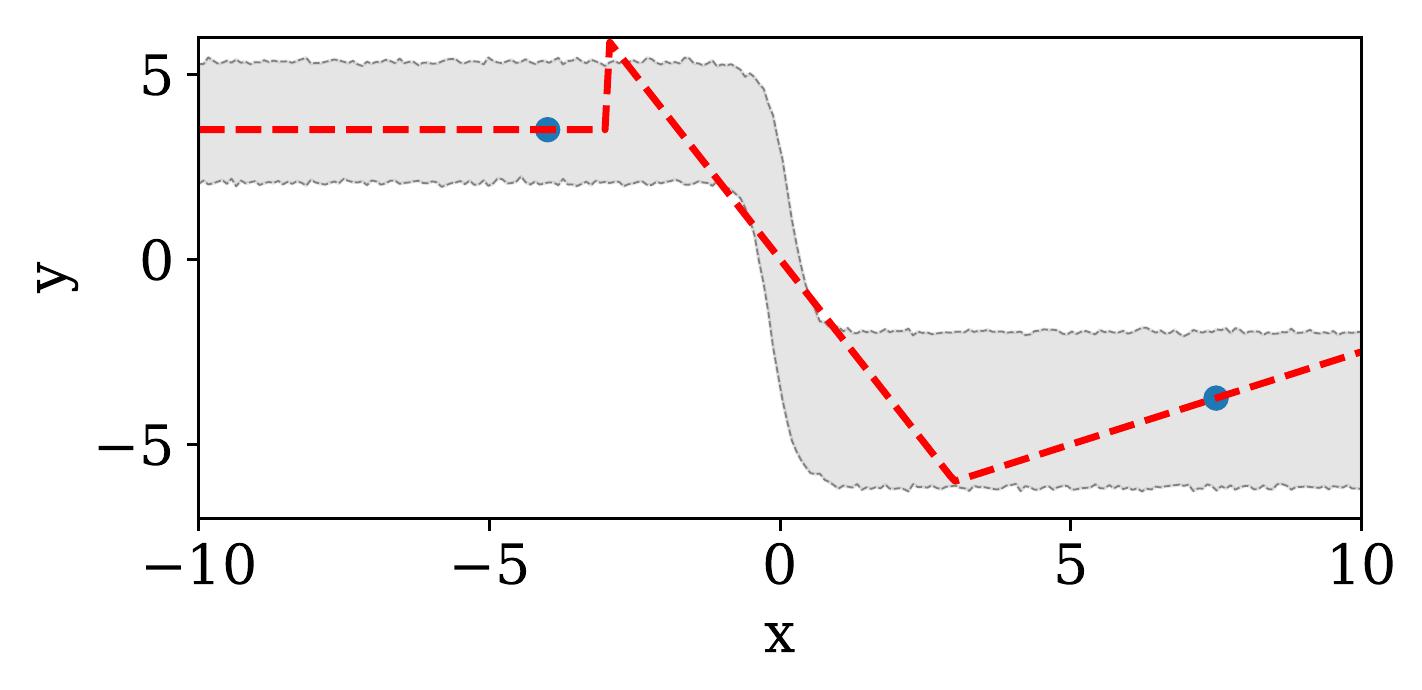}}%

\begin{center}
\hspace{0.3in}
\columnname{Phase Shift (PS) Model}
\betweenspace
\hspace{0.1in}
\columnname{Gaussian Process (GP)}
\betweenspace
\columnname{BPoE Ensemble (PS, GP)}\\

\vspace{-0.1in}

\leftspace
\rowname{\hspace{2mm}\underline{\parbox[c][0.2in]{0in}{}\hspace{4mm}$n$ = 2\hspace{4mm}}} \hspace{-0.1in}
\subfloat{
\includegraphics[width=\plotwidthThird]{fig/combine01}
\label{fig:combine1}}\betweenspace
\subfloat{
\includegraphics[width=\plotwidthThird]{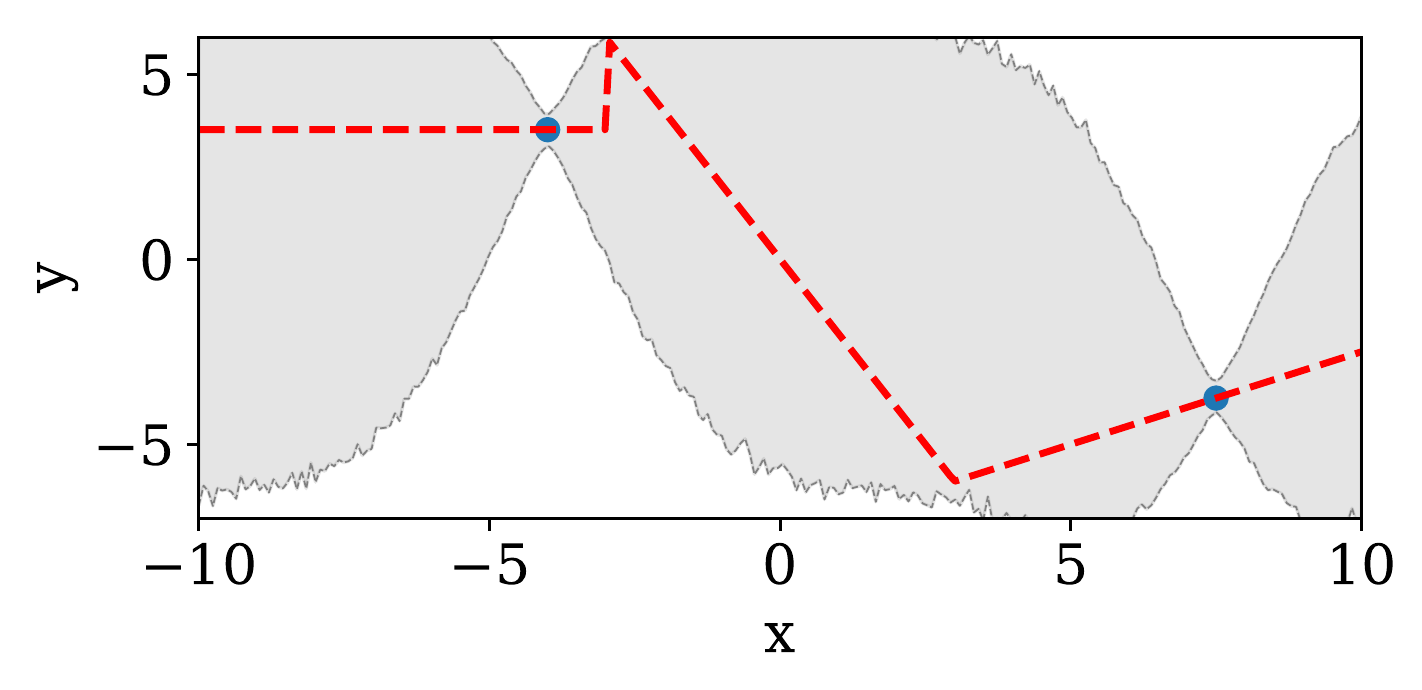}
\label{fig:combine2}}\betweenspace
\subfloat{
\includegraphics[width=\plotwidthThird]{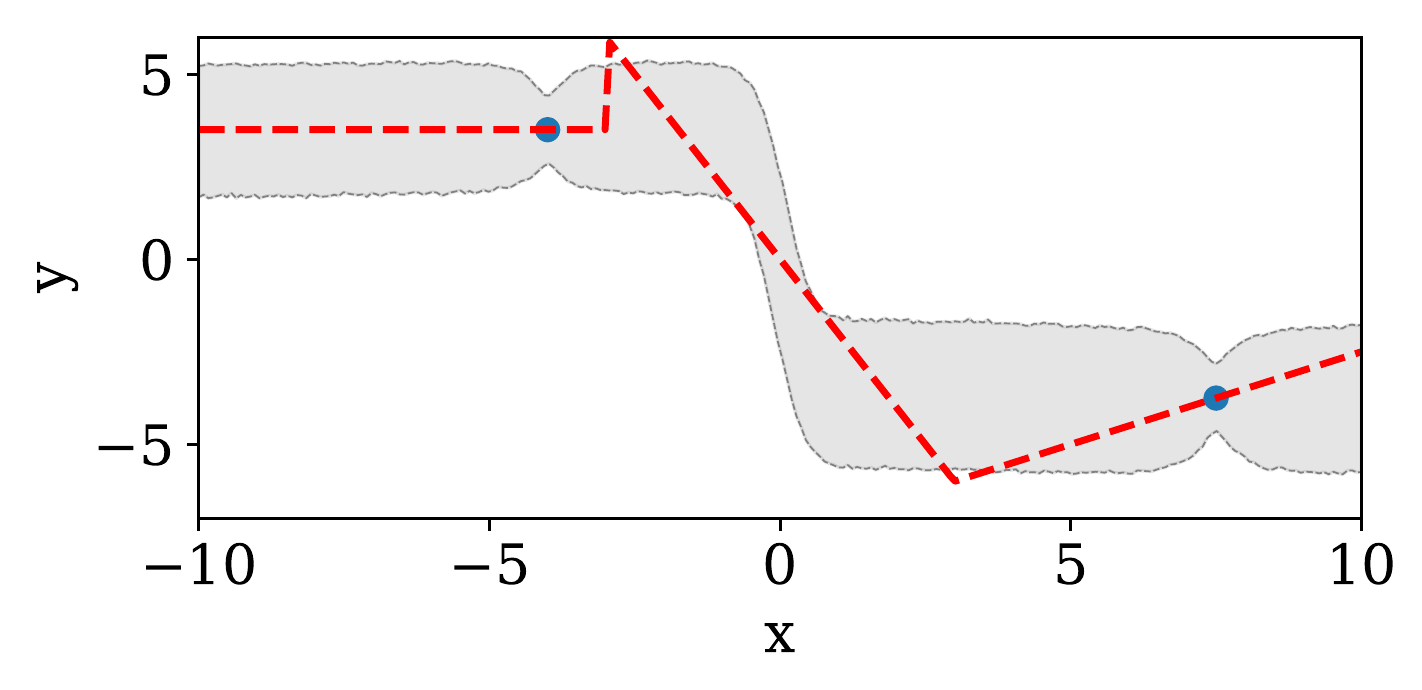}
\label{fig:combine3}}\betweenspace
\\
\vspace{-0.25in}
\leftspace
\rowname{\hspace{2mm}\underline{\parbox[c][0.2in]{0in}{}\hspace{4mm}$n$ = 50\hspace{4mm}}} \hspace{-0.1in}
\subfloat{
\includegraphics[width=\plotwidthThird]{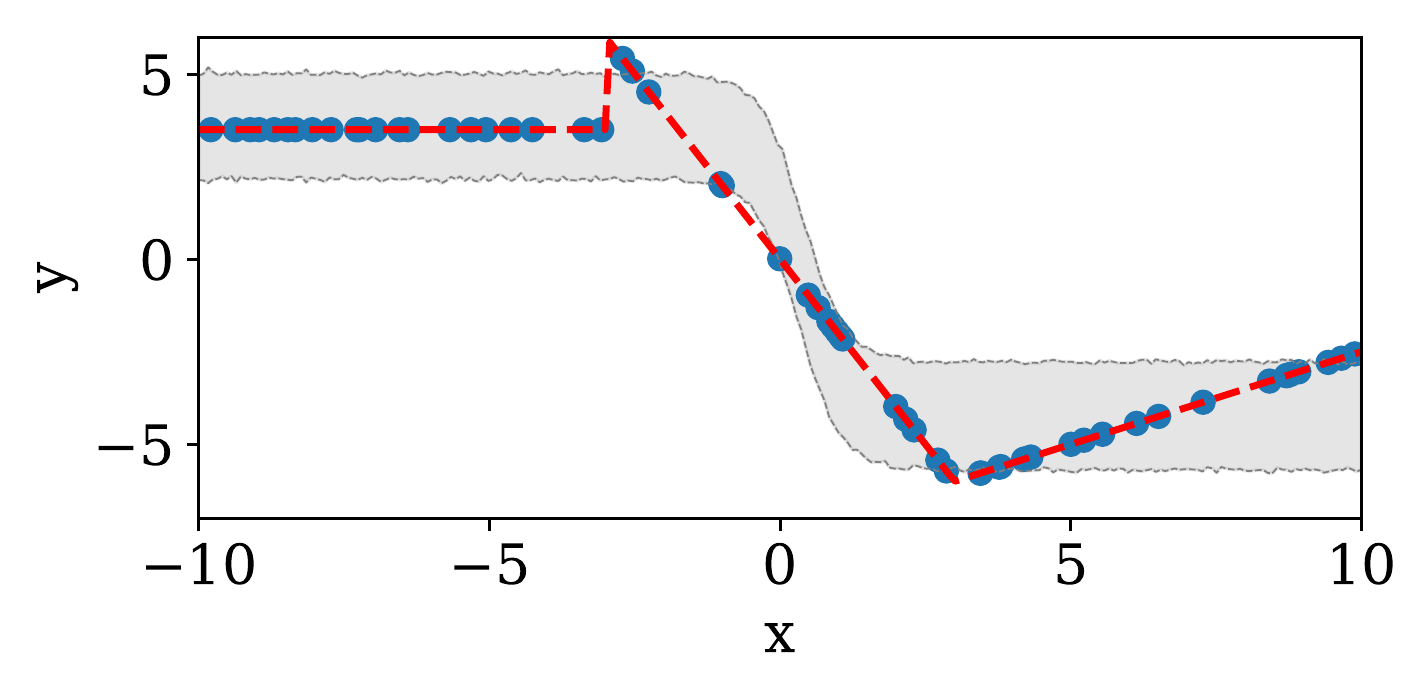}
\label{fig:combine4}}\betweenspace
\subfloat{
\includegraphics[width=\plotwidthThird]{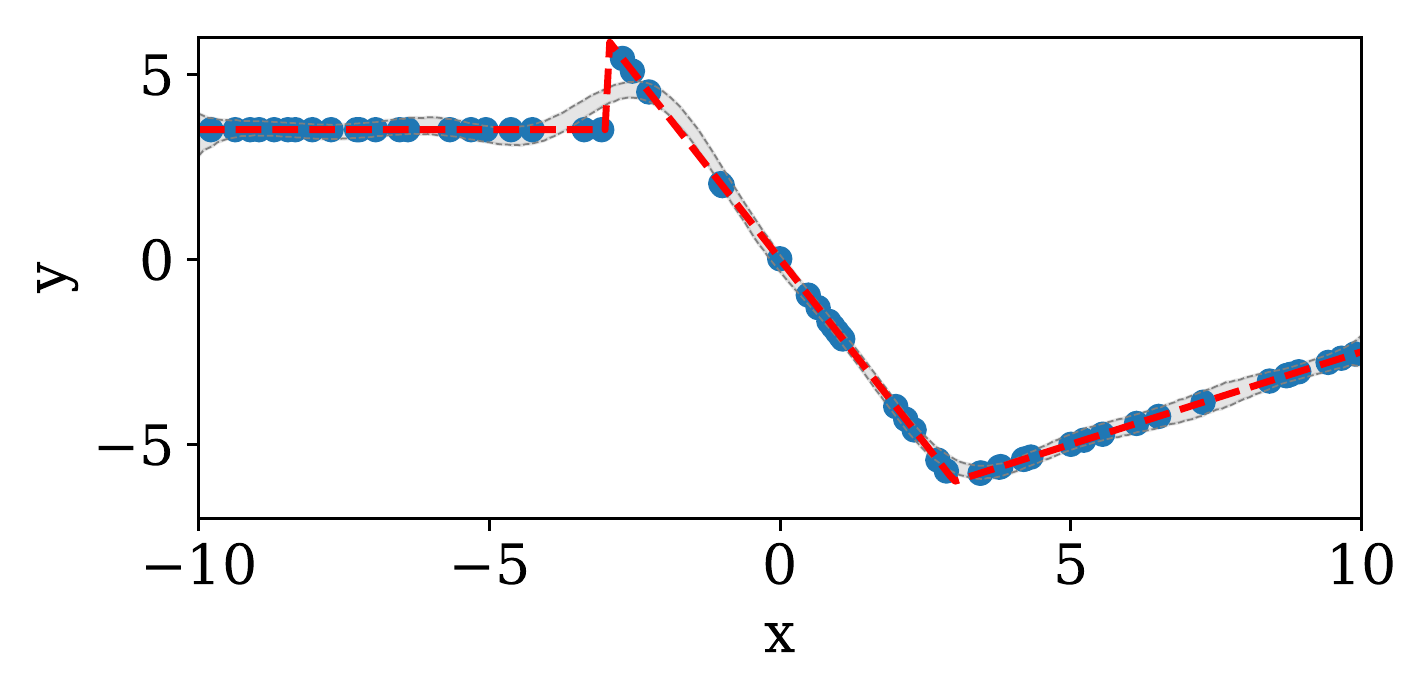}
\label{fig:combine5}}\betweenspace
\subfloat{
\includegraphics[width=\plotwidthThird]{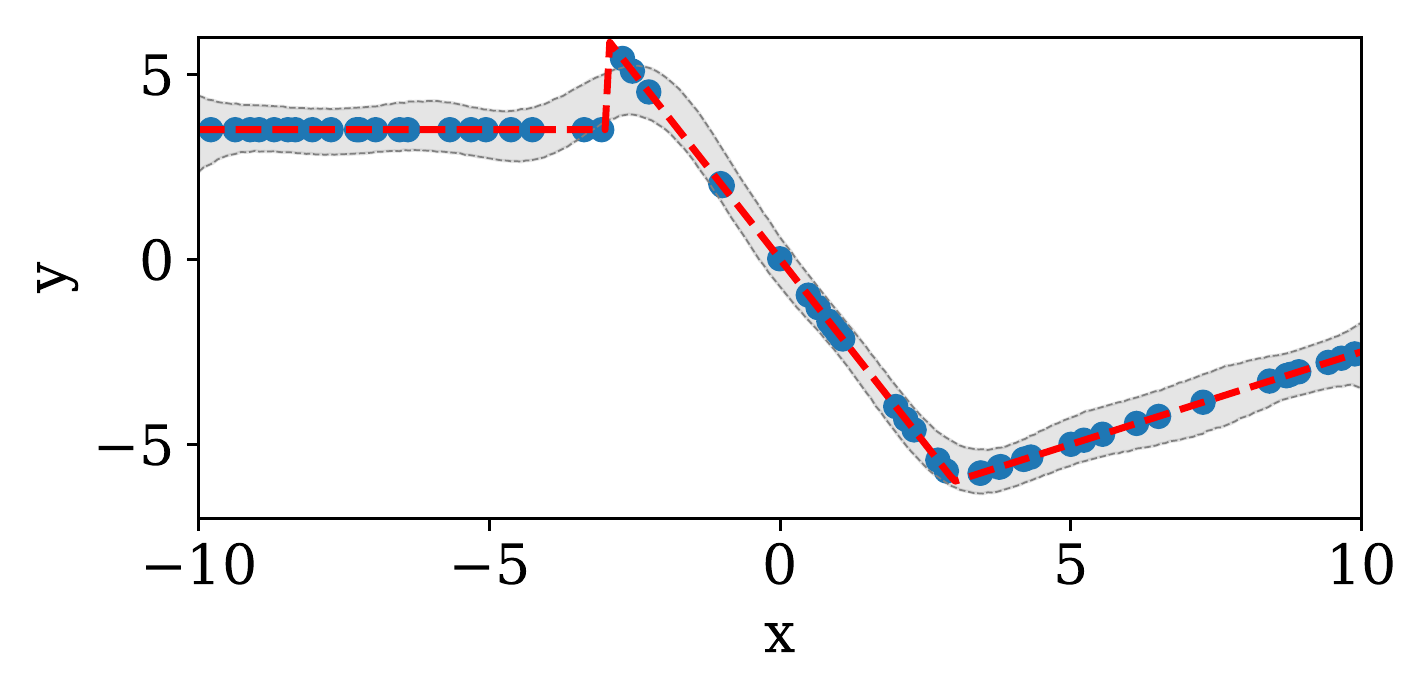}
\label{fig:combine6}}\betweenspace
\caption{
\small 
Visualization of the Bayesian product of experts (BPoE) ensemble model (column
3) of a phase shift (PS) model (column 1), defined in
Sec.~\ref{sec:phaseshift}, and a GP (column two). In the first row ($n=2$),
when $n$ is small, the BPoE ensemble more closely resembles the PS model.  In
the second row ($n=50$), when $n$ is larger, the BPoE ensemble more closely
resembles the GP model, and both accurately reflect the true landscape (red
dashed line). In all figures, the posterior predictive is shown in gray.
\label{fig:combine}
}
\end{center}
\end{figure*}

\subsection{Combination Algorithms for the \texttt{ensemble} Operation
  (Alg.~\ref{alg:bpoe})}
\label{sec:app_combine}

We make use of combination algorithms from the embarrassingly parallel MCMC
literature \cite{neiswanger2013asymptotically2, wang2015parallelizing,
neiswanger2017post}, to define the \texttt{ensemble} operation
(Alg.~\ref{alg:bpoe}) for use in applying the \probosp framework to a BPoE
model. We describe these combination algorithms here in more detail.

For convenience, we describe these methods for two Bayesian models,
$\mathcal{M}_1$ and $\mathcal{M}_2$, though these methods apply similarly to an
abitrarily large set of models.

The goal of these combination methods is to combine a set of $M$ samples
$y_{1,1:M} \sim p_1(y|\mathcal{D}_n;x)$
from the posterior predictive distribution of a model $\mathcal{M}_1$, with 
a disjoint set of $M$ samples
$y_{2,1:M} \sim p_2(y|\mathcal{D}_n;x)$
from the posterior predictive distribution of a model $\mathcal{M}_2$,
to produce samples
\begin{align}
  y_{3,1:M} \sim p(y|\mathcal{D}_n;x) \propto p_1(y|\mathcal{D}_n;x) p_2(y|\mathcal{D}_n;x),
\end{align}
where $p(y|\mathcal{D}_n;x)$ denotes the posterior predictive distribution of a
BPoE ensemble model $\mathcal{M}_e$, with constituent models $\mathcal{M}_1$
and $\mathcal{M}_2$.

We use the notation $\text{Combine}(y_{1,1:M},y_{2,1:M})$ to denote a
combination algorithm. We give a combination algorithm in
Alg.~\ref{alg:combine} for our setting based on a combination algorithm
presented in \cite{neiswanger2013asymptotically2}.

{\centering
\begin{algorithm}[H]
    \caption{\hspace{1mm} $\text{Combine}(y_{1,1:M},y_{2,1:M}))$  \hfill$\triangleright$ Combine sample sets}
    \label{alg:combine}
    \begin{algorithmic}[1]
      \State $t_1, t_2 \overset{\text{iid}}{\sim} \text{Unif}\left(\{1,\ldots,M\}\right)$
      \For{$i=1,\ldots,M$}
        \State $c_1, c_2 \overset{\text{iid}}{\sim} \text{Unif}\left(\{1,\ldots,M\}\right)$
        \State $u \sim \text{Unif}\left([0,1]\right)$
        \If{$u > \frac{w_{(c_1,c_2)}}{w_{(t_1,t_2)}}$}
          \State $t_1 \leftarrow c_1$
          \State $t_2 \leftarrow c_2$
        \EndIf
        \State $y_{3,i} \sim \mathcal{N} \left( \bar{y}_{(t_1,t_2)},\frac{i^{-1/2}}{2} \right)$
      \EndFor
      \State Return $y_{3,1:M}$.
    \end{algorithmic}
\end{algorithm}
}

\vspace{3mm}
We must define a couple of terms used in Alg.~\ref{alg:combine}. The
mean output $\bar{y}_{(t_1,t_2)}$, for indices $t_1,t_2 \in \{1,\ldots,M\}$, is
defined to be
\begin{align}
  \bar{y}_{(t_1,t_2)} = \frac{1}{2} \left( y_{1,t_1} + y_{2,t_2} \right),
\end{align}
and weights $w_{(t_1,t_2)}$ (alternatively, $w_{(c_1,c_2)}$), for indices $t_1,t_2
\in \{1,\ldots,M\}$, are defined to be
\begin{align}
  w_{(t_1,t_2)} =
  \mathcal{N}\left(y_{1,t_1}|\bar{y}_{(t_1,t_2)},i^{-1/2}\right)
  \mathcal{N}\left(y_{2,t_2}|\bar{y}_{(t_1,t_2)},i^{-1/2}\right).
\end{align}

Note that this $\text{Combine}(y_{1,1:M},y_{2,1:M})$ algorithm
(Alg.~\ref{alg:combine}) holds for sample sets from two arbitrary posterior
predictive distributions $p_1(y|\mathcal{D}_n;x)$ and $p_2(y|\mathcal{D}_n;x)$,
without any parametric assumptions such as Gaussianity.

\end{changemargin}
\end{document}